\documentclass[letterpaper,9pt]{article}
\usepackage[numbers]{natbib} 
\usepackage{authblk}
\usepackage{caption}
\usepackage{tabularx} 
\usepackage{amsmath,amssymb,slashed}   
\usepackage{graphicx} 
\usepackage[margin=1in,letterpaper]{geometry} 
\usepackage{tcolorbox} 
\usepackage{xcolor} 
\usepackage{hyperref} 
\usepackage{dirtytalk}

\tcbuselibrary{breakable}
\usepackage{pythonhighlight}
\usepackage{cancel}
\usepackage{multirow}
\usepackage{booktabs}
\usepackage{listings}

\usepackage{pgfplots}
\usepackage{tikz}
\usepackage{multirow}
\usepackage{booktabs}

\usepackage{listings}
\usepackage{tikz-feynman}
\usetikzlibrary{decorations.pathmorphing}

\tikzset{snake it/.style={decorate, decoration=snake}}

\hypersetup{
	colorlinks=true,       
	linkcolor=blue,        
	citecolor=blue,        
	filecolor=magenta,     
	urlcolor=blue         
}

\newcommand{\moritz}[1]{\textcolor{magenta}{[MM: #1]}}

\newcommand{\numberthis}{\stepcounter{equation}\tag{\theequation}}
\definecolor{standardcolor}{RGB}{255, 243, 205}  
\definecolor{attemptcolor}{RGB}{240, 240, 240}  
\definecolor{problemcolor}{RGB}{230, 245, 255}  

\definecolor{usercolor}{RGB}{33, 150, 243} 
\definecolor{modelcolor}{RGB}{76, 175, 80} 
\definecolor{expertcolor}{RGB}{220, 20, 60} 

\tcbuselibrary{breakable} 

\tcbset{
        colback=gray!10,       
        colframe=gray!40,      
        boxrule=0.3mm,         
        arc=3mm,               
        before skip=10pt,      
        after skip=10pt,  
        breakable,
        fonttitle=\bfseries,   
        coltitle=black,        
        colbacktitle=gray!20,  
        attach title to upper, 
}

\title{Theoretical Physics Benchmark (TPBench) - a Dataset and Study of AI Reasoning Capabilities in Theoretical Physics}

\author[1]{Daniel J.H. Chung} 
\author[2]{Zhiqi Gao} 
\author[1]{Yurii Kvasiuk} 
\author[1]{Tianyi Li} 
\author[1,5]{Moritz M\"unchmeyer} 
\author[3]{Maja Rudolph} 
\author[2]{Frederic Sala} 
\author[4]{Sai Chaitanya Tadepalli}

\affil[1]{Department of Physics, University of Wisconsin-Madison}
\affil[2]{Department of Computer Science, University of Wisconsin-Madison}
\affil[3]{Data Science Institute (DSI), University of Wisconsin-Madison}
\affil[4]{Department of Physics, Indiana University, Bloomington}
\affil[5]{NSF-Simons AI Institute for the Sky (SkAI), Chicago}

\def\s8m{\texttt{s8\_m} }
\def\s8p{\texttt{s8\_p} }

\begin{document}

\maketitle

\begin{abstract}
We introduce a benchmark to evaluate the capability of AI to solve problems in theoretical physics, focusing on high-energy theory and cosmology. The first iteration of our benchmark consists of 57 problems of varying difficulty, from undergraduate to research level. These problems are novel in the sense that they do not come from public problem collections.
We evaluate our data set on various open and closed language models, including o3-mini, o1, DeepSeek-R1, GPT-4o and versions of Llama and Qwen. While we find impressive progress in model performance with the most recent models, our research-level difficulty problems are mostly unsolved. We address challenges of auto-verifiability and grading, and discuss common failure modes. 
While currently state-of-the art models are still of limited use for researchers, our results show that AI assisted theoretical physics research may become possible in the near future. We discuss the main obstacles towards this goal and possible strategies to overcome them. The public problems and solutions, results for various models, and updates to the data set and score distribution, are available on the website of the dataset \url{tpbench.org}.

\vspace{7cm}

\begin{center}
\normalsize
    Website: \url{tpbench.org} \\
    Corresponding Email: \texttt{research@tpbench.org}
\end{center}

\clearpage

\tableofcontents

\clearpage

\end{abstract}

\section{Introduction}

Automated mathematical reasoning at research level with AI in theoretical physics 
may now be within reach.
Novel large language model (LLM)-based AI systems, powered by improved AI reasoning techniques at training and inference time, are potentially powerful tools for the theoretical physics community. If substantial parts of the theoretical research process could be performed by AI, this would allow to significantly accelerate progress in theoretical physics. If AI could act as a fast, reliable and skilled research assistant that can perform theoretical calculations and solve mathematical problems, human researchers could cover substantially more theoretical ground, evaluate more ideas for their promise, and thus make more theoretical discoveries. Even without super-human intelligence, an AI \say{craftsman} would allow humans to outsource tedious calculation work and to focus more on creative aspects of the theoretical research process. 

Recent advancements in LLMs have allowed models to solve progressively more difficult tasks that require abstract mathematical reasoning. While high-school level math competition benchmarks like \texttt{MATH} \cite{hendrycks2021measuring} are almost saturated by current models, the focus has recently turned to graduate level and research level mathematics. A main data set in this domain, the recently introduced \texttt{FrontierMath} \cite{glazer2024frontiermath}, which contains research level difficulty problems, is still mostly unsolved by frontier models. In theoretical physics (TP), which also requires extensive abstract mathematical reasoning, there has been comparatively less work than in mathematics. Existing benchmarks which include physics such as \texttt{JEEBench} \cite{arora2023llmsadvancedenoughchallenging}, \texttt{OlympiadBench} \cite{he2024olympiadbenchchallengingbenchmarkpromoting} and \texttt{PhysicsQA} \cite{jaiswal2024improvingphysicsreasoninglarge}, cover mostly high-school-level problems from college entrance exams or competitions. There is little existing work on mathematical reasoning for theoretical physics at graduate or research level. An exception is \cite{pan2024quantummanybodyphysicscalculations}, where the authors evaluate the performance of large language models for symbolic calculations in quantum many-body physics, however in the narrow context of a specific physical setting. Very recently, the \texttt{Humanity's Last Exam} dataset \cite{phan2025humanity} (HLE) appeared as a multi-domain benchmark that includes problems from theoretical physics. We provide a more complete list of available data sets in Sec. \ref{sec:related_benchmarks}. 

In the present work, we build a data set to test theoretical physics reasoning skill over a broad range of difficulty. We aim to answer the following questions:
\begin{itemize}
    \item How good is the current state-of-the-art AI for problem-solving in TP? Are existing models useful for research-level reasoning? 
    \item What are the most common failure modes? For example, are models performing correct reasoning but fail mostly at algebra (at which LLMs are known to perform poorly)? 
\end{itemize}
To answer these questions, we created a new benchmark data set \texttt{TPBench} of theoretical physics problems of varying degree of difficulty, from advanced undergraduate to research level. Our problems are novel, in the sense that they do not come from public problem collections (see Sec. \ref{sec:novelty} for detailed comments). For graduate level and research problems we focus in particular on problems from high-energy physics and cosmology. 
An important property of our data set is that it provides a \emph{continuum} of problem difficulty, from easy to research level, which few mathematical data sets do. This allows us to compare the performance of different models over a wide spectrum of difficulty. We invite the reader to skip ahead to App. \ref{app:sampleproblems} to get an impression of the difficulty of these problems. Before discussing our data set in detail, we begin with some general remarks about reasoning for TP and its relation to AI models. 

\paragraph{Differences between reasoning in math and TP.} 
Because TP is extremely broad and math is arguably even broader, any summary discussion of the differences between mathematical and physics reasoning is unlikely to be accurate in many examples in a generic comparison set. Nevertheless, in terms of modern graduate level and higher physics and mathematics comparisons, several aspects typically stand out. 

\begin{itemize}
\item Mathematical reasoning tends to focus on establishing exact broad statements constructed within a rigid logical framework, while TP reasoning mostly deals with approximate narrower statements constructed within a logical framework in which some of the less quantitatively relevant details are left unspecified but ``most likely" can be filled in such that the statements can be made arbitrarily precisely if desired.\footnote{Certain corners of TP such as formal general relativity and string theory come very close to the reasoning style of mathematics (e.g. \cite{Iyer:1994ys,Geroch:1968zm,Kontsevich:1992ti,Schon:1981vd,Aganagic:2024sww,Parker:1981uy,Hausel:2002ap}).  This will not be treated here, and this in some sense is covered by the LLM literature dealing with mathematics.}  This difference naturally stems from the different approximate goals of each discipline: a commonly accepted goal of TP is to model nature while a commonly accepted goal of mathematics to construct nontrivial, beautiful true statements connecting surprisingly disparate ideas \cite{Hardy1940}. The emphasis on rigidity is what naturally leads to the format of theorems and proofs in mathematics while the emphasis on quantitative modeling has allowed the Standard Model of particle physics to make successful predictions despite the evolving nature of its underlying mathematical structure.

\item TP reasoning primarily relies on techniques of direct computations, while mathematical reasoning tends to use more often indirect techniques such as contradiction and induction.  More explicitly, TP computations often utilize algorithmic methods in calculus, linear algebra, complex analysis, differential equations, differential geometry, and group representation theory.

\item TP reasoning often focuses on derivations of formulas whose parametric dependences as well as the overall normalization are implicitly defined in a narrow domain of physical relevance.  For example, if one writes down a quantum field theory Lagrangian and computes observables, the coupling constants with conventional normalization cannot be a large number such as 1000 since such theories are expected to have the field degrees of freedom reorganize into a different effective theory.  However, the exact parametric range of validity for the coupling constant is left implicit.  This is in contrast with much of mathematical reasoning, where parametric ranges are precisely defined.  This makes TP reasoning quite efficient at the expense of imprecision in the domain of validity.

\item TP typically focuses on approximations whose quantitative uncertainties are often left unspecified.  For example, one of the most popular computational techniques in TP is perturbation theory, a type of asymptotic expansion, which often has a zero radius of convergence, and because there is often no exact computation to compare to, there is no rigorous quantitative estimate of uncertainties in most cases.  One typically understands the estimate of the uncertainty to be the next order contribution in perturbation theory.  Researchers also implicitly understand that there are non-perturbative contributions such as instantons which have an exact representation of zero in perturbation theory that can become important in certain instances.

\end{itemize}
These properties make theoretical physics an exciting testbed for AI reasoning models, which has not been extensively explored, perhaps because models were not powerful enough to do so, until very recently. 

\paragraph{Generating novel research ideas/problems in TP.}  Novel research in TP, as in all fields of science, is usually incremental, and novel research ideas are  combinations or further developments of prior work. For example, once a novel method has been invented, it can often be applied to many different problems. Indeed, Feynman advised to keep a list of favorite problems, and to check whether any newly learned technique could be useful for one of these problems \cite{Rota1997}. Experienced researchers have an advantage over students at generating interesting research because their knowledge base is much larger and more interconnected.  Indeed, what makes a research level question different from a classroom question is often the novelty and connection with existing knowledge and not the reasoning difficulty.  It seems very plausible that machine learning models, with their ability to ingest vast amounts of knowledge during training or inference, could be particularly strong at finding promising combinations of novel results and techniques. A recent study in NLP research \cite{si2024can} found that LLM research ideas are rated more novel (but slightly less feasible) by human experts than human expert ideas.  Experienced researchers are also able to judge whether a mathematical result is interesting or surprising and deserves further investigation.  Such \say{theoretical taste} may be beyond existing AI models. With our data set, we are not currently aiming to test these aspects of theoretical research.

\paragraph{Reasoning abilities required to solve research problems in TP.} Researchers (consciously or unconsciously) have a number of techniques or heuristics to solve theoretical problems. A famous collection of problem solving techniques and advise is George Polya's book \textit{How to solve it} \cite{Polya1945} which lists about 50 heuristics with suitable examples in mathematics. Techniques include decomposing the problem, finding a related problem, generalization, and many less obvious ones. Most researchers have a more limited toolkit than Polya and many novel papers are somewhat straight forward combinations of reasoning steps contained in previous works. A main difficulty in this case is to understand this prior work and be able to recall and connect it when needed. Of course, insights are also often re-discovered independently. When solving a hard problem, researchers may try many different paths or heuristics, jump back and forth in their reasoning chain, analyse examples, answer subquestions, clear up their misunderstandings, read related literature, etc. In principle, given a large enough context window for prior thoughts and unlimited inference time, LLMs may be able to perform such very long thought processes, but currently available models (with a public reasoning chain) do not show very deep thought processes in our experience.

\paragraph{Technical (calculation) abilities required to solve research problems in TP.} Once a mathematical reasoning step has been proposed, it needs to be executed correctly. This step is in principle straightforward but error-prone for most humans. For example, one may decide to Taylor expand an expression to third order, perform a Gaussian integral, re-arrange terms, or even just multiply numbers. LLMs are well known to perform poorly at such tasks, but this problem can in principle be fixed by using computer algebra systems, if they can work with the required mathematical objects (which however is often not the case in TP).

\paragraph{Observations from our evaluation.} We list some observations from our experiments, which we discuss in more details in the following sections.
\begin{itemize}
    \item Progress has been very rapid with the most recent models. When we initiated this project, GPT-4o \cite{GPT-4o} (released on May 2024) was state-of-the-art and unable to solve almost any TP problem beyond undergraduate level. When the o1-preview model \cite{o1-preview} (released on Sep 2024) appeared, it could solve many easy graduate level problems, but rarely any harder ones. The o3-mini series \cite{o3-mini} (released on Jan 2025), is able to solve about half of our advanced graduate level problems and even a few research problems. Nevertheless, as we will see, research problems involving long mathematical arguments are generally unsolved. 
    \item Symbolic calculation mistakes. Existing models are known to perform poorly at mathematical calculations (see e.g. \cite{imani2023mathprompter}), which could be performed correctly with a computer algebra system such as \texttt{SymPy} or \texttt{Mathematica}. Such wrong intermediate results then lead to incorrect followup reasoning. It should be noted that humans tend to make similar mistakes in calculations, but are often able to spot them on revisiting. We made an initial attempt to encourage symbolic verification with python, which we describe in \ref{sec:pythonhelp}, but found that it barely improved results. Better symbolic tool integration would be very beneficial for TP reasoning.   
    \item Logical mistakes and lack of information about uncertainty. LLMs are generally poor at self-correcting \cite{huang2023large} and typically cannot provide very useful information of where they are uncertain \cite{yin2023large}. Many techniques have been proposed to mark mistakes (such as asking a different model to verify) \cite{kamoi2024when,dhuliawala2023chain,zhang2023sac3,jiang2024llms}, and for mathematical reasoning it would be particularly important to improve and include them. For lengthy reasoning chains, logical errors are a significant problem because human experts often need to perform solutions in detail themselves before being able to spot errors. Humans are often aware where in a derivation they are uncertain and can ask for help, or investigate further themselves. 
\end{itemize}

The paper is organized as follows. In Sec. \ref{sec:tpbench} we discuss the properties of our data set, including the origin of problems and our approach to verification and grading. In Sec. \ref{sec:evaluation} we benchmark popular closed source and open source models on this data set. In Sec. \ref{sec:failuremode} we analyze the output of these models in more detail, and categorize their failure modes. In Sec. \ref{sec:relatedwork} we discuss related work. Finally in Sec. \ref{sec:discussion} we discuss future directions to improve AI-based reasoning in TP.

\section{Properties of TPBench}
\label{sec:tpbench}

\subsection{Overview}
We have curated a dataset of problems and associated solutions in main areas of TP. For research level problems we currently focus on high-energy theory and cosmology, the main expertise of the authors. Problems in our collection should have the following properties (similar to \texttt{FrontierMath} \cite{glazer2024frontiermath}):
\begin{itemize}
    \item The problem is well-posed and the solution to the problem is unambiguous. An expert in the field, after reading the solution, should not have any objections.
    \item The problem is original. The solution to the problem cannot be easily found in the existing literature.
    \item The answer should be auto-verifiable. This is easily achieved for numerical answers or simple algebraic expressions, but more difficult for tensor expressions. We discuss this property further below.
    \item It should not be possible to guess the answer or remember it from the literature, despite a wrong reasoning chain. 
\end{itemize}
It is hard to strictly enforce all these conditions in TP, as we discuss further below. Problem originality and the possibility to guess the answer can be judged differently by different researchers. For this reason we also provide metadata for each problem individually. We point out potential shortcomings in instances where we are aware of them. We include problems of varying degrees of difficulty, from undergraduate to graduate and to research problems. Naturally, research problems are more difficult to create, especially when requiring the answers to be novel and unpublished. Furthermore, more difficult problems are often more novel than easier problems (since the space of possible problems grows rapidly with their complexity). We discuss the aspect of novelty of our problems in more detail below, as well as individually in the problem metadata. We also make sure that our problems do not contain steps where a human would need a calculator to solve them (e.g. no floating point operations). 

We now discuss the attributes of our data set in more detail, including their statistical distribution. We aim to enlarge and diversify the data set further in the future. We also provide ten sample problems in App. \ref{app:sampleproblems} and we encourage the reader to browse the problems to get an impression of the whole data set.

\subsection{Problem Statistics}

The dataset is categorized into five difficulty levels: \textit{1 - easy undergrad}, \textit{2 - undergrad}, \textit{3 - easy grad}, \textit{4 - grad}, and \textit{5 - research}. This classification ensures that the dataset can accommodate a wide range of use cases, from introductory studies to cutting-edge research challenges. The distribution of problems across these difficulty levels is detailed in Table~\ref{tab:difficulty_levels}. For difficulty level 1-4 this means that the problem could appear in a homework problem or exam for students. For level 5, this problem could appear as a nontrivial step in a publication: i.e. our research level problems are sub-problems that would constitute a part of a publication, and are not by themselves large enough to constitute an entire publication. Solving level 4 and 5 problems would make models useful for theoretical research, but would not mean that models could write their own publishable papers (by a significant margin).  Indeed, one of the most important steps in TP research is establishing why a particular question is important and organizing a string of level 5 type of steps to answer that question. Future iterations of this data set could include more open-ended research problems, more reminiscent of a research publication.

\begin{table}[h]
    \centering
    \footnotesize
    \renewcommand{\arraystretch}{1.2}
    \begin{tabular}{l c c}
        \toprule
        Difficulty Level & Number of Problems & Percentage \\
        \midrule
        1 - Easy Undergrad & 8  & 14.0\% \\
        2 - Undergrad      & 13 & 22.8\% \\
        3 - Easy Grad      & 11 & 19.3\% \\
        4 - Grad/Easy Research  & 14 & 24.6\% \\
        5 - Research       & 11 & 19.3\% \\
        \bottomrule
    \end{tabular}
    \caption{Distribution of Problems by Difficulty Level}
    \label{tab:difficulty_levels}
\end{table}

The problems in the dataset span specialized domains, including \textit{cosmology}, \textit{high energy theory}, and \textit{general relativity}.  The less difficult problems span a wide area including astrophysics, electromagnetism, quantum mechanics, statistical mechanics, and classical mechanics.  This domain-specific focus ensures the dataset's relevance to 
theoretical 
research related to the fundamental laws of nature, while the less difficult problems allow us to establish as a baseline what a successful AI performance looks like. Table~\ref{tab:problem_domains} provides an overview of the distribution of problems by domain.  In the future, we aim to include problems from other domains of theoretical physics, such as condensed matter theory.

\begin{table}[h]
    \centering
    \footnotesize
    \renewcommand{\arraystretch}{1.2}
    \begin{tabular}{l c c}
        \toprule
        Domain & Number of Problems & Percentage \\
        \midrule
        Cosmology            & 19 & 33.3\% \\
        High Energy Theory   & 18 & 31.6\% \\
        General Relativity   & 4  & 7.0\% \\
        Other                & 16  & 28.1\% \\
        \bottomrule
    \end{tabular}
    \caption{Distribution of Problems by Domain. 
    The ``Other" category includes astrophysics, electromagnetism, quantum mechanics, statistical mechanics, and classical mechanics. Many problems are in between areas. For example some Cosmology problems could also be classified as High Energy Theory.
    } 
    \label{tab:problem_domains}
\end{table}

The dataset includes problems from various sources, in particular unpublished research, private coursework, and recently published research papers. Almost half of the problems are novel (e.g. most of the level 3, 4, and 5 problems), having been created specifically for this dataset, while others draw on 
course-related material of the authors. A small number of problems have been taken from very recent publications (eg. \cite{kvasiuk2024talefieldsneuralnetworkenhanced}).

\subsection{Auto-Verification of Solutions}
\label{sec:autoverifier}
To automate the evaluation pipeline, we developed a system inspired by how coding competitions validate their results. We introduced the requirement that the final answer to each problem be provided as a \texttt{Python} callable with the specified signature. We then developed a simple automatic (not LLM-based) grading agent that, given the model's answer and the correct solution, extracts the code, creates, and executes a consistency-check script. This approach allows for efficient evaluation of algebraic answers and automatically ensures that equivalent correct answers are classified as such. Additionally, it is flexible enough to verify answers involving a variety of special functions or answers that involve several outputs. In some problems, the natural system of units ($c=\hbar=1$) is specified in the prompt, while in other cases we pass constants of nature as function arguments to be unit agnostic. Alternatively, we could have adopted other automatic verification strategies. We could have provided numerical test cases in the prompt, but this would have led to lengthy problem statements, floating point operations, and much less flexibility. Another option is to consider multiple-choice answers, but this would make it easier to guess the answer without detailed understanding. Yet another possibility is to use another LLM as a grading agent and instruct it to compare the given solution to the true one. However we found that this approach is very error prone and LLMs are often not able to check mathematical equivalence of expressions (see below). 

Our proposed scheme gives the flexibility to check the variety of classes of answers exactly.
The verification process consists of three components:

\begin{enumerate}
    \item \textbf{Code Extraction:} The system extracts Python functions from both the model's solution and the expert solution.
    \item \textbf{Test Case Execution:} Both functions are executed with identical test inputs across multiple parameter combinations.
    \item \textbf{Output Comparison:} Results are compared numerically with appropriate tolerances for floating-point arithmetic.
\end{enumerate}

Each problem in our dataset is accompanied by a comprehensive set of test cases, carefully designed to probe both the physical validity and mathematical correctness of solutions. These test cases span different parameter ranges (e.g. negative or complex arguments where appropriate), to ensure thorough verification. 

To illustrate this approach, consider the following undergraduate-level example:
\begin{tcolorbox}
\footnotesize
\textcolor{usercolor}{\textbf{Problem Statement:}}
A photon with the energy $E$ scatters on an electron at rest at angle $\theta$ in the electron's reference frame. Find the angular frequency $\omega$ of the scattered photon. 

\textcolor{usercolor}{\textbf{Answer Requirements:}}
Provide the answer in the form of a \texttt{python} function with the following signature:
\begin{python}
#let c be the speed of light, m_e - electron mass, h_bar - reduced Planck constant
def omega_scattered(E: float, m_e:float, theta:float, c:float, h_bar:float) -> float:
    pass
\end{python}
\end{tcolorbox}

\begin{tcolorbox}[standard]
\footnotesize
\textcolor{expertcolor}{\textbf{Model Answer:}}
\begin{equation*}
    \boxed{\omega = \frac{1}{\frac{\hbar}{E}+\frac{\hbar}{mc^2}(1-\cos{\theta})}}
\end{equation*}
\begin{python}
import math
def omega_scattered(E: float, m_e:float, theta:float, c:float, h_bar:float) -> float:
    return 1/(h_bar/E + h_bar/(m_e*c**2)*(1-math.cos(theta)))
\end{python}
\end{tcolorbox}
This example demonstrates several key aspects of our auto-verification approach. First, the problem statement is clear and unambiguous, requiring a specific physical quantity ($\omega$) to be calculated. Second, the answer requirements explicitly specify the expected format of the solution, including the function signature and parameter types. This standardization enables automated testing across different parameter regimes. Third, the model answer provides both the analytical expression and its implementation in Python code, allowing for direct numerical verification.

Furthermore, our verification system incorporates several safeguards to ensure reliable evaluation:

\begin{itemize}
    \item \textbf{Timeout Mechanisms:} Each function execution is limited to a maximum runtime of 30 seconds. This prevents infinite loops based on the model's incorrect reasoning while allowing sufficient time for complex calculations.
    
    \item \textbf{Error Handling:} The system catches and classifies runtime exceptions, including syntax errors and memory issues. Invalid solutions are automatically flagged incorrect.
        
    \item \textbf{Parameter Space Coverage:} Test cases are generated to cover different regimes of the parameter space while maintaining numerical stability.
\end{itemize}

While our verification system works well for many problems, certain theoretical physics problems present challenges:

\begin{itemize}
    \item \textbf{Tensor Expressions:} Problems involving abstract tensor expressions (e.g., $R_\mu^{\phantom{\mu}\nu} =d\omega_\mu^{\phantom{\mu}\nu} +\omega_\mu^{\phantom{\mu}\alpha} \wedge \omega_\alpha^{\phantom{\alpha}\nu} $, $\mathcal{L} = \epsilon_{\mu\nu\rho\theta}F^{\mu\nu}F^{\rho\theta}$, or $\nabla_{\mu}T^{\mu\nu} =0$) often have multiple equivalent representations due to symmetries. For instance, the Riemann tensor $R_{\mu\nu\alpha\beta}$ exhibits several symmetries including certain index permutations and the Bianchi identity.
    
    \item \textbf{Differential Expressions:} Verification of expressions involving derivatives, especially of fields, presents special challenges. Derivative expressions must satisfy constraints such as the product rule, chain rule, metric compatibility, and recognize the group representation of the field the derivative acts on: e.g. $D_\mu \phi$ has a different elementary calculus expression than $D_\mu \psi$ even for the same gauge group and symbol $D_\mu$ if $\phi$ and $\psi$ have different representations of the group.  Indeed in situations where the intermediate result is a differential equation in a system with gauge invariances, knowing whether the two sets of differential equations (here the differential equation itself being the solution to the physics related problem) are physically equivalent can become nontrivial.

    \item \textbf{Integral Expressions:} Integral expressions would be even more difficult to check numerically than differential expressions.  Furthermore, they have many of the same challenges for verification as the differential expressions in terms of equivalence classes, as can be seen in the expression $\int_{\mathcal{M}} d(H\wedge B)=\int _{\partial \mathcal{M}} H\wedge B$.  For the typical case of vanishing fields at infinity, there are also equivalences up to total derivative terms: e.g. $\int [d\phi \wedge *d\phi + dC]=\int d^4x \sqrt{-g}\partial_\mu \phi \partial^\mu \phi $.

    \item \textbf{Manifolds:} Furthermore, in cases where the solution to the problem is a manifold (often expressed as a metric), there is an infinite number of different equivalent algebraic expressions depending upon the coordinates used.  An example of this can be seen in asking for a noncompact, static, spherically symmetric, asymptotically flat vacuum solution to the Einstein equations which has a Komar mass of $M$.  A more abstract related situation of difficult-to-identify equivalence class is when two quantum field theories can be mapped to one another by integrating in and out different degrees of freedom (which abstractly covers the situation of renormalization group equivalence as well). 
\end{itemize}
Although this list is common for TP problems in the literature, it can be extended depending on the classes of mathematical objects that need to be covered.  The obvious common theme is the wealth of equivalence classes that the verification system needs to be aware of if it were to be generally applicable.

In our current data set, we only include problems where the above issues do not occur, i.e. where the final answer is an algebraic expression without tensors, derivatives, integrals, or manifolds. Of course, these objects do occur in the solution, but not in the final answer. In the future, it would be interesting to develop auto-verifiers for expressions involving these more general mathematical objects listed above. We have reserved a number of such problems for future iterations of the dataset that would be useful for testing dedicated more general verification codes.

\subsection{AI-Based Holistic Grading of the Entire Solution}

In addition to auto-verification, we also employ AI-based grading. In this process, the grader model has access to both the expert-labeled solution and the LLM-generated solutions from a separate model, and is tasked with assigning grades (A-D). This approach mirrors how a human teaching assistant grades homework, where partial credit is given for correct reasoning steps, even if the final solution is incorrect. Moreover, holistic grading can identify instances where a solution arrives at the correct answer using incorrect reasoning, which occurs in a small number of our problems. While holistic grading is conceptually preferred, we observe significant disagreement between different grader models as well as humans.

\subsection{Novelty and Difficulty of Our Problems}
\label{sec:novelty}

Most of the problems presented here are constructed based on those given in standard courses as well as unpublished research related notes.  For example, the solution to the research-level problem ``One pole problem" (see App. \ref{app:onepole}), without steps explained, is given in a footnote of \cite{Basso:2022tpd}.  Most of the research level problems would be readily doable by a good TP graduate student, and some of these are not much different from hard problems in graduate courses whose problems and solutions can be found publicly. However, we have made significant efforts to construct or modify problem statements so that the answers cannot be found by web search. Most of the research-level problems use typical or not-too-atypical notation to simulate a research setting, although this may facilitate literature recall (rather than reasoning) by the model.\footnote{An interesting followup study would be to vary variable naming and other notation to evaluate this point.}

The difficulty of a problem can vary along different axes, i.e. problems may be easy or hard for different reasons. We aimed to provide a sampling of this space:

\begin{itemize}
\item Some of the problems are difficult for a human researcher because they may not know that a similar problem has already been solved in the literature. Indeed, almost all solution techniques used in literature evolve over time incrementally as people build upon results of previous related computations. This gives LLMs an advantage for many problems, especially if the problem statement makes it clear what literature knowledge is required (which we try to avoid). Fortunately, publications often omit minor reasoning steps, and asking the model for detailed mathematical derivation can thus reveal such literature memory. For examples of models solving difficult problems by using \say{superhuman literature knowledge} see Sec. \ref{sec:omodelperformance}. Indeed, a key challenge in constructing this data set was to avoid this phenomenon as much as possible, to reveal true reasoning. 

\item Another obvious often encountered difficulty in research is simply the accuracy of routine calculus/algebraic manipulations.  The probability of errors increases with the number of steps needed to reach the answer as well as the number of variables that are involved. LLMs are not currently performing very well with such long calculations.

\item More truly physical setting (e.g. experimental setting) related theoretical physics problems contain larger number of seemingly-disorganized set of variables, in contrast with more formal setting theoretical physics problems that contain a well-organized set of variables (typically using group theoretic structure).  Some of our problems have been designed specifically to test whether the AI can reason using a seemingly-disorganized set of variables. 

\item Some of the problems have been given with a great deal of contextual information (such as the \say{One pole problem} in App. \ref{app:onepole}), but others require a much more contextual interpretation (e.g. \ref{sec:prob_bias}). In some sense, such \say{less specific} problems are similar in difficulty as the problems requiring literature recall.  If the LLM pattern matches the words in the problem to solution patterns in the literature, the LLM can be deemed to have understood the context.

\item The \say{One pole problem} in App. \ref{app:onepole}) also tests diagrammatic reasoning skills which are slightly more abstract than Feynman rules. This is part of a small number of problems in our data set where humans would use the help of diagrams to reason through them, and their expert solutions sometimes contain diagrams, usually in the \texttt{TikZ} LaTeX format.  More generally, graphical languages such as TikZ (particularly with its Feynman diagrammatic extension TikZ-Feynman \cite{Ellis:2016jkw} and other such extensions) might be a good language with which to develop an LLM's graphical reasoning skills because of its efficiency in capturing the mathematical content of the diagrams.

\end{itemize}

\subsection{Public and Private Data Set and Data Leakage Concerns}

We make 10 of our problems and solutions public (see App. \ref{app:sampleproblems} and \url{tpbench.org}), two for each difficulty level, such that they can be used to understand the data set, develop inference algorithms and examine failure modes. Naturally these problems will be part of future training data. To deal with this challenge, we also keep a large part of our data set private, currently about 50 problems. If you would like to evaluate your model on our private data set, please contact the authors directly.  

Guaranteeing that private data does not end up in future training data is challenging. OpenAI, which we have used extensively, adds user interface chats to its training data but does not add API calls. Correspondingly, we have generally used API calls for querying problem solutions. However, in early phases of this projects, some problems were run in the user interface. In future iterations of this project, we will emphasize data leakage control further, especially for research level problems. We note that a small number of research problems is sufficient to evaluate significant model progress, as long as for these problems data set leakage control and originality of the problem are flawless. For our current problem set, we only enforce that problems (and especially solutions) do not appear publicly accessible online. Furthermore, we took particular care that expert solutions to problems were never passed to the ChatGPT user interface, where they could be added to future training data.

\section{Model Performance Evaluation}
\label{sec:evaluation}

In this section, we evaluate the performance of several leading models on our dataset, TPBench, across five different difficulty levels, ranging from undergraduate to research-level problems. Closed-source models include OpenAI GPT-4o, o1, and o3-mini \cite{o1-preview, GPT-4o,o3-mini}. Open-source models which we were able to run locally on our hardware include small and intermediate sized Llama 3.1, Qwen 2.5, and Qwen-QwQ, which is an experimental LLM that focused on advancing reasoning developed by the Qwen Team  \cite{Metallama3.1, Qwen2.5, QwenQwQ32b}. We also include the recent open-source reasoning model DeepSeek R1 \cite{guo2025deepseek} and its base-model DeepSeek V3 \cite{bi2024deepseek} which we ran on \texttt{Together AI }API. Finally, we tried to solve a subset of our research problems with OpenAI's Deep Research, including the problem in App.  \ref{app:onepole}, primarily to spot solutions that could be found online. Deep Research was not able to solve any of these research problems. We believe our subset of models is representative of the spectrum of current LLM capabilities. 

We provide the prompts for inference in the Appendix \ref{sec:llmprompt}. The complete model answers from all models, for the public problems, can be found on the \url{tpbench.org} website.
The evaluation considers two grading schemes: \textit{answer-only} and \textit{holistic}.
\begin{itemize}
\item \textbf{Answer-Only (Auto-Verified) Evaluation}.
In the answer-only evaluation, models are tasked with producing a final answer to the problem, where correctness is assessed based on whether the model’s answer matches the expected correct solution. This evaluation process is fully automated as decribed in \ref{sec:autoverifier}, with the correctness of the answer validated through numeric verification by the program.
\item \textbf{Holistic AI-Based Grading.}
In the holistic grading approach, we assess the reasoning process and the steps taken by the models. A separate LLM is provided with the problem statement, the expert solution, and the model's solution. It then evaluates the model's answer on a grading scale ranging from A to D. This grading scale accounts not only for the correctness of the final answer but also for the quality of reasoning, intermediate steps, and overall approach. Holistic grading is more lenient with minor errors or missing intermediate steps, and it provides partial credit for well-reasoned solutions even if the final answer is incorrect.
\end{itemize}

These two choices on their own are imperfect. The first one might consider a solution as correct which has two or more self-annihilating mistakes, or a solution that arrived at the correct answer with inconsistent or false reasoning. If the task is to evaluate the reasoning in challenging problem-solving, the binary grading system might not be representative to a satisfactory level. The second has the disadvantage of being somewhat arbitrary on assignments of grades for partial correctness. Our core results will use answer-only solutions.

\subsection{Results for Auto-Verified Solutions}

We begin by discussing the answer-only results, which are the key empirical results of this paper. Our results are obtained using zero-shot reasoning where the model is given the problem statement and expected to reason through it without any prior examples. In fact, few-shot learning can degrade general performance in reasoning models \cite{guo2025deepseek}. We have experimented with prompt optimization, but found no significant differences (see App. \ref{sec:llmprompt} for our prompts). 

\begin{figure}[h!]
    \centering
    \includegraphics[width=0.95\linewidth]{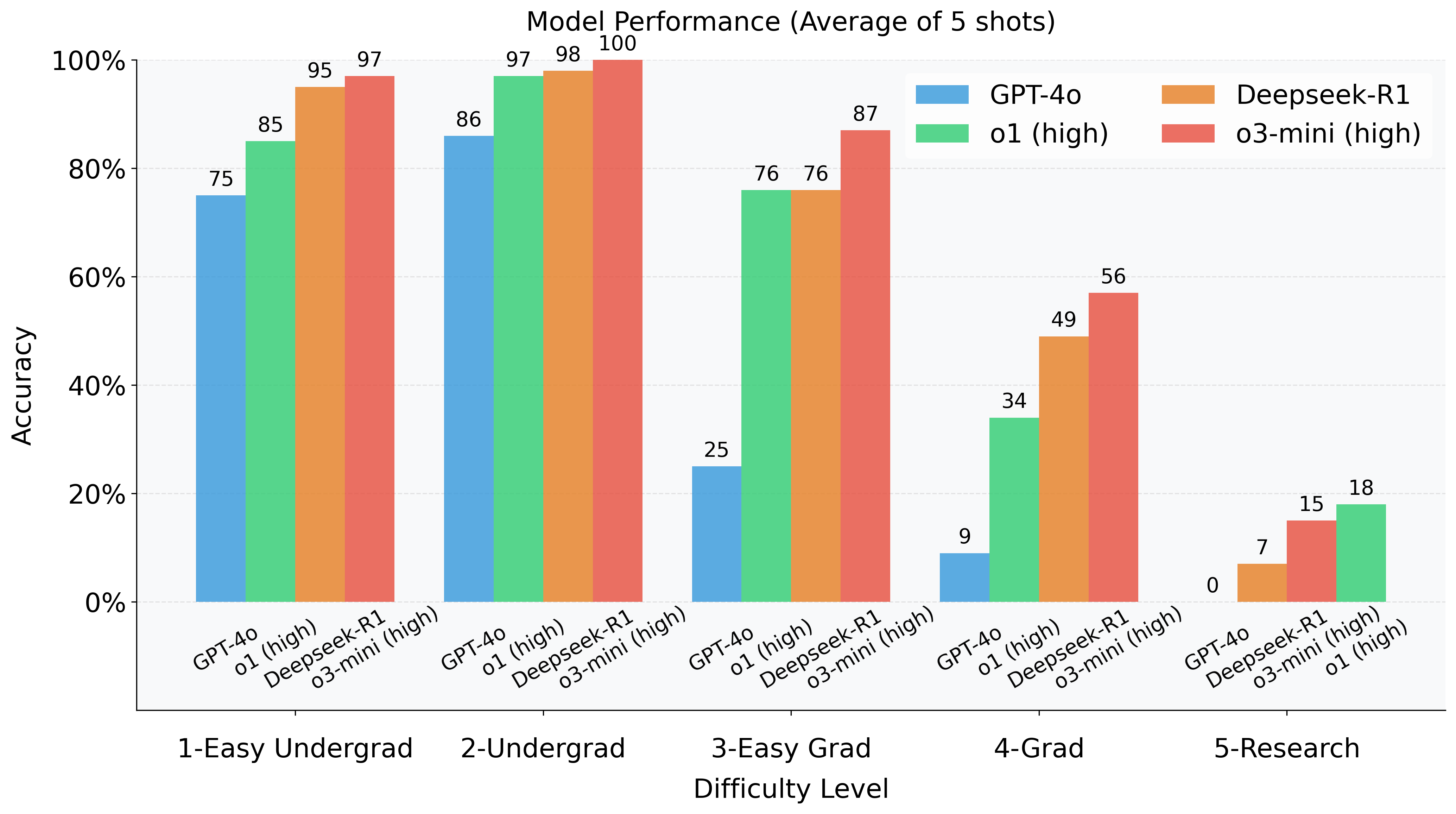}
    \captionsetup{justification=centering}
    \caption{Accuracy of SOTA Models by Difficulty Level.\\Note: ``high" in brackets indicates reasoning effort.}
    \label{fig:results_advanced}
\end{figure}

\begin{figure}[h!]
    \centering
    \includegraphics[width=0.95\linewidth]{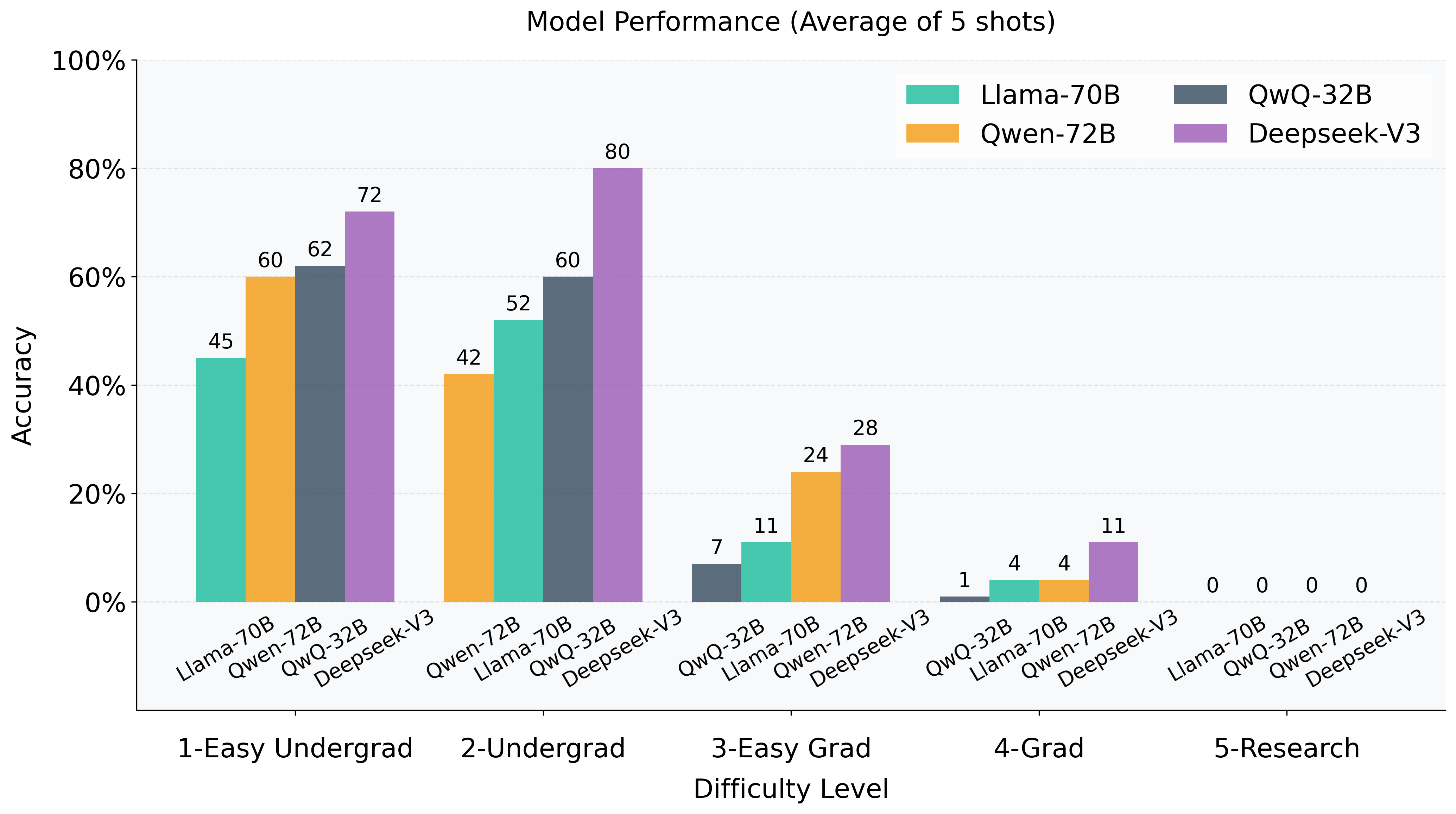}
    \caption{Accuracy of Common Open-Source Models by Difficulty Level}
    \label{fig:results_basic}
\end{figure}

Table~\ref{tab:model_performance} presents the performance of each model across various difficulty levels, ranging from easy undergraduate problems (Level 1) to research-level problems (Level 5). The table reports the percentage of problems solved by each model. 
The columns labeled \say{avg@5} represent the average score across five attempts, while the ``best@5" columns correspond to the average score of the best attempt out of five attempts. We visualize the \say{average of five} solution percentage in Fig. \ref{fig:results_advanced} (strong models) and Fig. \ref{fig:results_basic}. Finally, for our public problems, the individual results of models are given in App. \ref{app:sampleproblems}. For example, we include one level 5 research problem that top models can solve and one that they cannot. 

For the top models, o1, o3-mini and DeepSeek R1, undergraduate problems (level 1 and 2) are now essentially solved, with performance of $95\%$ to $100\%$ for the oX models. For easy graduate problems (level 3), the performance is around $80\%$. For our level 4 graduate problems, some of which could appear in research investigations, the best models o1 and o3-mini solve around $50\%$, with o3-mini slightly beating o1. Research problems are mostly unsolved at this stage with a score around $15\%$. o1 slightly beats o3-mini here, which may be due to it having a larger literature knowledge to draw on.  

Among mid-range models, GPT-4o and DeepSeek-V3 perform similary. They are between one and two levels of difficulty less powerful than the top models. Midrange models are essentially unable to solve problems above easy graduate level. Finally, lower parameter public models, which have the advantage that researchers can run them on invididual GPUs, cannot solve problems above undergraduate level. We also provide further model evaluation statistics of the data set on the website, including a unified model score over all difficulties.

\begin{table}[h!]
    \centering
    \scriptsize
    \begin{tabular}{lcccccccccc}
        \toprule
        \multirow{2}{*}{\textbf{Model}} & \multicolumn{2}{c}{\textbf{1-Easy Undergrad}} & \multicolumn{2}{c}{\textbf{2-Undergrad}} & \multicolumn{2}{c}{\textbf{3-Easy Grad}} & \multicolumn{2}{c}{\textbf{4-Grad}} & \multicolumn{2}{c}{\textbf{5-Research}} \\
        \cmidrule(lr){2-3} \cmidrule(lr){4-5} \cmidrule(lr){6-7} \cmidrule(lr){8-9} \cmidrule(lr){10-11}
        & avg@5 & best@5 & avg@5 & best@5 & avg@5 & best@5 & avg@5 & best@5 & avg@5 & best@5 \\
        \midrule
        GPT-4o   &0.75 (0.12)&0.88&0.86 (0.17)&1.00&0.25 (0.16)&0.45&0.09 (0.13)&0.29&0.00 (0.00)&0.00 \\
        o1 (high) &0.85 (0.05)&0.88&0.97 (0.04)&1.00&0.76 (0.24)&1.00&0.34 (0.13)&0.50&0.18 (0.07)&0.27 \\
        o3-mini (high) &0.97 (0.05)&1.00&1.00 (0.00)&1.00&0.87 (0.13)&1.00&0.57 (0.09)&0.64&0.15 (0.12)&0.27 \\
        DeepSeek-R1 &0.95 (0.06)&1.00&0.98 (0.03)&1.00&0.76 (0.23)&0.91&0.49 (0.20)&0.64&0.07 (0.08)&0.18 \\
        DeepSeek-V3 &0.72 (0.15)&0.88&0.80 (0.23)&1.00&0.29 (0.29)&0.64&0.11 (0.06)&0.21&0.00 (0.00)&0.00 \\
        Llama-3.1-8B &0.30 (0.06)&0.38&0.18 (0.20)&0.46&0.02 (0.04)&0.09&0.00 (0.00)&0.00&0.00 (0.00)&0.00 \\
        Llama-3.1-70B  &0.45 (0.36)&0.88&0.52 (0.22)&0.77&0.11 (0.11)&0.27&0.04 (0.06)&0.14&0.00 (0.00)&0.00\\
        Qwen2.5-7B &0.10 (0.11)&0.25&0.40 (0.21)&0.62&0.04 (0.07)&0.18&0.00 (0.00)&0.00&0.00 (0.00)&0.00\\
        Qwen2.5-72B &0.60 (0.11)&0.75&0.42 (0.23)&0.77&0.24 (0.16)&0.36&0.04 (0.06)&0.14&0.00 (0.00)&0.00 \\
        QwQ-32B  &0.62 (0.21)&0.75&0.60 (0.27)&0.92&0.07 (0.15)&0.36&0.01 (0.03)&0.07&0.00 (0.00)&0.00 \\
        \bottomrule
    \end{tabular}
    \caption{Fraction of problems solved for each difficulty for each model. Note: The number in the bracket is the average of model attempts' standard deviation per problems.} 
    \label{tab:model_performance}
\end{table}

\subsection{Results for Holistic AI-Based Grading}

Table~\ref{tab:python_verify_letter_grade} presents the results for the holistic AI-based grading, which involves assigning letter grades (A to D) based on the quality of reasoning and correctness of the solution. This grading is not limited to the final answer but considers the overall approach taken by the model in solving the problem.  We have used GPT-4o as a grader, as a currently mid-range model. We chose this model for cost efficiency reasons, and in the future we intend to use the most powerful model as a grader. The model was provided the grading prompt (App. \ref{sec:llmprompt}), the expert solution, and the model solution to grade, similar to the way a human teaching assistant would work. 

The models’ performances are shown across the five difficulty levels. The letter grades represent the models' ability to produce correct solutions while demonstrating sound reasoning. An ``A" indicates an excellent solution with minimal to no errors, a ``B" suggests a good solution with minor mistakes, ``C" indicates a solution with significant flaws, and ``D" represents a fundamentally incorrect solution. 

In principle, the holistic grading system provides insights into the models' reasoning capabilities beyond just final correctness. However, we find some difficulties with holistic grading as we now describe. This is consistent with results showing that LLM-as-a-judge approaches have considerable bias \cite{chen2024humansllmsjudgestudy}.

\begin{table}[h!]
    \centering
    \footnotesize
    \begin{tabular}{lcccc cccc cccc cccc cccc}
        \toprule
        \multirow{2}{*}{\textbf{Model}} 
        & \multicolumn{4}{c}{\textbf{1-Easy Undergrad}}
        & \multicolumn{4}{c}{\textbf{2-Undergrad}}
        & \multicolumn{4}{c}{\textbf{3-Easy Grad}}
        & \multicolumn{4}{c}{\textbf{4-Grad}}
        & \multicolumn{4}{c}{\textbf{5-Research}} \\
        \cmidrule(lr){2-5} \cmidrule(lr){6-9} \cmidrule(lr){10-13} \cmidrule(lr){14-17} \cmidrule(lr){18-21}
        & A & B & C & D 
        & A & B & C & D
        & A & B & C & D
        & A & B & C & D
        & A & B & C & D \\
        \midrule
        GPT-4o &28&0&11&1&50&6&8&1&20&4&29&2&8&5&51&6&1&4&50&0 \\
        o1 (high)&36&0&4&0&60&5&0&0&48&3&4&0&41&4&22&3&23&9&23&0 \\
        o3-mini (high)  &39&0&1&0&60&5&0&0&49&3&3&0&51&1&18&0&36&3&16&0 \\
        DeepSeek-R1  &31&2&7&0&58&6&1&0&41&2&10&2&25&3&23&19&6&0&26&23  \\
        DeepSeek-V3  &26&0&12&2&50&6&9&0&15&7&29&4&11&6&47&6&0&0&49&6 \\
        Llama-3.1-8B &11&1&15&13&4&7&25&29&0&1&13&41&0&0&15&55&0&0&8&47 \\
        Llama-3.1-70B &19&0&17&4&31&1&29&4&5&6&36&8&2&3&50&15&0&0&44&11 \\
        Qwen2.5-7B  &3&2&24&11&22&4&25&14&3&1&22&29&0&1&34&35&0&0&32&23 \\
        Qwen2.5-72B &25&0&13&2&35&5&23&2&11&5&30&9&8&2&47&13&1&1&46&7 \\
        QwQ-32B &25&3&11&1&38&9&18&0&11&1&33&10&2&2&49&17&1&1&37&16 \\
        \bottomrule
    \end{tabular}
        \caption{Letter grade received for different models. Note: the number of attempts per each level equals 5 shots times the number of problems in the level (see table \ref{tab:difficulty_levels}).}
        \label{tab:python_verify_letter_grade}
\end{table}

\begin{figure}
    \centering
    \includegraphics[width=0.50\linewidth]{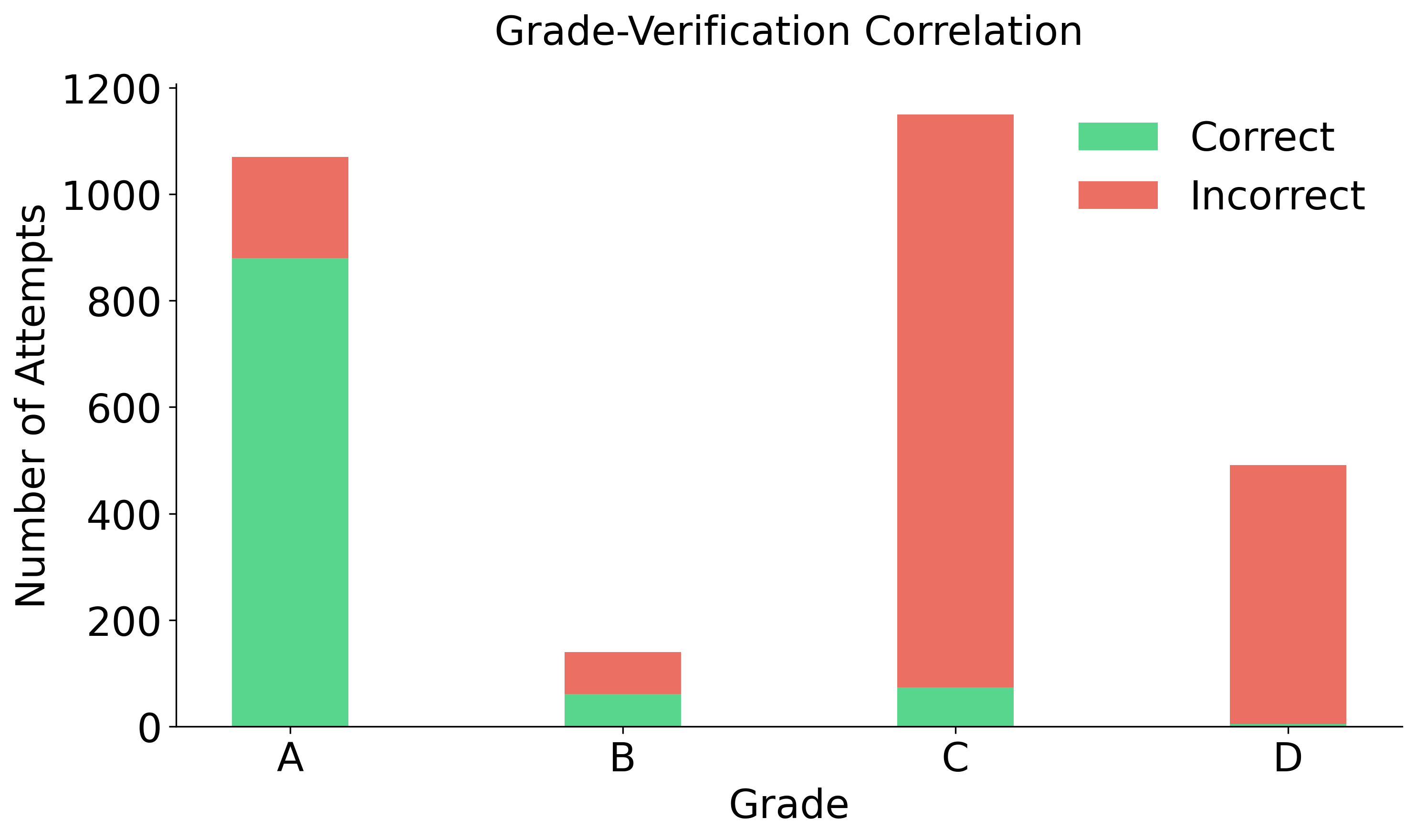}
    \caption{Stacked bar chart showing the number of solutions verified as correct (green) versus incorrect (red) across each letter grade.}
    \label{fig:grade_verification_bar}
\end{figure}

Table~\ref{tab:grade_verification} and the corresponding bar chart in Figure~\ref{fig:grade_verification_bar} summarize how the automatically verified results (\emph{Correct} vs. \emph{Incorrect}) align with the letter grades (\emph{A}, \emph{B}, \emph{C}, \emph{D}) assigned by the AI-based holistic grading. For \emph{A}-graded solutions, a large fraction (80.1\%) aligns with the auto-grader’s correct verification. By contrast, \emph{B}- and \emph{C}-graded solutions show substantially lower correctness rates (16.3\% and 4.9\% respectively). In the \emph{D} category, an overwhelming 99.5\% fail the auto-grader’s check, indicating that both holistic assessment and numeric verification typically reject these solutions.

Overall, there is a strong correlation between higher letter grades and positive verification outcomes, which validates that the AI-based grading system’s assessed quality generally corresponds to the auto-grader’s numeric correctness checks. At the same time, deviations exist in each category. For example, nearly 20\% of \emph{A}-graded solutions fail the numeric check, often because the AI holistic grader failed to correctly determine whether two answer expressions are equivalent. This is typically due to the expressions being overly complex. Conversely, a small fraction of lower-graded (\emph{C} or \emph{D}) responses may be mathematically correct in final form, yet insufficiently justified in intermediate steps, causing the holistic grader to assign a low grade despite correct numerical output.

Our findings illustrate that automatic verification and holistic AI-based grading are generally consistent: higher-quality solutions are confirmed as correct more frequently, while lower-quality solutions often fail numeric checks. 
Our current GPT-4o grader however has significant shortcomings. By cross-checking the grader with human grading, we find that LLM grading works reasonably well when grading solutions of low difficulty 1 to 2, but is not reliable at level 4 or 5. It seems likely that 4o is not strong enough to understand the logic of these higher difficulty problem solutions. In the present work, we thus focus on the auto-verifier results, and leave detailed exploration of holistic grading to future work.

\begin{table}[h]
\footnotesize
    \centering
    \renewcommand{\arraystretch}{1.2}
    \begin{tabular}{l c c c}
        \toprule
        Grade & Correct & Incorrect & Total \\
        \midrule
A & 880 (82.2\%) & 190 (17.8\%) & 1070 \\
B & 61 (43.6\%) & 79 (56.4\%)  & 140 \\
C & 74 (6.4\%) & 1075 (93.6\%) & 1149 \\
D & 5 (1.0\%) & 486 (99.0\%)  & 491 \\
        \midrule
\textbf{Total} & 972 (34.1\%) & 1878 (65.9\%)& 2850 \\
        \bottomrule
    \end{tabular}
    \caption{Grade Verification Results. Percentages in parentheses indicate the distribution of verification outcomes within each grade category. 
    Note: The total number 2850 results from 5 attempts for each of the 57 problems in the data set across 10 models.}
    \label{tab:grade_verification}
\end{table}

\subsection{Augmenting Inference With Python to Reduce Algebraic Mistakes}
\label{sec:pythonhelp}

We experimented with instructing models to break down calculations into smaller steps and verify these with python. Using a code interpreter was previously found to be beneficial in reducing algebraic mistakes in calculations (eg. \cite{kumar2023chatgpt4codeinterpreterused}) Our approach was based on the \texttt{MathChat} \cite{wu2023empirical} framework and prompt tuning. We instructed the model to write python (particularly \texttt{SymPy}) code for each calculation step and verify its result using this code. In a few cases, for low difficulty problems, our approach was able to spot and correct mistakes. However, more often the approach disrupted the reasoning chain and led to worse results. For complicated problems, models struggled to identify steps that can be checked with \texttt{SymPy}. We note that our problems do not include floating point calculations, where verification would be straightforward, but require more complicated algebraic operations. Recently, the FrontierMath paper \cite{glazer2024frontiermath} included a set of prompts to encourage LLMs to verify with python, but noted that advanced models barely made use of this possibility. While human theorists do sometimes check their results with computer algebra systems, especially \texttt{Mathematica}, this process is not straightforward, and there is likely limited existing training data for this approach. We aim to experiment with few-shot inference or fine-tuning in the future, showing the model handcrafted examples of \texttt{SymPy} or \texttt{Mathematica} verification in the prompt. Since our current \texttt{MathChat}-based results are not stable we chose to defer this direction to future work.

\section{Failure Mode Analysis}
\label{sec:failuremode}
We now discuss common classes of mistakes. We present a few examples highlighting the various types of errors that the LLMs make while attempting to solve problems in theoretical physics. We broadly classify these errors into four classes as shown below. Our examples mostly draw from GPT-4o and o1 model results.

\subsection{Background Knowledge of the Model}

Background knowledge is a strength of LLMs. Problem authors were impressed by models' ability to recall relevant mathematical definitions that were not included in the problem but are known to practicing researchers. This ability makes it much easier in principle to solve problems than with a computer algebra system like Mathematica. For example, consider the level 5 cosmology problem from App. \ref{sec:prob_bias}:

\begin{tcolorbox}
\footnotesize
\textcolor{usercolor}{\textbf{User:}}\\
In cosmology, large-scale cosmological dark-matter halo fields are biased tracers of the underlying Gaussian matter density $\delta_m$. Assume we have a sample $\delta_m$. We simulate a halo number density field by taking $n(\mathbf{x}) = \bar{n}\max(0,1+b\delta_m(\mathbf{x}))$, where bare number density $\bar{n}$ and bare bias $b$ are specified constants. What is the bias of the sampled halo field? Derive an equation to evaluate the bias which depends on the bare bias and the variance in each pixel.
\end{tcolorbox}  
While well-defined for a cosmologist, the problem does not define the mathematical quantities in detail, and would be hard to interpret by a non-cosmologist. Advanced models correctly recalled the required definitions and generally set up the problem correctly. 

However, while LLMs generally recall key definitions of various sub-fields of theoretical physics, they frequently encounter difficulties in accurately recalling more detailed mathematical information, as illustrated by the following two examples.

In one of the solutions to an undergraduate QM problem, the QwQ model incorrectly retrieves information about the Clebsch-Gordan coefficients. Specifically, it claims
\begin{tcolorbox}
\footnotesize
    From standard tables or textbooks, the Clebsch-Gordan coefficients are:\[\left\langle 1\;m_{1}\;1\;m_{2}|j\;m\right\rangle\] For \(j=1\), \(m=-1\):\[\left|1\;-1\right\rangle =\sqrt{\frac{2}{3}}\left|1\;-1\;1\;0\right\rangle +\sqrt{\frac{1}{3}}\left|1\;0\;1\;-1\right\rangle.\] 
\end{tcolorbox}The correct value of these coefficients are \(\mp1/\sqrt{2}\) for \(\left|1\;-1\;1\;0\right\rangle\)  and \(\left|1\;0\;1\;-1\right\rangle\) states respectively.

In the following snippet, generated from a model answer, the GPT-4o model incorrectly identifies the standard eigenstates of a particle in a 1-D infinite potential well ($|x|\leq L/2$) from existing results\footnote{The correct set of eigenvalues are \[\psi_{n}(x)=\sqrt{\frac{2}{L}}\sin\left(\frac{n\pi x}{L}\right)\quad\mbox{for \ensuremath{n=2,4,6,....}}, \qquad \psi_{n}(x)=\sqrt{\frac{2}{L}}\cos\left(\frac{n \pi x}{L}\right)\quad\mbox{for \ensuremath{n=1,3,5,....}}\]}
\begin{tcolorbox} \footnotesize
    \[\psi_{n}(x)=\sqrt{\frac{2}{L}}\sin\left(\frac{n\pi x}{L}\right)\quad\mbox{for \ensuremath{n=1,2,3,....}}\]and\[\psi_{n}(x)=\sqrt{\frac{2}{L}}\cos\left(\frac{n\pi x}{L}\right)\quad\mbox{for \ensuremath{n=2,4,6,....}}\]
\end{tcolorbox}

\subsection{Algebraic Mistakes}

A major challenge for models is to perform correct algebraic calculations. Consider the following relatively easy math problem that appears as an individual step in one of our problem solutions.

\begin{tcolorbox}
\footnotesize

\textcolor{usercolor}{\textbf{User:}}\\
Determine the leading real term of the expression
\[
F(k) = -1+\left(\frac{59 a^2 k^2}{15}+i a k+1\right)^5 \exp \left(\frac{-1}{6} a k (85 a k+6 i)\right)
\]
for real $a,k\in \mathrm{R}$ and \( k \ll 1 \).\\

\textcolor{expertcolor}{\textbf{Expert Solution :}}\\
The correct series expansion up to leading real and imaginary terms is 
\begin{eqnarray}
   \lim_{k\ll1} F(k) \approx \frac{793 a^4 k^4}{180}+4 i a k.
\end{eqnarray}

\end{tcolorbox}

We attempted this problem multiple times with o1 and o3-mini. In most of its responses, the LLMs did not expand the exponential term beyond the second order in $k$, falsely assuming that the leading real term must be proportional to $k^2$. In its best attempt, it expands up to quartic order in $k$, but fails to accurately combine the various terms to compute the coefficient of $k^4$. This is a good example of the promise of combining with computer algebra systems. If we add the prompt \say{Write and execute \texttt{SymPy} code to evaluate the expression.} models can generate Python code that calculates the correct expression.  
We have therefore tried to encourage python usage as discussed in Sec. \ref{sec:pythonhelp}, however with limited initial success. 

Algebraic mistakes are numerous, and often occur even in very simple calculations. Models often simply forget mathematical terms in an expression from one calculation step to the next. For example, the following is an arithmetic evaluation by GPT-4o,  in which it spuriously drops a factor of the imaginary number \(i\): 
\begin{tcolorbox}
    \footnotesize
    \[\nabla^{2}\vec{E}=i\omega(\sigma-i\omega)\vec{E}=\omega(\sigma+i\omega)\vec{E}.\]
\end{tcolorbox}
Similar cases of forgotten \(i\) factors, minus signs or constants occur frequently in many problem solution attempts.

Mathematical identities are also often applied incorrectly. For example, in the following case GPT-4o fails to implement the vector triple product \( \left((\vec{a}\times\vec{b})\times\vec{c}=(\vec{a}\cdot\vec{c})\vec{b}-(\vec{b}\cdot\vec{c})\vec{a}\right)\) correctly and writes
\begin{tcolorbox}
    \footnotesize
    \[(\vec{E}\times\hat{z})\times\hat{z}=\left[(\vec{E}\times\hat{z})\cdot\hat{z}\right]\hat{z}-(\vec{E}\times\hat{z})(\hat{z}\cdot\hat{z}).\]
\end{tcolorbox}
More powerful reasoning models tend to make less frequent ``simple" math mistakes such as the following:
\begin{tcolorbox}
    \footnotesize
    \textbf{o3-mini:} After performing the \(\eta-\)integrals (using the standard \(i\epsilon-\)prescription so that
\[
\int_{-\infty}^0 d\eta\,e^{iK\eta}=\frac{1}{iK},\qquad
\int_{-\infty}^0 d\eta\,\eta\,e^{iK\eta}=-\frac{1}{K^2}\,,
\]
\end{tcolorbox}
\noindent where the second integral erroneously contains a negative sign. It would be interesting to compare model performance on a set of automatically generated simple calculations typical for theoretical physics.

\subsection{Logical Mistakes}

We frequently observed that LLMs struggle to accurately account for the validity and applicability of advanced mathematical concepts, such as incorrectly applying theorems, misinterpreting definitions, or failing to recognize the limitations of certain mathematical techniques. Consider the following mathematical problem, which we will use to discuss logical errors made by the oX series models in their attempted solution.

\begin{tcolorbox}
\footnotesize

\textcolor{usercolor}{\textbf{User:}}\\
By Taylor expanding the integrand, find a $b$ cubic polynomial approximation to the integral 
\[
I(b) = \int_{b}^{1} \left( \frac{\sqrt{\pi}}{x} - \frac{\pi \, \text{erfc}\left(\frac{1}{\sqrt{x}}\right) \exp\left(\frac{1}{x}\right)}{x^{3/2}+x^{5/2}} \right) dx,
\]
that achieves a $90\%$ or better accuracy when \( b \) lies in the interval  \( [0, 1] \). 
\newline
\textcolor{expertcolor}{\textbf{Expert Solution :}}\\
First, we note that the integrand remains finite in the limit \( x \to 0 \), as the singular term proportional to \( 1/x \) cancels out. To assess the validity of a series expansion, we estimate the radius of convergence near the boundary points \( x = 0 \) and \( x = 1 \). This analysis shows that the integrand is convergent within the interval \( [0,1] \) when expanded about \( x = 1 \). Consequently, we perform a Taylor expansion around \( x = 1 \) retaining terms up to $(x-1)^2$:

\[
\frac{\sqrt{\pi}}{x} - \frac{\pi \, \text{erfc}\left(\frac{1}{\sqrt{x}}\right) \exp\left(\frac{1}{x}\right)}{x^{3/2}+x^{5/2}}
\approx \left(\sqrt{\pi}-\frac{1}{2} e \pi \, \text{erfc}(1)\right)-\left(\frac{27 \sqrt{\pi }}{4}-\frac{63}{8} e \pi  \,\text{erfc}(1)\right) (x-1) + \left(\frac{21 \sqrt{\pi }}{8}-\frac{51}{16} e \pi  \,\text{erfc}(1)\right) (x^2-1).
\]

Integrating over the interval \( [b,1] \), we obtain the approximate solution:

\[
I(b) \approx \left(b^3-1\right) \left(\frac{17}{16} e \pi \, \text{erfc}(1)-\frac{7 \sqrt{\pi }}{8}\right)+\left(b^2-1\right) \left(\frac{27 \sqrt{\pi }}{8}-\frac{63}{16} e \pi \, \text{erfc}(1)\right)+(b-1) \left(\frac{83}{16} e \pi \, \text{erfc}(1)-\frac{41 \sqrt{\pi }}{8}\right).
\]

In the limit \( b \to 0 \), our approximate expression evaluates to
\[
I(0) \approx 1.54633,
\]

which achieves approximately \( 93.6\% \) accuracy compared to the numerical result \( 1.65221 \). We note that this integral cannot be evaluated exactly using Mathematica or Maple software.

\end{tcolorbox}When the above problem was given to the o1 and o3-mini models, they demonstrated the following logical errors in their reasoning:
\begin{enumerate} \item The model begins by identifying that the integrand is finite at \( x = 0 \). However, it fails to recognize that the radius of convergence for the integrand around \( x = 0 \) is \( 0 \). This oversight leads to an improper application of approximations beyond the valid domain. Subsequently, the model factorizes the integral as \[ I(b) = \int_{b}^{1} f(x) \, dx = \int_{0}^{1} f(x) \, dx - \int_{0}^{b} f(x) \, dx. \] Assuming \( b \ll 1 \), the model proceeds to perform a Taylor expansion of the integrand around \( x = 0 \) and evaluates the second integral \( \int_{0}^{b} f(x) \, dx \) up to cubic order in \( b \). However, the Taylor expansion is only valid within the radius of convergence, and this restriction is not respected, rendering the approximation potentially invalid. 
\item To estimate the constant value of the first integral \( \int_{0}^{1} f(x) \, dx \), the model imposes the boundary condition at \( b = 1 \), equating \[ \int_{0}^{1} f(x) \, dx = \lim_{b \to 1} \int_{0}^{b} f(x) \, dx. \] The model substitutes the cubic-order Taylor expansion solution for \( \int_{0}^{b} f(x) \, dx \) derived in the previous step into the right-hand side (RHS) of this equation. This substitution constitutes a significant logical error in its reasoning, as the cubic-order approximation was determined only for \( b \ll 1 \), a fact that the model seemed to know but failed to implement. Extending this local approximation to \( b = 1 \), far beyond its domain of validity, leads to an erroneous evaluation of \( \int_{0}^{1} f(x) \, dx \). \end{enumerate}
Interestingly, the o3-mini model demonstrates two critical flaws: it not only arrives at logically inconsistent conclusions but occasionally also confidently hallucinates the claim $I(0)=\pi/2$, failing to furnish a coherent proof despite repeated prompting.

In a different problem involving particle physics, the GPT-4o model was asked to determine the effective mass of a spin 1/2 particle with action \[S=\int d^{4}x\bar{\psi}\left(ia\gamma^{\mu}\partial_{\mu}-c-i\frac{b}{\sqrt{3}}\gamma^{5}\right)\psi.\]The models did not understand and failed to reason out that the parameter \(c\) alone does not define the physical mass. The pseudoscalar \(\gamma^{5}\)-term must be included, corresponding to a chiral contribution to the mass.

For a much more basic example of failed logic, consider an example from QwQ. In one of our undergraduate Electrodynamics problems it produced the following expression followed by a faulty and rather incomplete reasoning:
\begin{tcolorbox}
    \footnotesize
   \[ (\vec{E}\times\hat{z})=b\vec{E}.\]
   But \(\vec{E}\times\hat{z}\) is perpendicular to both \(\vec{E}\) and \(\hat{z}\), which suggests that \(\vec{E}\) must be perpendicular to \(\hat{z}\) for this equation to hold.
\end{tcolorbox}
We found that advanced reasoning models such as o3-mini generally don't make such easy mistakes on undergraduate level physics problems. However, for difficult problems, they often oversimplify the problem due to a lack of detailed understanding. 
For instance, while solving the Level-5 problem detailed in App.~\ref{app:onepole}, o3-mini and other advanced reasoning models approximate scale factor $a(\eta)$ by expanding it linearly around the transition point, $\eta_e$, not realizing that the pole is far from the transition point and thus one needs to apply $a(\eta)\sim\eta^2$. For further details, we refer the readers to the expert solution detailed in App.~\ref{app:onepole}.

\subsection{Hallucinations}

Lastly, we present two instances where the LLM models generated new rules to obtain solutions that match with existing results in the literature. The following expression generated by GPT-4o represents an arithmetical hallucination error:
\begin{tcolorbox}
\footnotesize
\[k=\sqrt{\omega\sigma}\approx\sqrt{\omega\sigma}\left(\frac{1}{\sqrt{2}}+i\frac{1}{\sqrt{2}}\right).\]    
\end{tcolorbox}
The model performed the above arithmetic steps since it needed to determine the imaginary component of $k$. With this goal in mind, it carried out the above “illogical” mathematical step inventing new arithmetic rules to justify its approach to obtaining the imaginary component from a wrong answer.

Another example, from our problems, of o1 hallucinating non-existent rules to justify its approach is the following excerpt from its solution:
\begin{tcolorbox}
\footnotesize
    We can write
    \[T={\rm Tr}\left[\gamma^{\nu}\not p_{1}\gamma^{\mu}\not p_{2}\left(1+\gamma_{5}\right)\left(1-\gamma_{5}\right)\right].\]Note that \[\left(1+\gamma_{5}\right)\left(1-\gamma_{5}\right)=0.\]Therefore, to avoid vanishing of the trace, we need to consider that the \(\gamma_{5}\) matrices need to be kept separate. Instead, we should expand the trace without combining the projectors.
\end{tcolorbox}There is no such mathematical rule that an apparent zero can (must) be avoided by separating the terms and adding them later. The LLM invented this \say{rule} since it was working with incorrect expressions to nevertheless arrive at a correct solution, which in this case it was able to guess or recall (in only one of several attempts).

\subsection{Performance of Pre-O-Series Models}

In our experience, models that are not explicitly trained for reasoning (i.e. before the oX series) can be used to assist researchers that reason through a simple problem, but with significant shortcomings. Consider the following easy mathematical subproblem that appeared in one of our recent works  \cite{kvasiuk2024talefieldsneuralnetworkenhanced} in the context of cosmology, which we show here in simplified notation. 

\begin{tcolorbox}
\footnotesize

\textcolor{usercolor}{\textbf{User:}} \\
Assume $\mathbf{a}$, $\mathbf{b}$, and $\mathbf{c}$ are vectors, and $\mathbf{N}$ is a symmetric positive-definite matrix. Let $x$ and $y$ be real numbers. I want to minimize $\mathbf{a}^\top \mathbf{N} \mathbf{a}$ under the constraints $\mathbf{a}^\top \mathbf{b} = x$ and $\mathbf{a}^\top \mathbf{c} = y$. Solve this for $\mathbf{a}$, if possible. \\

\textcolor{expertcolor}{\textbf{Expert Solution:}}\\

We minimize $\mathbf{a}^\top \mathbf{N} \mathbf{a}$ subject to the constraints $\mathbf{a}^\top \mathbf{b} = x$ and $ \mathbf{a}^\top \mathbf{c} = y$. The Lagrangian \( \mathcal{L} \) for this optimization problem is defined as:
\[
\mathcal{L}(\mathbf{a}, \lambda, \mu) = \mathbf{a}^T N \mathbf{a} + \lambda (\mathbf{a}^T \mathbf{b} - x) + \mu (\mathbf{a}^T \mathbf{c} - y)
\]
Taking the gradient of \( \mathcal{L} \) with respect to \( \mathbf{a} \) and setting it to zero yields:
\[
\nabla_{\mathbf{a}} \mathcal{L} = 2N\mathbf{a} + \lambda \mathbf{b} + \mu \mathbf{c} = 0
\]
which we can solve for \( \mathbf{a} \) as:
\begin{align}
\label{eq:aweight}
\mathbf{a} = -\frac{1}{2} N^{-1} (\lambda \mathbf{b} + \mu \mathbf{c})
\end{align}
The constraint equations are:
\[
\mathbf{a}^T \mathbf{b} = x \quad \text{and} \quad \mathbf{a}^T \mathbf{c} = y
\]
Plugging the solution for \( \mathbf{a} \) into the constraint equations gives
\[
\begin{bmatrix}
\mathbf{b}^T N^{-1} \mathbf{b} & \mathbf{b}^T N^{-1} \mathbf{c} \\
\mathbf{c}^T N^{-1} \mathbf{b} & \mathbf{c}^T N^{-1} \mathbf{c}
\end{bmatrix}
\begin{bmatrix}
\lambda \\
\mu
\end{bmatrix}
=
\begin{bmatrix}
-2x \\
-2y
\end{bmatrix}
\]
which is of form 
\begin{align}
    \mathbf{M} \begin{bmatrix}
\lambda \\
\mu
\end{bmatrix} = -2
\begin{bmatrix}
x \\
y
\end{bmatrix} \nonumber
\end{align}
The above linear system is solved by (assuming the inverse exists, e.g., the two bias vectors are not co-linear):
\[
\begin{bmatrix}
\lambda \\
\mu
\end{bmatrix}
=
\frac{1}{\text{det}(M)}
\begin{bmatrix}
\mathbf{c}^T N^{-1} \mathbf{c} & -\mathbf{b}^T N^{-1} \mathbf{c} \\
-\mathbf{c}^T N^{-1} \mathbf{b} & \mathbf{b}^T N^{-1} \mathbf{b}
\end{bmatrix}
\begin{bmatrix}
-2x \\
-2y
\end{bmatrix}
\]
We then substitute the solution for $\lambda$ and $\mu$ back into \( \mathbf{a} \) using Eq. \eqref{eq:aweight}.

\end{tcolorbox}

That is a typical problem that GPT-4o and Llama-3 generally solve correctly, with correct mathematical derivation, although sometimes with a wrong numerical factor. It seems certain that this problem was in the training data of the model. Nevertheless, it is already time-saving for researchers to get answers to similar problems without manual labor. In particular for matrix algebra problems, existing computer algebra systems are not very strong or user friendly in our experience. However, the fact that models are very error prone limits their usefulness significantly. If every step needs to be checked in detail, the time saving can be minimal, or a wrong result can even confuse the user. Of course, human solutions can also have this property, depending on the skill and carefulness of the researcher.

\subsection{Performance of o1, o3-mini, and DeepSeek Reasoning Models}
\label{sec:omodelperformance}

From evaluating the output solutions generated by advanced LLM models such as o1, o3-mini and DeepSeek (DS), we observe that these models exhibit significantly stronger reasoning capabilities compared to other LLMs tested in our study. Notably, these models can perform more difficult algebraic manipulations, identify different components of a problem, and connect them with established concepts in the literature. This ability allows them to make meaningful progress in research-level problems, including those from topics such as quantum field theory (QFT) and String theory, by pinpointing key aspects of the question and recalling relevant background knowledge.

However, these models still struggle with detailed and systematic logical reasoning. When tasked with solving our Level-4 and Level-5 problems, these models often perform well in the initial phase of problem solving, demonstrating promising insights. Yet, for problems requiring extensive calculations combined with step-by-step logical rigor (e.g. loop integrals in QFT, tensor manipulations in general relativity) and systematic justfication of the assumptions, their performance deteriorates significantly. Our analysis of multiple solutions suggests that when intermediate steps become too complex, the models (including DS) often resort to literature memory from pre-training rather than performing detailed calculations. Rather than explicitly detailing intermediate steps, the models often present only their final answer, recalling related literature knowledge without references or resorting to vague assertions such as \say{after a lengthy (but straightforward) calculation} or \say{a short calculation shows}. While the full CoT of the o-series models is not public, we have no evidence that the models genuinely perform relevant calculations internally in these cases.

As an illustrative example, when asked to compute the one loop anomalous magnetic moment of a fermion (e.g. \cite{Peskin:1995ev}) including a contribution from a heavy scalar coupling, the model resorted to recalling existing solutions seen during pre-training rather than explicitly solving. However, it failed to recognize that the Yukawa interaction Lagrangian provided in our problem statement contained an additional factor of $1/\sqrt{2}$, which may deviate from the conventions in the literature. Consequently, its final answer overlooked this crucial modification. In a similar manner, when presented with the task of solving the Level-4 problem in \ref{L4-susy}, all advanced models (oX, DS-R1) initiate their response by articulating their interpretation of the problem statement and correctly identifying its connection to the standard supersymmetric transformations within the free Wess-Zumino model, as extensively documented in the literature. Subsequently, these models produce their final solution from memory. However, a consistent error emerges across all responses: the absence of the critical \say{negative sign} as seen in the solution given in Eq.\,(\ref{eq:L4-susy}). 
\begin{tcolorbox}
\footnotesize
    \textbf{o3-mini:} The well-known and consistent choice is
\[
\delta_{\eta}\phi = \sqrt{2}\,\eta^\alpha\,\xi_\alpha,
\]
with the Hermitian conjugate
\[
(\delta_{\eta}\phi)^\dagger = \sqrt{2}\,\bar{\eta}_{\dot{\alpha}}\,\bar{\xi}^{\dot{\alpha}}.
\]
This is verified by checking that the variations of all terms in \(\mathcal{L}\) under the full set of SUSY transformations (including the ones for \(\xi\) and \(F\)) cancel (up to a total derivative).
\end{tcolorbox}
While this might appear to be a minor discrepancy, it originates from a fundamental aspect of the problem. Specifically, the sign convention utilized in our given problem statement likely differed from the convention commonly adopted in the literature (for instance refer to Sec.~5.2 in \cite{Bailin1994}) used within the models' training samples, thereby necessitating a corresponding modification in the final transformation rule. This seemingly subtle yet conceptually significant detail indicates a potential cognitive limitation in these AI models, reflecting an over-reliance on memorized patterns rather than a systematic, first-principles approach, as well as a failure to validate the appropriateness of the retrieved solution within the context of the specific problem statement.

As another example of literature memory, one of the problems in TPBench involves solving a nonlinear differential equation in a manner similar to how Chandrasekhar presents the Kerr solution in \cite{Chandrasekhar:1985kt}. The number of steps to reach the answer is long and complicated. Such (complicated) recall problems are expected to be solvable by an AI due to its vast knowledge of the literature. Indeed, on one of the attempts, the AI can recognize the literature and write an answer to this problem, but even in that instance, it does not reason through the problem but just states: 
\begin{tcolorbox}
\footnotesize
    \textbf{o3-mini:} In fact, after a (lengthy) calculation one finds that the only solution (consistent with the field equations and the asymptotic condition) is
\[
e^X=\frac{r^2+C_2\mu^2}{\sqrt{\Delta(r)}}.
\]
\end{tcolorbox}

These inconsistencies suggest that models' solutions often fluctuate based on how their internal sampling mechanism recalls (pre-)training data, rather than adhering to a logically coherent problem-solving strategy. This underscores a fundamental issue: unlike a proficient researcher who would maintain logical consistency across different attempts, these models exhibit uncertainty in their outputs, lacking a clear measure of confidence in their solutions. Such limitations and the opaque structure of the training and inference process (especially of closed-source models) present obstacles to their applicability in research settings. It appears that successfully solved high-difficulty problems often benefit from the very deep and interconnected literature memory of these models, in addition with their ability to translate this knowledge to the problem setting. While this ability is useful for research, it may not be sufficient to create novel TP results without human assistance. In summary, current model performance perhaps resembles a student with superhuman literature knowledge but low intellectual rigor and technical expertise.

\section{Related work}
\label{sec:relatedwork}

Despite significant advances in the mathematical reasoning capabilities of large language models, accurately solving reasoning problems in specialized domains, such as theoretical physics remains a persistent challenge. In math reasoning, the landscape of existing benchmarks has been instrumental for the evaluation of LLM reasoning capabilities and the development of more robust and interpretable reasoning strategies. We review related benchmarks in Sec.~\ref{sec:related_benchmarks} as well as common strategies for eliciting more accurate reasoning from LLMs in Sec.~\ref{sec:related_methods}.

\subsection{Mathematical Reasoning Benchmarks}
\label{sec:related_benchmarks}
Recent progress in large language models (LLMs) has enabled these models to tackle increasingly complex tasks that demand high-level abstract mathematical reasoning. A significant body of work has focused on datasets for mathematical reasoning at the middle-school (e.g., \citep{cobbe2021training}), high-school (e.g., \citep{hendrycks2021measuring}), or undergraduate level (e.g., \citep{ling2017program}), which often cover arithmetic, geometry, or math word problems. Other benchmarks are focused on theorem proving \citep{tsoukalas2024putnambenchevaluatingneuraltheoremprovers,zheng2022minif2fcrosssystembenchmarkformal,liu2023fimochallengeformaldataset}. For example, the recently introduced \texttt{PutnamBench} \cite{tsoukalas2024putnambenchevaluatingneuraltheoremprovers} provides a collection of formalized theorems from the Putnam competition, while \texttt{MiniF2F} \cite{zheng2022minif2fcrosssystembenchmarkformal} and \texttt{FIMO} \cite{liu2023fimochallengeformaldataset} offer datasets of formalized proof problems drawn from competitions like the IMO, AIME, and AMC. In addition, \texttt{ProofNet} \cite{azerbayev2023proofnetautoformalizingformallyproving} comprises both natural language and formalized theorem statements and proofs at the undergraduate level. Complementary to these are natural language datasets that feature problems of varying difficulty \cite{hendrycks2021measuring, cobbe2021trainingverifierssolvemath}, as well as benchmarks like GPQA diamond \citep{rein2023gpqa}, which are designed to be hard.
Even more recently, the \texttt{Humanity's Last Exam} dataset \cite{phan2025humanity} (HLE) is an industry-curated, multi-domain benchmark that includes very challenging problems, among them some from theoretical physics. However, problems in HLE are constrained to numerical answers or multiple choice formats, there is no spectrum of difficulty, and it is not specifically designed to probe reasoning capabilities in theoretical physics.

While lower difficulty math benchmarks such as \texttt{MATH} \cite{hendrycks2021measuring} have nearly been mastered by current LLMs, the \texttt{FrontierMath} ~\cite{glazer2024frontiermath} dataset, which includes research-level problems curated by working mathematicians, remain largely unsolved. \texttt{FrontierMath} spans a range of difficulties from high-school to research level and features properties like auto-verifiability and rich metadata, design principles we have also incorporated into TPBench. However, \citet{glazer2024frontiermath} provide limited information about the difficulty distribution and the specifics of the problems that have been solved by advanced models.

In the realm of physics, which also demands extensive abstract mathematical reasoning, the focus has been predominantly on high-school level challenges as seen in datasets such as \texttt{JEEBench} \cite{arora2023llmsadvancedenoughchallenging}, \texttt{OlympiadBench} \cite{he2024olympiadbenchchallengingbenchmarkpromoting}, and \texttt{PhysicsQA} \cite{jaiswal2024improvingphysicsreasoninglarge}. Beyond undergraduate-level problems, very little work has addressed mathematical reasoning for theoretical physics. One notable exception is \cite{pan2024quantummanybodyphysicscalculations}, which examines symbolic calculations, albeit within the narrow context of a specific class of quantum many-body physics problems.

Our new dataset, TPBench addresses the gap in theoretical physics reasoning benchmarks beyond the undergraduate level. TPBench encompasses problems ranging from undergraduate to research level, with research problems reflecting challenges typical of those found in theoretical physics publications (rather than representing entire publications in themselves). 
Importantly, TPBench is designed to be independent of industry control, ensuring that the theoretical physics research community has access to a reasoning benchmark that is not susceptible to data leakage from future training data. We look forward to sharing this dataset with collaborators under appropriate data leakage controls.

\subsection{Reasoning Capabilities of LLMs}
\label{sec:related_methods}
Despite the remarkable fluency of LLMs in generating human-like text, their capacity to perform reliable multi-step reasoning remains a challenge~\citep{mirzadeh2024gsm}. Many LLMs still struggle with complex arithmetic and logical inference tasks. In this section, we review state-of-the-art methods, spanning both training-time and inference-time techniques that have been developed to boost the reasoning capabilities of LLMs.

\paragraph{Training-Time Methods for Improved Reasoning}
Training-time methods encompass all strategies where pre-trained language models are fine-tuned or otherwise modified to improve their reasoning capabilities. The most popular approaches in this category rely on either supervised fine tuning \citep{team2025kimi,xu2025redstardoesscalinglongcot,bespoke_stratos}, or reinforcement learning \citep{guo2025deepseek,team2025kimi} (or both \citep{guo2025deepseek}).
In supervised fine-tuning \citep{yu2023metamath,zelikman2022star,zelikman2024quiet, shao2024deepseekmathpushinglimitsmathematical}, high-quality reasoning chains are curated and used to fine-tune models to display more accurate reasoning behavior. \citet{chen2024self} demonstrate that self-play fine-tuning can improve model reasoning. 

\paragraph{Inference-Time Methods for Improved Reasoning}
Test-time methods aim at improving reasoning capabilities by either designing prompts that elicit good reasoning behavior or by building reasoning systems which prompt the LLM over and over to arrive at a solution in a systematic way. The most popular strategy for prompting large language models to reason is Chain-of-Thought \citep{wei2023chainofthoughtpromptingelicitsreasoning}, where the prompt includes instructions to ``think step-by-step". This is a type of test-time approach \citep{snell2024scalingllmtesttimecompute,welleck2024decodingmetagenerationinferencetimealgorithms,muennighoff2025s1}, as it typically leads to longer token sequences generated by the LLM. The default prompt (see App. \ref{sec:llmprompt}) we use to evaluate various LLMs on TPBench is a customized variation of chain-of-though -- it includes the tips from Polya's famous manual \say{How to solve it} \citep{Polya1945} which was originally intended to teach students how to solve mathematical problems. Related advances include prompting the model to break down the problem into simpler subproblems \citep{khot2022decomposed,zhou2022least,hao2023reasoning}, or seeking abstractions \citep{zheng2023take}. Other prompting strategies encourage models to self-verify \citep{lightman2023letsverifystepstep,selfevaluation}, self-improve \citep{chen2024self, chen2024boosting}, or iteratively refine their answer~\citep{selfrefine,selfrefinereport}.

Other strategies to elicit reasoning behavior involve the generation of multiple reasoning chains which can then be sampled from (as in best-of-$n$ \cite{beirami2025theoreticalguaranteesbestofnalignment}) or combined via majority voting or by ensuring self-consistency \citep{selfconsistency}. Methods that improve reasoning through planning \citep{hao2023reasoning,yao2023tree, qi2024mutual,zhang2024llamaberrypairwiseoptimizationo1like,kang2024mindstar} roll out multiple reasoning chains hierarchically and explore the space with Monte-Carlo Tree Search \citep{kocsis2006bandit}. The success of these methods depends on how the different reasoning chains are evaluated and can be achieved either through other language models \citep{qi2024mutual} or through external tools, e.g. \citep{zhang2024llamaberrypairwiseoptimizationo1like}. Tool usage in reasoning is explored next.

\paragraph{Verifiers and Tool Usage.}
Another avenue for boosting the performance of LLMs is by allowing tool usage \citep{schick2024toolformer, saad2024archon} either during the reasoning phase \citep{chen2022program}, or to verify intermediate reasoning steps \citep{zhang2024llamaberrypairwiseoptimizationo1like} and solutions \citep{imani2023mathprompter}. Verifiers and tools are compatible both with training-time and test-time methods. Since each of the problems in TPBench has an auto-verifier, one could consider giving the LLM under evaluation access to the auto-verifier to test if, by using it, it can achieve better results.

\section{Discussion} 
\label{sec:discussion}
We developed the dataset TPBench to test TP reasoning capabilities of AI models. We note that our problems were not constructed to match a particular target error rate (o3-mini and DeepSeek R1 appeared after most problems were finalized), but rather to reflect real problems encountered by theoretical physicists at each career level. Our theoretical physics reasoning results are consistent with studies from more general benchmarks, and illustrate the speed of progress in AI. The most advanced models are able to solve some problems at graduate level, but are not yet capable of solving most research level problems. While advanced models demonstrate remarkable proficiency in algebraic and conceptual problem-solving, they struggle with structured logical reasoning and transparent step-by-step calculations, particularly in complex, research-level problems. Their reliance on literature recall without verification or referencing and their lack of consistency in detailed reasoning remain key limitations in their problem-solving capabilities. We discussed these shortcomings and summarized common failure modes.

Progress has been rapid, even during the creation of this data set. If models could solve level 5 research problems consistently, their impact on theoretical physics would be substantial. However, even then, AI models could not perform independent research without further developments. We now discuss some future directions related to our work, that could make LLMs more powerful for TP research.

\paragraph{Updates to the TPBench data set and score board.} We will update the score board for novel SOTA models. Results will be published on the website of the data set \url{tpbench.org}. The website also contains additional model evaluation metrics, which assign a unified model score over all difficulties. We aim to add more problems to the data set in the future, both public and private problems. 
It would be particularly interesting to design more research problems which are clearly outside of the training data. This could be achieved by curating research problems specifically from the newest arxiv publications, before the current knowledge cutoff. We invite interested researchers to contribute new problems and collaborate on future TPBench updates (see website for details).

\paragraph{Automatic problem scraping from publication archives.} To improve inference methods specifically for TP, for example by reinforcement learning of reasoning chains (e.g. Deepeek R1 \cite{guo2025deepseek}), it would be important to have a large collection of verifiable problems. If problems could be extracted automatically from publications, perhaps after a training data cutoff, this would allow generation of training data without human labor at industrial scale. An initial exploration of LLM-based problem extraction from papers has revealed that this is difficult in TP because calculations are often spread over the paper and it is not clear to the model what information is needed to state the problem and what the answer is. This is more obvious in mathematical papers that clearly mark theorems and proofs (e.g. with latex tags), however those are more difficult to auto-verify. Nevertheless, this is an exciting direction for future work, especially since large industry labs keep their training data for reasoning models private.  

\paragraph{Automatic verification for non-algebraic expressions.} We were somewhat constrained in our choice of problems by the criterium of auto-verifiability. Many TP results can be written in inequivalent ways, and models are not currently good at judging equivalence of expressions. Large collections of verifiable problems are also important for reinforcement learning-based training of reasoning models, see e.g. the recent DeepSeek R1 \cite{guo2025deepseek}. Generating stronger verifiers that work for a wider class of problems is a very interesting direction for future work, where theoretical and computational physics domain expertise is valuable. Some challenges were listed in Sec. \ref{sec:autoverifier}. We note that results in TP are often symbolic expressions, which are more suited for auto-verification than mathematical proofs (which need to be checked by proof assistants).

\paragraph{Improving reasoning methods for TP.} We have reviewed methods to improve reasoning capabilities of LLMs in Sec \ref{sec:related_methods}.
It is clear from our experiments that a significant gain could be obtained if tools such as \texttt{SymPy} or \texttt{Mathematica} would be used consistently to check symbolic calculations where this is possible. Few shot learning or fine-tuning could be used to improve models ability to call symbolic software packages. However, many TP calculations require specific packages and do not come with a lot of training data.  
Further, the human TP research process involves reading publications, and looking up results or methods when needed. References are also used to spot mistakes in calculations by comparing to known published results where possible. While LLMs can parse literature with techniques such as RAG \cite{gao2023retrieval}, to our knowledge this has not been demonstrated to lead to performance gains in mathematical reasoning. The fact that models cannot point to a specific source for mathematical statements lowers their trustworthiness. 
Finally, inference methods that provide more information about uncertainty in individual steps would be particularly beneficial for difficult TP problems. This would pave the way for trustworthy, automated TP research assistants that reliably solve some aspects of a problem, but then ask for help for the parts they are uncertain about.

\paragraph{Diagrammatic and spatial reasoning.} Theoretical physicists like to reason using spatial diagrams such as Feynman diagrams or drawing integration contours. In principle, such diagrams can be encoded in some formal language and multi-modality for spatial reasoning may not be necessary. For example, some of our problem solutions include Feynman diagrams or integration contours encoded with the \texttt{TikZ} LaTeX library (e.g. Fig. \ref{fig:The-original-contour}). For some of our problems humans would have trouble reasoning through them without the ability to draw on some scratchpad. It would be interesting to see whether multi-modal language and spatial reasoning models could make models stronger. Visualizing the problem (e.g. \say{running an example in your head}) is a common strategy and could be particularly powerful for models to develop truly novel ideas. 

\paragraph{Training reasoning models on TPBench.} While we designed TPBench for the evaluation of the reasoning capabilities of large language models, it would also be very interesting to curate a dataset for supervised fine-tuning or for reinforcement learning purposes. While we expect fine-tuning to increase the TP specific reasoning capabilities of LLMs, it is equally important to avoid data leakage to avoid problems that are later used for evaluation to seep into the training data. For this reason we choose not to publish all of our problems in TPBench at this time. Instead we encourage researchers who wish to have their models evaluated on TPBench to reach out to us.

\paragraph{Open-ended research problems.} If models could solve well-posed problems such as the research problems in our collection reliably, this would speed up TP research projects considerably. However, a large part of research consists of arriving at well-posed problems, which are interesting to answer and can be answered. It could be possible to design more open-ended tasks, where the goal is to \say{derive interesting results} based on some set of initial constraints or observations. The AI model could suggest assumptions to include or drop, design its own problem statements, and attempt to judge the importance of its results (develop \say{theoretical taste}). It would be exciting and challenging to set up such a more open-ended benchmark.

\paragraph{Community efforts by the TP community.} With reasoning models being developed primarily by industry, usually with proprietary and closed data sets, it is important to consider how the open research community can contribute to AI driven TP reasoning. It now seems possible that AI models will be able to do significant theoretical research within a few years. The TP community should work towards the goal that such research remains open and accessible, rather than being performed exclusively at a few select industry labs. While pre-training may be financially inaccessible to publicly funded research, supervised fine-tuning, reinforcement learning, and algorithm development require more moderate resources. As an example, the community could build data sets for both TP reasoning training and benchmarking that are available to both the community and AI labs (with some data leakage control). These could also include examples of tool usage such as Mathematica. A large community-curated data set of verifiable TP problems would in particular allow supervised fine-tuning and Reinforcement Learning specifically for TP. Our data set is a first step in that direction. We hope that this work will contribute to engaging theoretical physicists in this exciting research direction.

\section*{Acknowledgements}

We thank Kendrick Smith and Matthew Johnson for discussions. M.M. and D.J.H.C. acknowledge the support by the U.S. Department of Energy, Office of Science, Office of High Energy Physics under Award Number DE-SC0017647. M.M. also acknowledges the support by the National Science Foundation (NSF) under Grant Number 2307109 and
the Wisconsin Alumni Research Foundation (WARF). F.S. is grateful for the support of the NSF under CCF2106707 and
the Wisconsin Alumni Research Foundation (WARF).

\bibliography{bibliography}

\begin{thebibliography}{84}
\providecommand{\natexlab}[1]{#1}
\providecommand{\url}[1]{\texttt{#1}}
\expandafter\ifx\csname urlstyle\endcsname\relax
  \providecommand{\doi}[1]{doi: #1}\else
  \providecommand{\doi}{doi: \begingroup \urlstyle{rm}\Url}\fi

\bibitem[Hendrycks et~al.(2021)Hendrycks, Burns, Kadavath, Arora, Basart, Tang, Song, and Steinhardt]{hendrycks2021measuring}
Dan Hendrycks, Collin Burns, Saurav Kadavath, Akul Arora, Steven Basart, Eric Tang, Dawn Song, and Jacob Steinhardt.
\newblock Measuring mathematical problem solving with the math dataset.
\newblock \emph{arXiv preprint arXiv:2103.03874}, 2021.

\bibitem[Glazer et~al.(2024)Glazer, Erdil, Besiroglu, Chicharro, Chen, Gunning, Olsson, Denain, Ho, Santos, et~al.]{glazer2024frontiermath}
Elliot Glazer, Ege Erdil, Tamay Besiroglu, Diego Chicharro, Evan Chen, Alex Gunning, Caroline~Falkman Olsson, Jean-Stanislas Denain, Anson Ho, Emily de~Oliveira Santos, et~al.
\newblock Frontiermath: A benchmark for evaluating advanced mathematical reasoning in ai.
\newblock \emph{arXiv preprint arXiv:2411.04872}, 2024.

\bibitem[Arora et~al.(2023)Arora, Singh, and Mausam]{arora2023llmsadvancedenoughchallenging}
Daman Arora, Himanshu~Gaurav Singh, and Mausam.
\newblock Have llms advanced enough? a challenging problem solving benchmark for large language models, 2023.
\newblock URL \url{https://arxiv.org/abs/2305.15074}.

\bibitem[He et~al.(2024)He, Luo, Bai, Hu, Thai, Shen, Hu, Han, Huang, Zhang, Liu, Qi, Liu, and Sun]{he2024olympiadbenchchallengingbenchmarkpromoting}
Chaoqun He, Renjie Luo, Yuzhuo Bai, Shengding Hu, Zhen~Leng Thai, Junhao Shen, Jinyi Hu, Xu~Han, Yujie Huang, Yuxiang Zhang, Jie Liu, Lei Qi, Zhiyuan Liu, and Maosong Sun.
\newblock Olympiadbench: A challenging benchmark for promoting agi with olympiad-level bilingual multimodal scientific problems, 2024.
\newblock URL \url{https://arxiv.org/abs/2402.14008}.

\bibitem[Jaiswal et~al.(2024)Jaiswal, Jain, Popat, Anand, Dharmadhikari, Marathe, and Shah]{jaiswal2024improvingphysicsreasoninglarge}
Raj Jaiswal, Dhruv Jain, Harsh~Parimal Popat, Avinash Anand, Abhishek Dharmadhikari, Atharva Marathe, and Rajiv~Ratn Shah.
\newblock Improving physics reasoning in large language models using mixture of refinement agents, 2024.
\newblock URL \url{https://arxiv.org/abs/2412.00821}.

\bibitem[Pan et~al.(2024)Pan, Mudur, Taranto, Tikhanovskaya, Venugopalan, Bahri, Brenner, and Kim]{pan2024quantummanybodyphysicscalculations}
Haining Pan, Nayantara Mudur, Will Taranto, Maria Tikhanovskaya, Subhashini Venugopalan, Yasaman Bahri, Michael~P. Brenner, and Eun-Ah Kim.
\newblock Quantum many-body physics calculations with large language models, 2024.
\newblock URL \url{https://arxiv.org/abs/2403.03154}.

\bibitem[Phan et~al.(2025)Phan, Gatti, Han, Li, Hu, Zhang, Shi, Choi, Agrawal, Chopra, et~al.]{phan2025humanity}
Long Phan, Alice Gatti, Ziwen Han, Nathaniel Li, Josephina Hu, Hugh Zhang, Sean Shi, Michael Choi, Anish Agrawal, Arnav Chopra, et~al.
\newblock Humanity's last exam, 2025.
\newblock URL \url{https://arxiv.org/abs/2501.14249}.

\bibitem[Iyer and Wald(1994)]{Iyer:1994ys}
Vivek Iyer and Robert~M. Wald.
\newblock {Some properties of Noether charge and a proposal for dynamical black hole entropy}.
\newblock \emph{Phys. Rev. D}, 50:\penalty0 846--864, 1994.
\newblock \doi{10.1103/PhysRevD.50.846}.

\bibitem[Geroch(1968)]{Geroch:1968zm}
Robert~P. Geroch.
\newblock {Spinor structure of space-times in general relativity. i}.
\newblock \emph{J. Math. Phys.}, 9:\penalty0 1739--1744, 1968.
\newblock \doi{10.1063/1.1664507}.

\bibitem[Kontsevich(1992)]{Kontsevich:1992ti}
M.~Kontsevich.
\newblock {Intersection theory on the moduli space of curves and the matrix Airy function}.
\newblock \emph{Commun. Math. Phys.}, 147:\penalty0 1--23, 1992.
\newblock \doi{10.1007/BF02099526}.

\bibitem[Schon and Yau(1981)]{Schon:1981vd}
Richard Schon and Shing-Tung Yau.
\newblock {Proof of the positive mass theorem. 2.}
\newblock \emph{Commun. Math. Phys.}, 79:\penalty0 231--260, 1981.
\newblock \doi{10.1007/BF01942062}.

\bibitem[Aganagic et~al.(2024)Aganagic, Danilenko, Li, Shende, and Zhou]{Aganagic:2024sww}
Mina Aganagic, Ivan Danilenko, Yixuan Li, Vivek Shende, and Peng Zhou.
\newblock {Quiver Hecke algebras from Floer homology in Couloumb branches}.
\newblock \emph{arXiv preprint arXiv: 2406.04258}, 6 2024.

\bibitem[Parker and Taubes(1982)]{Parker:1981uy}
Thomas Parker and Clifford~Henry Taubes.
\newblock {On Witten's Proof of the Positive Energy Theorem}.
\newblock \emph{Commun. Math. Phys.}, 84:\penalty0 223, 1982.
\newblock \doi{10.1007/BF01208569}.

\bibitem[Hausel and Thaddeus(2003)]{Hausel:2002ap}
Tamas Hausel and Michael Thaddeus.
\newblock {Mirror symmetry, Langlands duality, and the Hitchin system}.
\newblock \emph{Invent. Math.}, 153:\penalty0 197, 2003.
\newblock \doi{10.1007/s00222-003-0286-7}.

\bibitem[Hardy(1940)]{Hardy1940}
G.~H. Hardy.
\newblock \emph{A Mathematician's Apology}.
\newblock Cambridge University Press, London, 1940.
\newblock URL \url{https://www.cambridge.org/core/books/mathematicians-apology/A344F9D097F5AFF45BDA21B57B54BDCA}.
\newblock Foreword by C. P. Snow.

\bibitem[Rota(1997)]{Rota1997}
Gian-Carlo Rota.
\newblock Ten lessons i wish i had been taught.
\newblock \emph{Notices of the American Mathematical Society}, 44\penalty0 (1):\penalty0 22--25, 1997.
\newblock URL \url{https://www.ams.org/notices/199701/comm-rota.pdf}.

\bibitem[Si et~al.(2024)Si, Yang, and Hashimoto]{si2024can}
Chenglei Si, Diyi Yang, and Tatsunori Hashimoto.
\newblock Can llms generate novel research ideas? a large-scale human study with 100+ nlp researchers.
\newblock \emph{arXiv preprint arXiv:2409.04109}, 2024.

\bibitem[Pólya(1945)]{Polya1945}
George Pólya.
\newblock \emph{How to Solve It: A New Aspect of Mathematical Method}.
\newblock Princeton University Press, Princeton, NJ, 1945.
\newblock ISBN 978-0-691-11966-3.

\bibitem[OpenAI(2024{\natexlab{a}})]{GPT-4o}
OpenAI.
\newblock gpt-4o, 2024{\natexlab{a}}.
\newblock URL \url{https://openai.com/index/hello-gpt-4o/}.

\bibitem[OpenAI(2024{\natexlab{b}})]{o1-preview}
OpenAI.
\newblock Introducing openai o1-preview, 2024{\natexlab{b}}.
\newblock URL \url{https://openai.com/index/introducing-openai-o1-preview/}.

\bibitem[OpenAI(2025)]{o3-mini}
OpenAI.
\newblock o3-mini, 2025.
\newblock URL \url{https://openai.com/index/openai-o3-mini/}.

\bibitem[Imani et~al.(2023)Imani, Du, and Shrivastava]{imani2023mathprompter}
Shima Imani, Liang Du, and Harsh Shrivastava.
\newblock Mathprompter: Mathematical reasoning using large language models.
\newblock \emph{arXiv preprint arXiv:2303.05398}, 2023.

\bibitem[Huang et~al.(2023)Huang, Chen, Mishra, Zheng, Yu, Song, and Zhou]{huang2023large}
Jie Huang, Xinyun Chen, Swaroop Mishra, Huaixiu~Steven Zheng, Adams~Wei Yu, Xinying Song, and Denny Zhou.
\newblock Large language models cannot self-correct reasoning yet.
\newblock \emph{arXiv preprint arXiv:2310.01798}, 2023.

\bibitem[Yin et~al.(2023)Yin, Sun, Guo, Wu, Qiu, and Huang]{yin2023large}
Zhangyue Yin, Qiushi Sun, Qipeng Guo, Jiawen Wu, Xipeng Qiu, and Xuanjing Huang.
\newblock Do large language models know what they don't know?
\newblock \emph{arXiv preprint arXiv:2305.18153}, 2023.

\bibitem[Kamoi et~al.(2024)Kamoi, Zhang, Zhang, Han, and Zhang]{kamoi2024when}
Ryo Kamoi, Yusen Zhang, Nan Zhang, Jiawei Han, and Rui Zhang.
\newblock When can llms actually correct their own mistakes? a critical survey of self-correction of llms.
\newblock \emph{arXiv preprint arXiv:2406.01297}, 2024.

\bibitem[Dhuliawala et~al.(2023)Dhuliawala, Komeili, Xu, Raileanu, Li, Celikyilmaz, and Weston]{dhuliawala2023chain}
Shehzaad Dhuliawala, Mojtaba Komeili, Jing Xu, Roberta Raileanu, Xian Li, Asli Celikyilmaz, and Jason Weston.
\newblock Chain-of-verification reduces hallucination in large language models.
\newblock \emph{arXiv preprint arXiv:2309.11495}, 2023.

\bibitem[Zhang et~al.(2023)Zhang, Li, Das, Malin, and Kumar]{zhang2023sac3}
Jiaxin Zhang, Zhuohang Li, Kamalika Das, Bradley Malin, and Sricharan Kumar.
\newblock Sac3: Reliable hallucination detection in black-box language models via semantic-aware cross-check consistency.
\newblock \emph{arXiv preprint arXiv:2311.01740}, 2023.

\bibitem[Jiang et~al.(2024)Jiang, Peng, Feng, Li, and Li]{jiang2024llms}
Zhuoxuan Jiang, Haoyuan Peng, Shanshan Feng, Fan Li, and Dongsheng Li.
\newblock Llms can find mathematical reasoning mistakes by pedagogical chain-of-thought.
\newblock \emph{arXiv preprint arXiv:2405.06705}, 2024.

\bibitem[Kvasiuk et~al.(2024)Kvasiuk, Münchmeyer, and Smith]{kvasiuk2024talefieldsneuralnetworkenhanced}
Yurii Kvasiuk, Moritz Münchmeyer, and Kendrick Smith.
\newblock A tale of two fields: Neural network-enhanced non-gaussianity search with halos, 2024.
\newblock URL \url{https://arxiv.org/abs/2410.01007}.

\bibitem[Basso et~al.(2022)Basso, Chung, Kolb, and Long]{Basso:2022tpd}
Edward Basso, Daniel J.~H. Chung, Edward~W. Kolb, and Andrew~J. Long.
\newblock {Quantum interference in gravitational particle production}.
\newblock \emph{JHEP}, 12:\penalty0 108, 2022.
\newblock \doi{10.1007/JHEP12(2022)108}.

\bibitem[Ellis(2017)]{Ellis:2016jkw}
Joshua Ellis.
\newblock {TikZ-Feynman: Feynman diagrams with TikZ}.
\newblock \emph{Comput. Phys. Commun.}, 210:\penalty0 103--123, 2017.
\newblock \doi{10.1016/j.cpc.2016.08.019}.

\bibitem[AI(2024)]{Metallama3.1}
Meta AI.
\newblock Meta llama 3.1, 2024.
\newblock URL \url{https://ai.meta.com/blog/meta-llama-3-1/}.

\bibitem[Team(2024{\natexlab{a}})]{Qwen2.5}
Qwen Team.
\newblock Qwen2.5, 2024{\natexlab{a}}.
\newblock URL \url{https://qwenlm.github.io/blog/qwen2.5/}.

\bibitem[Team(2024{\natexlab{b}})]{QwenQwQ32b}
Qwen Team.
\newblock Qwen qwq 32b preview, 2024{\natexlab{b}}.
\newblock URL \url{https://qwenlm.github.io/blog/qwq-32b-preview/}.

\bibitem[Guo et~al.(2025)Guo, Yang, Zhang, Song, Zhang, Xu, Zhu, Ma, Wang, Bi, et~al.]{guo2025deepseek}
Daya Guo, Dejian Yang, Haowei Zhang, Junxiao Song, Ruoyu Zhang, Runxin Xu, Qihao Zhu, Shirong Ma, Peiyi Wang, Xiao Bi, et~al.
\newblock Deepseek-r1: Incentivizing reasoning capability in llms via reinforcement learning.
\newblock \emph{arXiv preprint arXiv:2501.12948}, 2025.

\bibitem[Bi et~al.(2024)Bi, Chen, Chen, Chen, Dai, Deng, Ding, Dong, Du, Fu, et~al.]{bi2024deepseek}
Xiao Bi, Deli Chen, Guanting Chen, Shanhuang Chen, Damai Dai, Chengqi Deng, Honghui Ding, Kai Dong, Qiushi Du, Zhe Fu, et~al.
\newblock Deepseek llm: Scaling open-source language models with longtermism.
\newblock \emph{arXiv preprint arXiv:2401.02954}, 2024.

\bibitem[Chen et~al.(2024{\natexlab{a}})Chen, Chen, Liu, Jiang, and Wang]{chen2024humansllmsjudgestudy}
Guiming~Hardy Chen, Shunian Chen, Ziche Liu, Feng Jiang, and Benyou Wang.
\newblock Humans or llms as the judge? a study on judgement biases, 2024{\natexlab{a}}.
\newblock URL \url{https://arxiv.org/abs/2402.10669}.

\bibitem[Kumar and Kats(2023)]{kumar2023chatgpt4codeinterpreterused}
Tanuj Kumar and Mikhail~A. Kats.
\newblock Chatgpt-4 with code interpreter can be used to solve introductory college-level vector calculus and electromagnetism problems, 2023.
\newblock URL \url{https://arxiv.org/abs/2309.08881}.

\bibitem[Wu et~al.(2023)Wu, Jia, Zhang, Li, Zhu, Wang, Lee, Peng, Wu, and Wang]{wu2023empirical}
Yiran Wu, Feiran Jia, Shaokun Zhang, Hangyu Li, Erkang Zhu, Yue Wang, Yin~Tat Lee, Richard Peng, Qingyun Wu, and Chi Wang.
\newblock Mathchat: Converse to tackle challenging math problems with llm agents.
\newblock \emph{arXiv preprint arXiv:2306.01337}, 2023.

\bibitem[Peskin and Schroeder(1995)]{Peskin:1995ev}
Michael~E. Peskin and Daniel~V. Schroeder.
\newblock \emph{{An Introduction to quantum field theory}}.
\newblock Addison-Wesley, Reading, USA, 1995.
\newblock ISBN 978-0-201-50397-5, 978-0-429-50355-9, 978-0-429-49417-8.
\newblock \doi{10.1201/9780429503559}.

\bibitem[Bailin and Love(1994)]{Bailin1994}
D.~Bailin and Alexander Love.
\newblock \emph{Supersymmetric Gauge Field Theory and String Theory}.
\newblock CRC Press, Boca Raton, 1994.
\newblock ISBN 978-0750302678.
\newblock \doi{10.1201/9780367805807}.
\newblock URL \url{https://library.oapen.org/handle/20.500.12657/50873}.

\bibitem[Chandrasekhar(1985)]{Chandrasekhar:1985kt}
Subrahmanyan Chandrasekhar.
\newblock \emph{{The mathematical theory of black holes}}.
\newblock 1985.
\newblock ISBN 978-0-19-850370-5.

\bibitem[Cobbe et~al.(2021{\natexlab{a}})Cobbe, Kosaraju, Bavarian, Chen, Jun, Kaiser, Plappert, Tworek, Hilton, Nakano, Hesse, and Schulman]{cobbe2021training}
Karl Cobbe, Vineet Kosaraju, Mohammad Bavarian, Mark Chen, Heewoo Jun, Lukasz Kaiser, Matthias Plappert, Jerry Tworek, Jacob Hilton, Reiichiro Nakano, Christopher Hesse, and John Schulman.
\newblock Training verifiers to solve math word problems, 2021{\natexlab{a}}.

\bibitem[Ling et~al.(2017)Ling, Yogatama, Dyer, and Blunsom]{ling2017program}
Wang Ling, Dani Yogatama, Chris Dyer, and Phil Blunsom.
\newblock Program induction by rationale generation: Learning to solve and explain algebraic word problems.
\newblock In \emph{Proceedings of the 55th Annual Meeting of the Association for Computational Linguistics (Volume 1: Long Papers)}, pages 158--167, 2017.

\bibitem[Tsoukalas et~al.(2024)Tsoukalas, Lee, Jennings, Xin, Ding, Jennings, Thakur, and Chaudhuri]{tsoukalas2024putnambenchevaluatingneuraltheoremprovers}
George Tsoukalas, Jasper Lee, John Jennings, Jimmy Xin, Michelle Ding, Michael Jennings, Amitayush Thakur, and Swarat Chaudhuri.
\newblock Putnambench: Evaluating neural theorem-provers on the putnam mathematical competition, 2024.
\newblock URL \url{https://arxiv.org/abs/2407.11214}.

\bibitem[Zheng et~al.(2022)Zheng, Han, and Polu]{zheng2022minif2fcrosssystembenchmarkformal}
Kunhao Zheng, Jesse~Michael Han, and Stanislas Polu.
\newblock Minif2f: a cross-system benchmark for formal olympiad-level mathematics, 2022.
\newblock URL \url{https://arxiv.org/abs/2109.00110}.

\bibitem[Liu et~al.(2023)Liu, Shen, Xin, Liu, Yuan, Wang, Ju, Zheng, Yin, Li, Zhang, and Liu]{liu2023fimochallengeformaldataset}
Chengwu Liu, Jianhao Shen, Huajian Xin, Zhengying Liu, Ye~Yuan, Haiming Wang, Wei Ju, Chuanyang Zheng, Yichun Yin, Lin Li, Ming Zhang, and Qun Liu.
\newblock Fimo: A challenge formal dataset for automated theorem proving, 2023.
\newblock URL \url{https://arxiv.org/abs/2309.04295}.

\bibitem[Azerbayev et~al.(2023)Azerbayev, Piotrowski, Schoelkopf, Ayers, Radev, and Avigad]{azerbayev2023proofnetautoformalizingformallyproving}
Zhangir Azerbayev, Bartosz Piotrowski, Hailey Schoelkopf, Edward~W. Ayers, Dragomir Radev, and Jeremy Avigad.
\newblock Proofnet: Autoformalizing and formally proving undergraduate-level mathematics, 2023.
\newblock URL \url{https://arxiv.org/abs/2302.12433}.

\bibitem[Cobbe et~al.(2021{\natexlab{b}})Cobbe, Kosaraju, Bavarian, Chen, Jun, Kaiser, Plappert, Tworek, Hilton, Nakano, Hesse, and Schulman]{cobbe2021trainingverifierssolvemath}
Karl Cobbe, Vineet Kosaraju, Mohammad Bavarian, Mark Chen, Heewoo Jun, Lukasz Kaiser, Matthias Plappert, Jerry Tworek, Jacob Hilton, Reiichiro Nakano, Christopher Hesse, and John Schulman.
\newblock Training verifiers to solve math word problems, 2021{\natexlab{b}}.
\newblock URL \url{https://arxiv.org/abs/2110.14168}.

\bibitem[Rein et~al.(2023)Rein, Hou, Stickland, Petty, Pang, Dirani, Michael, and Bowman]{rein2023gpqa}
David Rein, Betty~Li Hou, Asa~Cooper Stickland, Jackson Petty, Richard~Yuanzhe Pang, Julien Dirani, Julian Michael, and Samuel~R Bowman.
\newblock Gpqa: A graduate-level google-proof q\&a benchmark.
\newblock \emph{arXiv preprint arXiv:2311.12022}, 2023.

\bibitem[Mirzadeh et~al.(2024)Mirzadeh, Alizadeh, Shahrokhi, Tuzel, Bengio, and Farajtabar]{mirzadeh2024gsm}
Iman Mirzadeh, Keivan Alizadeh, Hooman Shahrokhi, Oncel Tuzel, Samy Bengio, and Mehrdad Farajtabar.
\newblock Gsm-symbolic: Understanding the limitations of mathematical reasoning in large language models.
\newblock \emph{arXiv preprint arXiv:2410.05229}, 2024.

\bibitem[Team et~al.(2025)Team, Du, Gao, Xing, Jiang, Chen, Li, Xiao, Du, Liao, et~al.]{team2025kimi}
Kimi Team, Angang Du, Bofei Gao, Bowei Xing, Changjiu Jiang, Cheng Chen, Cheng Li, Chenjun Xiao, Chenzhuang Du, Chonghua Liao, et~al.
\newblock Kimi k1. 5: Scaling reinforcement learning with llms.
\newblock \emph{arXiv preprint arXiv:2501.12599}, 2025.

\bibitem[Xu et~al.(2025)Xu, Wu, Wang, Li, Zheng, Chen, Hu, Kang, Ji, Zhang, Guo, Yang, Zhang, and Zhang]{xu2025redstardoesscalinglongcot}
Haotian Xu, Xing Wu, Weinong Wang, Zhongzhi Li, Da~Zheng, Boyuan Chen, Yi~Hu, Shijia Kang, Jiaming Ji, Yingying Zhang, Zhijiang Guo, Yaodong Yang, Muhan Zhang, and Debing Zhang.
\newblock Redstar: Does scaling long-cot data unlock better slow-reasoning systems?, 2025.
\newblock URL \url{https://arxiv.org/abs/2501.11284}.

\bibitem[Labs(2025)]{bespoke_stratos}
Bespoke Labs.
\newblock Bespoke-stratos: The unreasonable effectiveness of reasoning distillation, 2025.
\newblock URL \url{https://hf.co/bespokelabs/Bespoke-Stratos-32B}.
\newblock Accessed: 2025-01-22.

\bibitem[Yu et~al.(2024)Yu, Jiang, Shi, Yu, Liu, Zhang, Kwok, Li, Weller, and Liu]{yu2023metamath}
Longhui Yu, Weisen Jiang, Han Shi, Jincheng Yu, Zhengying Liu, Yu~Zhang, James~T. Kwok, Zhenguo Li, Adrian Weller, and Weiyang Liu.
\newblock Metamath: Bootstrap your own mathematical questions for large language models, 2024.
\newblock URL \url{https://arxiv.org/abs/2309.12284}.

\bibitem[Zelikman et~al.(2022)Zelikman, Wu, Mu, and Goodman]{zelikman2022star}
Eric Zelikman, Yuhuai Wu, Jesse Mu, and Noah Goodman.
\newblock Star: Bootstrapping reasoning with reasoning.
\newblock \emph{Advances in Neural Information Processing Systems}, 35:\penalty0 15476--15488, 2022.

\bibitem[Zelikman et~al.(2024)Zelikman, Harik, Shao, Jayasiri, Haber, and Goodman]{zelikman2024quiet}
Eric Zelikman, Georges Harik, Yijia Shao, Varuna Jayasiri, Nick Haber, and Noah~D Goodman.
\newblock Quiet-star: Language models can teach themselves to think before speaking.
\newblock \emph{arXiv preprint arXiv:2403.09629}, 2024.

\bibitem[Shao et~al.(2024)Shao, Wang, Zhu, Xu, Song, Bi, Zhang, Zhang, Li, Wu, and Guo]{shao2024deepseekmathpushinglimitsmathematical}
Zhihong Shao, Peiyi Wang, Qihao Zhu, Runxin Xu, Junxiao Song, Xiao Bi, Haowei Zhang, Mingchuan Zhang, Y.~K. Li, Y.~Wu, and Daya Guo.
\newblock Deepseekmath: Pushing the limits of mathematical reasoning in open language models, 2024.
\newblock URL \url{https://arxiv.org/abs/2402.03300}.

\bibitem[Chen et~al.(2024{\natexlab{b}})Chen, Deng, Yuan, Ji, and Gu]{chen2024self}
Zixiang Chen, Yihe Deng, Huizhuo Yuan, Kaixuan Ji, and Quanquan Gu.
\newblock Self-play fine-tuning converts weak language models to strong language models.
\newblock \emph{arXiv preprint arXiv:2401.01335}, 2024{\natexlab{b}}.

\bibitem[Wei et~al.(2023)Wei, Wang, Schuurmans, Bosma, Ichter, Xia, Chi, Le, and Zhou]{wei2023chainofthoughtpromptingelicitsreasoning}
Jason Wei, Xuezhi Wang, Dale Schuurmans, Maarten Bosma, Brian Ichter, Fei Xia, Ed~Chi, Quoc Le, and Denny Zhou.
\newblock Chain-of-thought prompting elicits reasoning in large language models, 2023.
\newblock URL \url{https://arxiv.org/abs/2201.11903}.

\bibitem[Snell et~al.(2024)Snell, Lee, Xu, and Kumar]{snell2024scalingllmtesttimecompute}
Charlie Snell, Jaehoon Lee, Kelvin Xu, and Aviral Kumar.
\newblock Scaling llm test-time compute optimally can be more effective than scaling model parameters, 2024.
\newblock URL \url{https://arxiv.org/abs/2408.03314}.

\bibitem[Welleck et~al.(2024)Welleck, Bertsch, Finlayson, Schoelkopf, Xie, Neubig, Kulikov, and Harchaoui]{welleck2024decodingmetagenerationinferencetimealgorithms}
Sean Welleck, Amanda Bertsch, Matthew Finlayson, Hailey Schoelkopf, Alex Xie, Graham Neubig, Ilia Kulikov, and Zaid Harchaoui.
\newblock From decoding to meta-generation: Inference-time algorithms for large language models, 2024.
\newblock URL \url{https://arxiv.org/abs/2406.16838}.

\bibitem[Muennighoff et~al.(2025)Muennighoff, Yang, Shi, Li, Fei-Fei, Hajishirzi, Zettlemoyer, Liang, Cand{\`e}s, and Hashimoto]{muennighoff2025s1}
Niklas Muennighoff, Zitong Yang, Weijia Shi, Xiang~Lisa Li, Li~Fei-Fei, Hannaneh Hajishirzi, Luke Zettlemoyer, Percy Liang, Emmanuel Cand{\`e}s, and Tatsunori Hashimoto.
\newblock s1: Simple test-time scaling.
\newblock \emph{arXiv preprint arXiv:2501.19393}, 2025.

\bibitem[Khot et~al.(2022)Khot, Trivedi, Finlayson, Fu, Richardson, Clark, and Sabharwal]{khot2022decomposed}
Tushar Khot, Harsh Trivedi, Matthew Finlayson, Yao Fu, Kyle Richardson, Peter Clark, and Ashish Sabharwal.
\newblock Decomposed prompting: A modular approach for solving complex tasks.
\newblock \emph{arXiv preprint arXiv:2210.02406}, 2022.

\bibitem[Zhou et~al.(2022)Zhou, Sch{\"a}rli, Hou, Wei, Scales, Wang, Schuurmans, Bousquet, Le, and Chi]{zhou2022least}
Denny Zhou, Nathanael Sch{\"a}rli, Le~Hou, Jason Wei, Nathan Scales, Xuezhi Wang, Dale Schuurmans, Olivier Bousquet, Quoc Le, and Ed~Chi.
\newblock Least-to-most prompting enables complex reasoning in large language models.
\newblock \emph{arXiv preprint arXiv:2205.10625}, 2022.

\bibitem[Hao et~al.(2023)Hao, Gu, Ma, Hong, Wang, Wang, and Hu]{hao2023reasoning}
Shibo Hao, Yi~Gu, Haodi Ma, Joshua~Jiahua Hong, Zhen Wang, Daisy~Zhe Wang, and Zhiting Hu.
\newblock Reasoning with language model is planning with world model.
\newblock \emph{arXiv preprint arXiv:2305.14992}, 2023.

\bibitem[Zheng et~al.(2023)Zheng, Mishra, Chen, Cheng, Chi, Le, and Zhou]{zheng2023take}
Huaixiu~Steven Zheng, Swaroop Mishra, Xinyun Chen, Heng-Tze Cheng, Ed~H Chi, Quoc~V Le, and Denny Zhou.
\newblock Take a step back: Evoking reasoning via abstraction in large language models.
\newblock \emph{arXiv preprint arXiv:2310.06117}, 2023.

\bibitem[Lightman et~al.(2023)Lightman, Kosaraju, Burda, Edwards, Baker, Lee, Leike, Schulman, Sutskever, and Cobbe]{lightman2023letsverifystepstep}
Hunter Lightman, Vineet Kosaraju, Yura Burda, Harri Edwards, Bowen Baker, Teddy Lee, Jan Leike, John Schulman, Ilya Sutskever, and Karl Cobbe.
\newblock Let's verify step by step, 2023.
\newblock URL \url{https://arxiv.org/abs/2305.20050}.

\bibitem[Ren et~al.(2023)Ren, Zhao, Vu, Liu, and Lakshminarayanan]{selfevaluation}
Jie Ren, Yao Zhao, Tu~Vu, Peter~J Liu, and Balaji Lakshminarayanan.
\newblock Self-evaluation improves selective generation in large language models.
\newblock \emph{arXiv preprint arXiv:2312.09300}, 2023.

\bibitem[Chen et~al.(2024{\natexlab{c}})Chen, Li, and Niu]{chen2024boosting}
Sijia Chen, Baochun Li, and Di~Niu.
\newblock Boosting of thoughts: Trial-and-error problem solving with large language models.
\newblock \emph{arXiv preprint arXiv:2402.11140}, 2024{\natexlab{c}}.

\bibitem[Madaan et~al.(2024)Madaan, Tandon, Gupta, Hallinan, Gao, Wiegreffe, Alon, Dziri, Prabhumoye, Yang, et~al.]{selfrefine}
Aman Madaan, Niket Tandon, Prakhar Gupta, Skyler Hallinan, Luyu Gao, Sarah Wiegreffe, Uri Alon, Nouha Dziri, Shrimai Prabhumoye, Yiming Yang, et~al.
\newblock Self-refine: Iterative refinement with self-feedback.
\newblock \emph{Advances in Neural Information Processing Systems}, 36, 2024.

\bibitem[Forsman(2024)]{selfrefinereport}
Anton Forsman.
\newblock Analyzing the performance of self-refine on different large language models, 2024.
\newblock URL \url{https://github.com/anforsm/self-refine/blob/main/report.pdf}.

\bibitem[Beirami et~al.(2025)Beirami, Agarwal, Berant, D'Amour, Eisenstein, Nagpal, and Suresh]{beirami2025theoreticalguaranteesbestofnalignment}
Ahmad Beirami, Alekh Agarwal, Jonathan Berant, Alexander D'Amour, Jacob Eisenstein, Chirag Nagpal, and Ananda~Theertha Suresh.
\newblock Theoretical guarantees on the best-of-n alignment policy, 2025.
\newblock URL \url{https://arxiv.org/abs/2401.01879}.

\bibitem[Wang et~al.(2023)Wang, Wei, Schuurmans, Le, Chi, Narang, Chowdhery, and Zhou]{selfconsistency}
Xuezhi Wang, Jason Wei, Dale Schuurmans, Quoc~V Le, Ed~H. Chi, Sharan Narang, Aakanksha Chowdhery, and Denny Zhou.
\newblock Self-consistency improves chain of thought reasoning in language models.
\newblock In \emph{The Eleventh International Conference on Learning Representations}, 2023.
\newblock URL \url{https://openreview.net/forum?id=1PL1NIMMrw}.

\bibitem[Yao et~al.(2023)Yao, Yu, Zhao, Shafran, Griffiths, Cao, and Narasimhan]{yao2023tree}
Shunyu Yao, Dian Yu, Jeffrey Zhao, Izhak Shafran, Tom Griffiths, Yuan Cao, and Karthik Narasimhan.
\newblock Tree of thoughts: Deliberate problem solving with large language models.
\newblock \emph{Advances in neural information processing systems}, 36:\penalty0 11809--11822, 2023.

\bibitem[Qi et~al.(2024)Qi, Ma, Xu, Zhang, Yang, and Yang]{qi2024mutual}
Zhenting Qi, Mingyuan Ma, Jiahang Xu, Li~Lyna Zhang, Fan Yang, and Mao Yang.
\newblock Mutual reasoning makes smaller llms stronger problem-solvers.
\newblock \emph{arXiv preprint arXiv:2408.06195}, 2024.

\bibitem[Zhang et~al.(2024)Zhang, Wu, Lei, Che, Li, Xie, Huang, Zhang, Pavone, Li, Ouyang, and Zhou]{zhang2024llamaberrypairwiseoptimizationo1like}
Di~Zhang, Jianbo Wu, Jingdi Lei, Tong Che, Jiatong Li, Tong Xie, Xiaoshui Huang, Shufei Zhang, Marco Pavone, Yuqiang Li, Wanli Ouyang, and Dongzhan Zhou.
\newblock Llama-berry: Pairwise optimization for o1-like olympiad-level mathematical reasoning, 2024.
\newblock URL \url{https://arxiv.org/abs/2410.02884}.

\bibitem[Kang et~al.(2024)Kang, Li, Chen, Kazemi, Sun, Chen, Li, He, He, Wen, et~al.]{kang2024mindstar}
Jikun Kang, Xin~Zhe Li, Xi~Chen, Amirreza Kazemi, Qianyi Sun, Boxing Chen, Dong Li, Xu~He, Quan He, Feng Wen, et~al.
\newblock Mindstar: Enhancing math reasoning in pre-trained llms at inference time.
\newblock \emph{arXiv preprint arXiv:2405.16265}, 2024.

\bibitem[Kocsis and Szepesv{\'a}ri(2006)]{kocsis2006bandit}
Levente Kocsis and Csaba Szepesv{\'a}ri.
\newblock Bandit based monte-carlo planning.
\newblock In \emph{European conference on machine learning}, pages 282--293. Springer, 2006.

\bibitem[Schick et~al.(2024)Schick, Dwivedi-Yu, Dess{\`\i}, Raileanu, Lomeli, Hambro, Zettlemoyer, Cancedda, and Scialom]{schick2024toolformer}
Timo Schick, Jane Dwivedi-Yu, Roberto Dess{\`\i}, Roberta Raileanu, Maria Lomeli, Eric Hambro, Luke Zettlemoyer, Nicola Cancedda, and Thomas Scialom.
\newblock Toolformer: Language models can teach themselves to use tools.
\newblock \emph{Advances in Neural Information Processing Systems}, 36, 2024.

\bibitem[Saad-Falcon et~al.(2024)Saad-Falcon, Lafuente, Natarajan, Maru, Todorov, Guha, Buchanan, Chen, Guha, R{\'e}, et~al.]{saad2024archon}
Jon Saad-Falcon, Adrian~Gamarra Lafuente, Shlok Natarajan, Nahum Maru, Hristo Todorov, Etash Guha, E~Kelly Buchanan, Mayee Chen, Neel Guha, Christopher R{\'e}, et~al.
\newblock Archon: An architecture search framework for inference-time techniques.
\newblock \emph{arXiv preprint arXiv:2409.15254}, 2024.

\bibitem[Chen et~al.(2022)Chen, Ma, Wang, and Cohen]{chen2022program}
Wenhu Chen, Xueguang Ma, Xinyi Wang, and William~W Cohen.
\newblock Program of thoughts prompting: Disentangling computation from reasoning for numerical reasoning tasks.
\newblock \emph{arXiv preprint arXiv:2211.12588}, 2022.

\bibitem[Gao et~al.(2023)Gao, Xiong, Gao, Jia, Pan, Bi, Dai, Sun, and Wang]{gao2023retrieval}
Yunfan Gao, Yun Xiong, Xinyu Gao, Kangxiang Jia, Jinliu Pan, Yuxi Bi, Yi~Dai, Jiawei Sun, and Haofen Wang.
\newblock Retrieval-augmented generation for large language models: A survey.
\newblock \emph{arXiv preprint arXiv:2312.10997}, 2023.

\bibitem[Srednicki(1993)]{Srednicki:1993im}
Mark Srednicki.
\newblock {Entropy and area}.
\newblock \emph{Phys. Rev. Lett.}, 71:\penalty0 666--669, 1993.
\newblock \doi{10.1103/PhysRevLett.71.666}.

\end{thebibliography}

\appendix

\section{Summary of Problem Data}

For each problem we collect the following data.

\begin{itemize}
    \item \textbf{Problem Title}: Up to one sentence describing the problem.    
    \item \textbf{Problem Statement}: The problem statement in LaTeX.
    \item \textbf{Problem Solution}: The full solution to the problem in LaTeX.
    \item \textbf{Public problem}: yes/no.  
    \item \textbf{Auto-verifiable}: yes/no. All problems in the data set for this publication are auto-verifiable.
    \item \textbf{Auto-verifier instructions}: Instructions how to output the solution for the auto-verifier. See Sec. \ref{sec:autoverifier}.
    \item \textbf{Domain of theoretical physics}: e.g. High Energy Theory.
    \item \textbf{Difficulty level}: 1-5
    \item \textbf{Authors}: The contributors of the problem and solution.
    \item \textbf{Reviewers}: The reviewers of the problem and solution.
    \item \textbf{Problem origin and novelty}: How closely existing published work contains the solution (only above undergraduate level).  
    \item \textbf{Problem ID}: Unique Problem ID in our catalogue.  
    \item \textbf{Problem Version}: In some cases there may be errors or ambiguities in a problem. For this case we track a version number. 
    \item \textbf{Variation of a different problem}: In the future, we aim to provide minor modifications of existing problems to check stability of the reasoning chain (as opposed to memorization). Standard: No
    \item \textbf{Date problem was added to the data set}: Allows us to track new problems. Format: 01/31/2025.
    \item \textbf{Author comments}: Any additional comments the author has about the problem.
\end{itemize}

\section{Prompts}
\label{sec:llmprompt}
\subsection{Prompts to Query Problem Solutions}

We used two different system prompts to initialize the LLMs, as well as a unique user prompt to query individual solutions. 

\subsubsection*{Simple System Prompt}
Our simple system prompt only specifies the required output format and encourages complete calculations.

\begin{tcolorbox}
\footnotesize
\textcolor{usercolor}
{\textbf{System:}}
You are a mathematical problem-solving assistant specializing in theoretical physics.\\ Input problems will be provided in LaTeX format, and you must provide your solutions in LaTeX format as well.\\Please provide detailed step-by-step solutions and clearly mark your final answer with `Final Answer:' at the end.\\When writing equations, ensure proper LaTeX formatting including appropriate equation environments and mathematical notation.
\end{tcolorbox}

\subsubsection*{Extended System Prompt with CoT advise}

Our extended system prompt includes additional problem solving advice inspired by Polya's book \say{How to Solve It}. \cite{Polya1945}. We have used this system prompt as our default. However, we did not find a systematic difference between these prompts as illustrated in Table \ref{tab:difficulty_comparison} for a subset of problems.

\begin{tcolorbox}
\footnotesize
\textcolor{usercolor}
{\textbf{System:}} You are a mathematical problem-solving assistant specializing in theoretical physics.\\Input problems will be provided in LaTeX format, and you must provide your solutions in LaTeX format as well.\\Please provide detailed step-by-step solutions and clearly mark your final answer with `Final Answer:' at the end.\\When writing equations, ensure proper LaTeX formatting including appropriate equation environments and mathematical notation.\\Please follow a structured and logical approach. Here are your key steps for solving any problem:\\1. Understand the Problem:\\- Identify the unknown, the given data, and the conditions.\\- Evaluate if the conditions are sufficient, redundant, or contradictory.\\- Break down and analyze the different parts of the condition.\\2. Devise a Plan:\\- Explore connections between the data and the unknown.\\- If necessary, consider auxiliary problems to bridge gaps.\\- Reflect on whether you have encountered similar problems or solutions before.\\- Look for related problems, theorems, or methods that might apply.\\- Consider simplifying or reformulating the problem to make it more accessible.\\- Use definitions and explore analogous, general, or special cases.\\3. Carry Out the Plan:\\- Execute your solution step by step, ensuring each step is clear and logically valid.\\- Confirm the correctness of each step and justify your reasoning.\\For each problem, ensure clarity, logical rigor, and consistency. You may iterate to refine and improve your solution.
\end{tcolorbox}

\begin{table}[h]
    \footnotesize
    \centering
    \renewcommand{\arraystretch}{1.2}
    \begin{tabular}{l | c c | c c | c c | c c | c c | c c}
        \toprule
        \multirow{2}{*}{Difficulty Level} 
        & \multicolumn{2}{c|}{A} & \multicolumn{2}{c|}{B} & \multicolumn{2}{c|}{C} 
        & \multicolumn{2}{c|}{D} & \multicolumn{2}{c|}{avg@5} & \multicolumn{2}{c}{best@5} \\
        & Ext & Std & Ext & Std & Ext & Std 
        & Ext & Std & Ext & Std & Ext & Std \\
        \midrule
        1-Easy Undergrad   &20&23&3&1&2&1&0&0&0.72&0.76&0.8&0.8 \\
        2-Undergrad        &22&17&1&0&2&8&0&0&0.88&0.68&1.0&1.0\\
        3-Easy Grad        &8&10&3&3&14&12&0&0&0.16&0.24&0.4&0.6\\
        \bottomrule
    \end{tabular}
    \caption{Performance Comparison for different system prompts, using the  GPT-4o model, on a subset of problems.}
    \label{tab:difficulty_comparison}
\end{table}

\subsubsection*{User Prompt}
\begin{tcolorbox}
\footnotesize
\textcolor{usercolor}
{\textbf{User:}} 
Problem:\\
{problem[``problem\_details"][``Problem Statement"]}\\

IMPORTANT SOLUTION REQUIREMENTS:\\
1. You MUST FIRST solve this problem using mathematical reasoning and symbolic calculations:\\
   - Use proper mathematical notation and symbols\\
   - Arrive at a final symbolic mathematical expression\\
   
2. ONLY AFTER completing the mathematical solution:\\
   - Convert your final mathematical expression into Python code\\
   - The code must satisfy these requirements:\\
{problem[``problem\_details"][``Answer Requirements"]}\\

Code Format Requirements:\\
1. Your solution MUST include the final executable Python code as required by the ``Answer Requirements"\\
2. You MUST wrap the final Python code between \texttt{\`{}\`{}\`{}}python and \texttt{\`{}\`{}\`{}} tags\\
3. Ensure the code is complete and can run independently\\
4. The code should NOT contain ANY externally defined variables, including physical constants.\\
\end{tcolorbox}

\subsection{Prompts to Query Grading of Solutions}

\subsubsection*{System Prompt}
\begin{tcolorbox}
\footnotesize
\textcolor{usercolor}
{\textbf{System:}}
You are a grader for machine learning model solutions of theoretical physics problems. I will provide you with a correct expert solution to the problem for your reference, and a model solution for you to grade.\\
Grade solutions using A/B/C/D grades where:\\A=Excellent: The solution is mathematically equivalent to the expert solution, even if the symbolic expression differs (e.g. terms are arranged differently). The solution includes all necessary steps and the reasoning in each step is correct. Different but valid solution methods are acceptable.\\B=Good with minor issues: Generally correct solution with small errors such as: arithmetic mistakes that don't affect the main approach, missing intermediate steps, or minor notation issues. The problem was correctly understood and the reasoning of the solution is generally correct.\\C=Significant issues but partially correct: Shows basic understanding but has major flaws such as: incorrect application of formulas, missing crucial steps, or computational errors that lead to wrong final answer. The approach has some merit despite errors.\\D=Incorrect or major issues: Fundamentally flawed approach, completely incorrect calculations, or missing essential components. Shows little to no understanding of the mathematical concepts involved.\\When comparing final answers, verify that the equations or expressions are mathematically equivalent (e.g., $2x + 2$ is equivalent to $2(x + 1)$). Always format your notes using LaTeX notation for mathematical expressions. Provide evaluation in compact JSON format with only 'grade' and 'notes' fields. Format all mathematical expressions in your notes using LaTeX notation (e.g., \$x\texttt{\^}2\$, \$\textbackslash
frac\{1\}\{2\}\$, \$\textbackslash sqrt\{x\}\$
).
\end{tcolorbox}

\subsubsection*{User Prompt}
\begin{tcolorbox}
\footnotesize
\textcolor{usercolor}
{\textbf{User:}}
Compare the following model solution detailed steps with the expert solution, along with the code verification result which check the equivalence of 2 expression numerically, and evaluate its correctness. \\Expert Solution: \{expert\_solution\}\\Model Solution \{model\_solution\} \\ Code Verification result: \{code\_verification\_result\}\\Format your response as JSON with the following structure:\\\{\\    ``grade": ``A/B/C/D",\\    ``notes": ``your notes here with LaTeX math notation"\\ \}
\end{tcolorbox}

\clearpage

\section{Public Problems and Solutions}
\label{app:sampleproblems}
\normalsize
We list ten public sample problems along with their solutions. AI model results for these problems are available on the dataset website \url{tpbench.org}. 
Table \ref{tab:model_scores} summarizes the performance of different AI models on these problems, covering a range of topics and difficulty levels from Level 1 (L1) to Level 5 (L5). The scores indicate the average accuracy of the 5 attempts of each model.

\begin{table}[h]
\centering
\begin{tabular}{lccccc}
\hline
Problem ID & Llama-70B & GPT-4o & R1 & o1 & o3-mini \\
\hline
Boosted Parabolic Trajectory (L1) & 0.60 & 1.00 & 1.00 & 1.00 & 1.00 \\
Blackbody in d Dimensions (L1) & 0.20 & 0.40 & 1.00 & 1.00 & 1.00 \\
A 3-State QM Problem (L2) & 0.40 & 0.80 & 1.00 & 1.00 & 1.00\\
Dark Matter Capture as a Function of Time (L2) & 0.60 & 1.00 & 1.00 & 1.00 & 1.00 \\
Slow-Roll Inflation (L3) & 0.00 & 0.00 & 1.00 & 1.00 & 1.00 \\
Scalar Particle Scattering (L3) & 0.00 & 0.40 & 0.80 & 0.40 & 0.40 \\
SHO Vacuum Entanglement (L4) & 0.00 & 0.00 & 0.80 & 0.00 & 1.00 \\
SUSY-Symmetry (L4) & 0.00 & 0.00 & 0.00 & 0.00 & 0.00 \\
Bias of a Sampled Halo Field (L5) & 0.00 & 0.00 & 0.60 & 1.00 & 0.80 \\
One-Pole Problem (L5) & 0.00 & 0.00 & 0.00 & 0.00 & 0.00 \\
\hline
\end{tabular}
\caption{Model Average Scores by Problem}
\label{tab:model_scores}
\end{table}

\subsection{Level 5 - One-Pole Problem}
\label{app:onepole}
\footnotesize

\subsubsection*{Problem Statement}

Consider the conformally coupled scalar field $\phi$ 
\begin{equation}
\mathcal{L}=\frac{1}{2}\left[g^{\mu\nu}\partial_{\mu}\phi\partial_{\nu}\phi-\left(m^{2}-\frac{1}{6}R\right)\phi^{2}\right]
\end{equation}
 in curved spacetime
\[
ds^{2}=a^{2}(\eta)\left(d\eta^{2}-|d\vec{x}|^{2}\right)
\]
where the Ricci scalar is 
\begin{equation}
R=-6\frac{a''(\eta)}{a(\eta)}
\end{equation}
and $a$ satisfies the differential equation
\begin{equation}
\frac{d}{dt}\ln a=\Theta(t_{e}-t)H_{I}+\Theta(t-t_{e})\frac{H_{I}}{1+\frac{3}{2}H_{I}(t-t_{e})}\label{eq:scalefactor-onepole}
\end{equation}
with $t_{e}$ a finite positive number, the $\Theta$ function having
the steplike behavior
\begin{equation}
\Theta(t-t_{e})\equiv\begin{cases}
1 & t\geq t_{e}\\
0 & \mbox{otherwise}
\end{cases},
\end{equation}
and $t$ being the comoving proper time related to $\eta$ through
\begin{equation}
t=t_{e}+\int_{\eta_{e}}^{\eta}a(y)dy.
\end{equation}
The boundary condition for the differential equation (in comoving
proper time) is $a|_{t=t_{e}}=a_{e}$.

In the limit that $k/(a_{e}H_{I})\rightarrow\infty$, using the steepest
descent approximation starting from the dominant pole $\tilde{\eta}$
(with $\Re\tilde{\eta}>0$) of the integrand factor $\omega_{k}'(\eta)/\left(2\omega_{k}(\eta)\right)$,
compute the Bogoliubov coefficient magnitude $|\beta(k)|$ approximated
as 
\begin{equation}
|\beta(k)|\approx\left|\int_{-\infty}^{\infty}d\eta\frac{\omega_{k}'(\eta)}{2\omega_{k}(\eta)}e^{-2i\int_{\eta_{e}}^{\eta}d\eta'\omega_{k}(\eta')}\right|\label{eq:originalbeta}
\end{equation}
for particle production where the dispersion relationship given by
\begin{equation}
\omega_{k}^{2}(\eta)=k^{2}+m^{2}a^{2}(\eta)
\end{equation}
with $0<m\lesssim H_{I}$. Use a one pole approximation which dominates
in this limit.

\subsubsection*{Answer Requirements}
Provide the answer in the form of the \texttt{python} code. Implement the following function.
\begin{python}
def abs_beta(k:float, a_e:float, m:float, H_I:float) -> float:
    pass
\end{python}

\subsubsection*{Comments about the Problem}

\textit{This is an example of a difficult problem from Quantum Field Theory in curved spacetime, dealing with gravitational particle production, that appears out of reach of current models. This is part of a published research work and the solution, without steps explained, is given in a footnote of \cite{Basso:2022tpd}, but would be difficult to locate (in fact we tried, without success, with OpenAI's Deep Research). 
}
\subsubsection*{Solution}

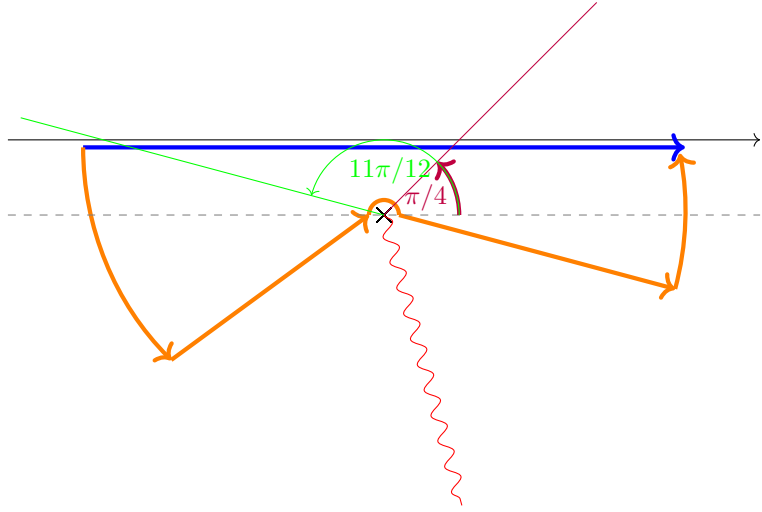
\begin{figure}
\begin{centering}
\begin{tikzpicture} 
\draw[->] (-5,0) -- (5,0); 
\draw[blue,ultra thick, ->] (-4,-0.1) -- (4,-0.1); 
\draw plot[only marks, mark=x, mark size=4pt](0,-1);
\draw[ultra thick, orange, ->] (-4,-.1) arc (180:225:4);
\draw[ultra thick, orange, ->] (-2.82843, -2.92843) -- (-.2,-1);
\draw[ultra thick, orange] (-.2,-1) arc (180:0:.2);
\draw[ultra thick, orange,->] (+.2,-1) -- (3.87052, -1.98351);
\draw[ultra thick, orange, ->] (3.87052, -1.98351) arc (345:371:4);
\draw[gray, dashed] (-5,-1) -- (5, -1);
\draw[purple] (0,-1) -- (2.82843, 1.82843);
\draw[green] (0,-1) -- (-4.82963, 0.294095);
\draw[draw=red, snake it] (0,-1) -- (1.03528, -4.8637);
\usetikzlibrary {angles,quotes} 
\draw [transparent] (5,-1) coordinate (A) -- (0,-1) coordinate (B)
         -- (2.82843, 1.82843) coordinate (C);
\draw pic ["$\pi/4$",ultra thick, draw,->,purple,angle radius=1cm] {angle = A--B--C}; 
\draw [transparent] (5,-1) coordinate (D) -- (0,-1) coordinate (E)
         -- (-4.82963, 0.294095) coordinate (F);
\draw pic ["$11\pi/12$", draw,->,green,angle radius=1cm] {angle = D--E--F}; 

\end{tikzpicture}
\par\end{centering}
\caption{\label{fig:The-original-contour}\footnotesize The original contour in blue is deformed
into the orange contour in the lower half complex plane of $\eta$.
The large radius arcs have vanishing contributions, and one-pole approximation
has been taken. The upper green and purple boundaries correspond to
where integrations over any arcs extended beyond this boundary would not converge. The
dashed horizontal curve is parallel to the real axis.  The red squiggly line is the branch cut at $-5 \pi/12$.}
\end{figure}

To find the pole of $\omega_{k}'(\eta)/\omega_{k}(\eta)$, we need
$a(\eta)$ from the given differential equation
\begin{equation}
\frac{d\ln a}{dt}=\Theta(t_{e}-t)H_{I}+\Theta(t-t_{e})\frac{H_{I}}{1+\frac{3}{2}H_{I}(t-t_{e})}.\label{eq:Hubble}
\end{equation}
Integrating from time $t=t_{e}$, we find
\begin{align}
\ln\frac{a}{a_{e}} & =\int_{t_{e}}^{t}dT\frac{H_{I}}{1+\frac{3}{2}H_{I}(T-t_{e})}\\
 & =\frac{2}{3}\ln\left[1+\frac{3}{2}H_{I}(T-t_{e})\right]_{t_{e}}^{t}\\
 & =\frac{2}{3}\ln\left[1+\frac{3}{2}H_{I}(t-t_{e})\right]
\end{align}
for $t\geq t_{e}$. In other words, this scale factor
\begin{equation}
\frac{a}{a_{e}}=\left[1+\frac{3}{2}H_{I}(t-t_{e})\right]^{2/3}\label{eq:aHeq}
\end{equation}
behaves as a typical coherent oscillations spacetime minus the oscillatory
effects. Hence, note that for $t\gg t_{e}$, the scale factor can
be approximated as 
\begin{equation}
a(\eta)\approx c_{1}\eta^{2}
\end{equation}
for $\eta\gg\eta_{e}$ (where $\eta_{e}$ is the corresponding conformal
time for $t_{e}$) where we see by matching 
\begin{equation}
\int_{\eta_{i}}^{\eta}a(\eta)d\eta=t-t_{i}
\end{equation}
with $\eta_{i}\gg\eta_{e}$ and $t_{i}\gg t_{e}$, we can write
\begin{equation}
\frac{1}{3}c_{1}\eta^{3}\approx t
\end{equation}
for times much larger than $\eta_{i}$. This means that at time $\eta_{i}\gg\eta_{e}$,
we have
\begin{equation}
c_{1}\approx\frac{2}{H(\eta_{i})\eta_{i}^{3}}
\end{equation}
(where the Hubble expansion rate is $H(\eta)=a'(\eta)/a^{2}(\eta)$)
which gives
\begin{equation}
a(\eta)\approx\frac{2\eta^{2}}{H(\eta_{i})\eta_{i}^{3}}
\end{equation}
for $\eta>\eta_{i}$ where the choice of $\eta_{i}$ controls the
approximation error proportional to positive power of $\eta_{e}/\eta_{i}$.
Since $\eta_{i}\gg\eta_{e}>0$, we can approximate $\eta=0$ to be
equivalent to $\eta-\eta_{i}\rightarrow-\infty$. In other words,
when we analytically continue and consider the poles of the integrand,
we will consider only the region with $\Re\eta>0$.

Next, note the pole of

\begin{equation}
\frac{\omega'}{2\omega}=\frac{m^{2}\partial_{\eta}a^{2}}{4\left(k^{2}+m^{2}a^{2}\right)}
\end{equation}
is at $\tilde{\eta}$ defined by 
\begin{equation}
k^{2}=-m^{2}a^{2}(\tilde{\eta})\label{eq:poleexpl}
\end{equation}
which means
\begin{align*}
\tilde{\eta} & =\sqrt{\frac{H(\eta_{i})\eta_{i}^{3}}{2}}\left(\frac{-k^{2}}{m^{2}}\right)^{1/4}\\
 & =\eta_{i}\sqrt{\frac{1}{a(\eta_{i})}}\left(\frac{-k^{2}}{m^{2}}\right)^{1/4}\\
 & =\eta_{i}e^{i(2l+1)\pi/4}\frac{\sqrt{k/a(\eta_{i})}}{\sqrt{m}}\numberthis \label{eq:branchpoints}
\end{align*}
where $l$ is an integer. We see that $\Re\tilde{\eta}\gg\eta_{i}$
for $k/a(\eta_{e})\gg k/a(\eta_{i})\gg m$. We also see that $l\in\{1,2\}$
have negative $\Re\tilde{\eta}$ which are in the region that we excised
with the $\eta-\eta_{i}\rightarrow-\infty$ discussed above. That
means we can consider either $l\in\{3,4\}$. We will see below that
one of these poles is irrelevant.

Eq.~(\ref{eq:originalbeta}) tells us that

\begin{align*}
\left|\beta(\eta)\right| & =\left|\int_{-\infty}^{\infty}d\eta\frac{\omega_{k}'(\eta)}{2\omega_{k}(\eta)}e^{-2i\int_{\eta_{e}}^{\eta}d\eta'\omega_{k}(\eta')}\right|\\
 & =\left|\int_{-\infty}^{\infty}d\eta\frac{\omega_{k}'(\eta)}{2\omega_{k}(\eta)}e^{-2i\int_{\eta_{i}}^{\eta}d\eta'\omega_{k}(\eta')}e^{-2i\int_{\eta_{e}}^{\eta_{i}}d\eta'\omega_{k}(\eta')}\right|\\
 & =\left|\int_{-\infty}^{\infty}d\eta\frac{\omega_{k}'(\eta)}{2\omega_{k}(\eta)}e^{-2i\int_{\eta_{i}}^{\eta}d\eta'\omega_{k}(\eta')}\right|.\numberthis
\end{align*}
With the steepest descent technique starting from the pole of $\omega_{k}'/\omega_{k}$,
we write after analytically continuing $\eta$
\begin{align*}
\left|\beta\right| & =\left|\int_{-\infty}^{\infty}d\eta\frac{\omega_{k}'(\eta)}{2\omega_{k}(\eta)}e^{-2i\left[\int_{\eta_{i}}^{\tilde{\eta}}d\eta'\omega_{k}(\eta')+\int_{\tilde{\eta}}^{\eta}d\eta'\omega_{k}(\eta')\right]}\right|\\
 & =\left|e^{-2i\int_{\eta_{i}}^{\tilde{\eta}}d\eta'\omega_{k}(\eta')}v\right|\numberthis\label{eq:betaeq}
\end{align*}
where $\tilde{\eta}$ is the pole of $\omega_{k}'(\eta)/\omega_{k}(\eta)$
and  $v$ is the part obtained from the steepest descent. The factor
in the integrand of Eq.~(\ref{eq:originalbeta}) is therefore
\begin{equation}
\frac{\omega'}{2\omega}\approx\frac{1}{4(\eta-\tilde{\eta})}
\end{equation}
which implies $v$ in eq.~(\ref{eq:betaeq}) is
\begin{equation}
v=\int_{-\infty}^{\infty}\frac{d\eta}{4(\eta-\tilde{\eta})}e^{-\frac{4}{3}im\sqrt{C'(\tilde{\eta})}(\eta-\tilde{\eta})^{3/2}}\label{eq:vexplic}
\end{equation}
where 
\begin{equation}
C(\eta)\equiv a^{2}(\eta).
\end{equation}
Deforming the integration contour as shown in Fig.~\ref{fig:The-original-contour}
allows us to rewrite this as
\begin{equation}
v=\int_{\mathcal{C}}\frac{d\eta}{4(\eta-\tilde{\eta})}e^{-\frac{4}{3}im\sqrt{C'(\tilde{\eta})}(\eta-\tilde{\eta})^{3/2}}
\end{equation}
where the $\mathcal{C}$ is the orange part of the contour in the
lower half plane. 

To define the contour, one must understand the complex values of $C'(\tilde{\eta})$.
To this end, let
\begin{equation}
-i\sqrt{C'(\tilde{\eta})}=U+iW
\end{equation}
where the imaginary part generically is nonvanishing. The branch points
are given by eqs.~(\ref{eq:branchpoints}) which gives
\begin{align}
C'(\tilde{\eta}) & =\frac{4a^{2}(\eta_{i})}{\eta_{i}}e^{\frac{3}{4}i(2l+1)\pi}\left(\frac{k/a(\eta_{i})}{m}\right)^{3/2}
\end{align}
which says 
\begin{align*}
U+iW & =\frac{2a(\eta_{i})}{\sqrt{\eta_{i}}}e^{\frac{1}{8}i(6l-1)\pi}\left(\frac{k/a(\eta_{i})}{m}\right)^{3/4}\\
 & =a^{3/2}(\eta_{i})\sqrt{2H(\eta_{i})}e^{\frac{1}{8}i(6l-1)\pi}\left(\frac{k/a(\eta_{i})}{m}\right)^{3/4}.\numberthis
\end{align*}

To deform the contour, we need regions where the arcs with large radius
does not contribute to the integral. Note that if we define $\delta\equiv\eta-\tilde{\eta}=Re^{i\theta}$,
we have
\begin{equation}
\delta^{3/2}=R^{3/2}e^{i3\theta/2}=R^{3/2}\left(\cos\frac{3\theta}{2}+i\sin\frac{3\theta}{2}\right)
\end{equation}
making the exponent in $v$ 
\begin{equation}
-\frac{4}{3}im\sqrt{C'(\tilde{\eta})}(\eta-\tilde{\eta})^{3/2}=\frac{4}{3}mR^{3/2}(U+iW)\left(\cos\frac{3\theta}{2}+i\sin\frac{3\theta}{2}\right)
\end{equation}
which is damped only if 
\begin{equation}
U\cos(3\theta/2)-W\sin(3\theta/2)<0.
\end{equation}
For the case of Eq.~(\ref{eq:poleexpl}), we need 
\begin{equation}
\cos\left[\frac{\pi}{8}(6l-1)\right]\cos(3\theta/2)-\sin\left[\frac{\pi}{8}(6l-1)\right]\sin(3\theta/2)<0
\end{equation}
for one choice of $l$. For the choice of $l=3$, we can choose the
arc regions to be $\theta\in[\frac{-5\pi}{12},\frac{\pi}{4}]$ and
another arc region to be $\theta\in[\frac{11\pi}{12},\frac{19\pi}{12}]$
with a branch cut at $-5\pi/12$. 

Choosing $l=3$, we find the steepest descent contour shown in orange
in Fig.~\ref{fig:The-original-contour}. The left contour is $5\pi/4$
and the right contour is at $-\pi/12$, along which 
\begin{align*}
-\frac{4}{3}im\sqrt{C'(\tilde{\eta})}(\eta-\tilde{\eta})^{3/2} & =-\frac{4}{3}mR^{3/2}a^{3/2}(\eta_{i})\sqrt{2H(\eta_{i})}\left(\frac{k/a(\eta_{i})}{m}\right)^{3/4}
\end{align*}
gives a damped exponential in eq.~(\ref{eq:vexplic}). Hence, the
integral is
\begin{align*}
v & =\frac{1}{4}\int_{\infty}^{\epsilon}\frac{dR}{R}e^{-\frac{4}{3}mR^{3/2}a^{3/2}(\eta_{i})\sqrt{2H_{e}}\left(\frac{k/a(\eta_{i})}{m}\right)^{3/4}}+\frac{1}{4}\int_{\epsilon}^{\infty}\frac{dR}{R}e^{-\frac{4}{3}mR^{3/2}a^{3/2}(\eta_{i})\sqrt{2H_{e}}\left(\frac{k/a(\eta_{i})}{m}\right)^{3/4}}\nonumber\\
 & + \frac{1}{4}\int_{5\pi/4}^{-\pi/12}id\theta\exp\left[-\frac{4}{3}im\sqrt{C'(\tilde{\eta})}(\epsilon\, e^{i\theta})^{3/2}\right]\\
 & =\frac{i}{4}\left[\frac{-\pi}{12}-\frac{15\pi}{12}\right]=\frac{-i\pi}{3}\numberthis
\end{align*}
where in the first line we have introduced a regulator $\epsilon\rightarrow0$.

The final piece in eq.~(\ref{eq:betaeq}) is
\begin{align}
I & =e^{-2i\int_{\eta_{i}}^{\tilde{\eta}}d\eta'\omega_{k}(\eta')}.
\end{align}
Use the expansion 
\begin{align*}
I & =e^{-2i\int_{\eta_{i}}^{\tilde{\eta}}d\eta'\omega_{k}(\eta')}\\
 & =\exp\left(-2i\left[\Phi+J\right]\right)\numberthis
\end{align*}
where $\Phi$ is real and $J$ is purely imaginary. We take the path
to be along the real axis until $\eta=\Re\tilde{\eta}$ and then integrate
in the imaginary $\eta$ direction:
\begin{equation}
J=i\Im\int_{\Re\tilde{\eta}}^{\Re\tilde{\eta}+i\Im\tilde{\eta}}d\eta'\omega_{k}(\eta').
\end{equation}
 This gives
\begin{align}
J & \approx-i\frac{2}{3}\sqrt{2\pi}\frac{\Gamma(5/4)}{\Gamma(3/4)}\frac{(k/a(\eta_{i}))^{3/2}}{H(\eta_{i})\sqrt{m}}.
\end{align}
Now, note from Eq.~(\ref{eq:aHeq}), we can compute
\begin{align*}
\frac{1}{a_{e}^{3/2}} & =\frac{1}{a^{3/2}(\eta_{i})}\left[1+\frac{3}{2}H_{I}(t_{i}-t_{e})\right]\\
 & \approx\frac{1}{a^{3/2}(\eta_{i})}\frac{3}{2}H_{I}t_{i}\\
 & \approx\frac{1}{a^{3/2}(\eta_{i})}\frac{H_{I}}{H(\eta_{i})}\numberthis
\end{align*}
where we used Eq.~(\ref{eq:Hubble}). Eq.~(\ref{eq:betaeq}) then
becomes
\begin{equation}
\boxed{|\beta|\approx\frac{\pi}{3}\exp\left(-\frac{4}{3}\sqrt{2\pi}\frac{\Gamma(5/4)}{\Gamma(3/4)}\frac{(k/a_{e})^{3/2}}{H_{I}\sqrt{m}}\right)}.\label{eq:exampleresult}
\end{equation}

\clearpage

\subsection{Level 5 - Bias of a Sampled Halo Field}
\label{sec:prob_bias}

\subsubsection*{Problem Statement}

In cosmology, large-scale cosmological dark-matter halo fields are biased tracers of the underlying Gaussian matter density $\delta_m$. Assume we have a sample $\delta_m$. We simulate a halo number density field by taking $n(\mathbf{x}) = \bar{n}\max(0,1+b\delta_m(\mathbf{x}))$, where bare number density $\bar{n}$ and bare bias $b$ are specified constants. What is the bias of the sampled halo field? Derive an equation to evaluate the bias which depends on the bare bias and the variance in each pixel.

\subsubsection*{Answer Requirements}
Provide the answer in the form of the \texttt{python} code. Implement the following function.
\begin{python}
#let b_in stand for bare bias
def b_eff(sigma: float, b_in:float) -> float:
    pass
\end{python}

\subsubsection*{Comments about the Problem}

\textit{This is an example of a cosmology research problem that is being solved correctly by advanced reasoning models. This may be because the calculation is similar to existing calculations in the literature. However, this is a genuine research problem, which we solved independently, for an upcoming cosmology publication. The problem requires to retrieve some background knowledge, such as the definition of the matter power spectrum in cosmology.}

\subsubsection*{Solution}

The solution to this question involves some domain knowledge, parts of which were given in the problem's statement, some approximations sourced by the domain knowledge, and some mathematical calculations. The domain knowledge is very basic and should be known to anyone in the field. Approximations are intuitive and also, mostly, inspired by the domain knowledge. Following Polya, we can organize it as follows: \\

\textbf{Understand the problem.} The number density of halos $n_h(\mathbf{x})$ is defined as
\begin{equation}    N_h = \int_{V} n_h(\mathbf{x})d\mathbf{x}.\end{equation}
The overdensity is defined as \begin{equation}    \delta_h(\mathbf{x}) = \frac{n_h(\mathbf{x})-\langle n_h(\mathbf{x})\rangle}{\langle n_h(\mathbf{x})\rangle}.\end{equation}
Linear bias is defined in terms of Fourier-transformed quantities:\begin{equation}    \delta_h(\mathbf{k}) = b\delta_m(\mathbf{k}).\end{equation} 
This is an approximation that holds on sufficiently large scales (small $k$). $\delta_m(\mathbf{k})$ and $\delta_h(\mathbf{k})$ are Gaussian random fields with zero mean and their variance depends only on the magnitude of the wave-vector $k=|\mathbf{k}|$: 
\begin{equation}    \delta_m \sim \mathcal{N}(0,P_{mm}(k)),\ \delta_h \sim \mathcal{N}(0,P_{hh}(k)).\end{equation}The quantity $P(k)$ is called the power spectrum and is defined as \begin{equation}    \langle\delta(\mathbf{k})\delta(\mathbf{k'})\rangle = (2\pi)^3\delta^D(\mathbf{k+k'})P(k).\end{equation}
It immediately follows that\begin{equation}    P_{hh}(k) = b^2P_{mm}(k).\end{equation}
We are given the expression in real space. In real space, the quantity $\delta_m(\mathbf{x})$ is also a Gaussian random field:\begin{equation}    \delta_m(\mathbf{x}) \sim \mathcal{N}(0, \xi_m),\ \delta_h(\mathbf{x}) \sim \mathcal{N}(0, \xi_h).\end{equation}
Quantity $\xi$ is called a two-point (real-space) correlation function and is defined as\begin{equation}    \langle\delta(\mathbf{x})\delta(\mathbf{x'})\rangle = \xi(|\mathbf{x}-\mathbf{x'}|).\end{equation}This quantity is sufficiently small when $|\mathbf{x}-\mathbf{x'}|\gg1$. We are asked to find what is the expression for $b$ in the equation $\delta_h(k) = b\delta_m(\mathbf{k})$, given the real-space expression for the number density $n_h(\mathbf{x})$ in terms of real-space sample of $\delta_m(\mathbf{x})$.\\

\textbf{Devise a plan.} The key point to solve this problem should be that real-space correlation function for halos $\xi_h$ should also be equal to $b^2\xi_m$. We want to calculate that correlation function. It should be expressed in terms of $\langle n(\mathbf{x})\rangle$ and $\langle n_h(\mathbf{x})n_h(\mathbf{x'})\rangle$. We expect to be able to calculate these expectations since they are the expectations of functions of the Gaussian random variables. We are given the pixel variance $\sigma$. How does it connect to the other quantities we know? In principle, that's also the part of domain knowledge but it also can be deducted from the definitions already given. A discretized version of the correlation function is \begin{equation}    \xi_{ij} = \langle\delta_{\mathbf{x_i}}\delta_{\mathbf{x_j}}\rangle.\end{equation}
When $i=j$, it becomes the pixel variance $\sigma$. \textit{Aside, we could have given instead of $\sigma$, the quantity $P_{mm}(k)$, that is a common description of a cosmological dark-matter field. In that case, from the definitions of $\xi(r)$ and $P_{mm}(k)$, we could have deduced that $\sigma = \frac{1}{V}\sum_{k}P_{mm}(k)$}. Then we pick the ensemble of all the pixels at given fixed large distance $r=|\mathbf{x_i}-\mathbf{x_j}|$. The key is to recognize that it is fully described by a correlated bivariate Gaussian distribution. \begin{equation}    (\delta^m_{i},\delta^m_{j}) \sim \mathcal{N}(0,\Sigma) \end{equation}with a covariance\begin{equation}    \Sigma = \begin{pmatrix}        \sigma^2 & \xi^m_r  \\      \xi^m_r & \sigma^2    \end{pmatrix}.\end{equation}In general, the integrals from the expectation values are cumbersome, but we should expect some simplifications from the fact that $\xi$ is small and we can Taylor-expand the pdf. \\

\textbf{Carry out the plan.} It's more convenient to define $\hat{\delta}_{i} = \delta^m_i/\sigma$ and $\hat{\xi} = \xi^m_r/\sigma^2$, and $\phi_2$ - a correlated bivariate Gaussian pdf - then \begin{equation}    (\hat{\delta}_{i},\hat{\delta}_{j}) \sim \frac{e^{-\frac{1}{2(1-\hat{\xi}^2)}[\hat{\delta}^2_{i}+\hat{\delta}^2_{j}-2\hat{\xi}\hat{\delta}_{i}\hat{\delta}_{j}]}}{2\pi\sqrt{1-\hat{\xi}^2}} \equiv \phi_2(\hat{\delta}_{i},\hat{\delta}_{j}|\hat{\xi}).\end{equation}We note that \begin{equation}    \xi^{n}_r = \frac{\langle n_in_j\rangle}{\langle n \rangle^2}-1.\end{equation}The quantity $\langle n\rangle$ is the actual mean number density:\begin{equation*}    \bar{n}^{'} = \langle n \rangle = \langle n_i \rangle = \int n^{loc}(\delta_i,b,\bar{n}) \phi_2(\hat{\delta}_{i},\hat{\delta}_{j}|\hat{\xi}) d\hat{\delta}_{i}d\hat{\delta}_{j} = \int n^{loc}_i \phi_1(\hat{\delta_i})d\hat{\delta_i}.\end{equation*}Here, $\phi_1$ - is a standard normal pdf. It is expected that it's not dependent on the correlation $\hat{\xi}$, but only on $b$ and $\sigma$, just as the marginal of 2D correlated Gaussian distribution is 1D Gaussian that's not dependent on the cross-correlation. To the linear order in $\hat{\xi}$, \begin{equation}    \phi_2(x,y|\hat{\xi}) \approx \phi_1(x)\phi_1(y)(1+\hat{\xi}xy).\end{equation} So that the two-point function neatly factorizes:\begin{align*}    &\langle n_in_j\rangle = \int n^{loc}(\delta_i,b,\bar{n})n^{loc}(\delta_j,b,\bar{n}) \phi_2(\hat{\delta}_{i},\hat{\delta}_{j}|\hat{\xi}) d\hat{\delta}_{i}d\hat{\delta}_{j} \\ &\approx \int n^{loc}_i \phi_1(\hat{\delta_i})d\hat{\delta_i}\int n^{loc}_j \phi_1(\hat{\delta_j})d\hat{\delta_j} + \hat{\xi}\int n^{loc}_i \phi_1(\hat{\delta_i})\hat{\delta_i}d\hat{\delta_i}\int n^{loc}_j \phi_1(\hat{\delta_j})\hat{\delta_j}d\hat{\delta_j} \\ &\equiv \langle n\rangle^2 + \hat{\xi} \langle n\hat{\delta }\rangle^2.\numberthis\end{align*}
Substituting the results for $\langle n\rangle$ and $\langle n_in_j\rangle$ in  the equation for $\xi^n_r$, we can read off the bias:\begin{equation}b^{'2} = \frac{\xi^n_r}{\sigma^2\hat{\xi}} = \frac{\langle n\hat{\delta}\rangle^2}{\sigma^2\langle n\rangle^2}.\end{equation}
All that is left is to calculate the expectations. One can evaluate for $b\geq0$ 
\begin{align*}    
\langle n\rangle &= \int n^{loc}_i \phi_1(\hat{\delta_i})d\hat{\delta_i} = \int \bar{n}\max(0, 1+b\sigma x)\phi_1(x)dx \\  &= \bar{n}\int_{-\frac{1}{b\sigma}}^{+\infty}(1+b\sigma x)\phi_1(x)dx = \bar{n}\left[\Phi_1\left(\frac{1}{b\sigma}\right)+b\sigma\phi_1\left(\frac{1}{b\sigma}\right)\right].\numberthis \end{align*}
For $b<0$ it's, however, 
\begin{align*}    
\langle n\rangle &= \bar{n}\int_{-\infty}^{+\frac{1}{|b|\sigma}}(1-|b|\sigma x)\phi_1(x)dx\\ &= \bar{n}\left[\Phi_1\left(\frac{1}{|b|\sigma}\right)+|b|\sigma\phi_1\left(\frac{1}{|b|\sigma}\right)\right].\numberthis\end{align*}So we conclude that the latter expression is valid for all $b$. Similarly, one can show that\begin{equation}\langle n\hat{\delta }\rangle = \bar{n}\int\max(0,1+b\sigma x)x\phi_1(x)dx = \bar{n}b\sigma \Phi_1\left(\frac{1}{|b|\sigma}\right)\end{equation}
where $\Phi_1(x) = \int_{-\infty}^x\phi_1(x)dx$ - normal cdf. Finally, one can get
\begin{equation}\boxed{b^{'} = \frac{b \Phi_1\left(\frac{1}{|b|\sigma}\right)}{\Phi_1\left(\frac{1}{|b|\sigma}\right)+|b|\sigma\phi_1\left(\frac{1}{|b|\sigma}\right)}}.\end{equation}

Note: We also accept solutions as correct if they omit the $|~|$ around the bias, since halo bias is usually positive.

\clearpage

\subsection{Level 4 - SHO Vacuum Entanglement}

\subsubsection*{Problem Statement}

Consider a coupled simple harmonic oscillator governed by the Hamiltonian
\begin{align}
H & =\sum_{i=1}^{2}\frac{1}{2}\left(\frac{p_{i}^{2}}{m}+kx_{i}^{2}\right)+g\frac{(x_{1}-x_{2})^{2}}{2}.
\end{align}
If the ground state is $|\Omega\rangle$ and the operator $\hat{\rho}$
is the vacuum density matrix partially traced over the $|w\rangle_{x_{2}}$
components (satisfying $\hat{x}_{2}|w\rangle_{x_{2}}=w|w\rangle_{x_{2}}$),
i.e.
\begin{equation}
\hat{\rho}\equiv\int dx_{1}''\int dx_{1}'\int dw\left(|x_{1}''\rangle_{x_{1}}\,_{x_{1}}\langle x_{1}''|\otimes\,_{x_{2}}\langle w|\right)\left(|\Omega\rangle\langle\Omega|\right)\left(|x_{1}'\rangle_{x_{1}}\otimes|w\rangle_{x_{2}}\,_{x_{1}}\langle x_{1}'|\right)
\end{equation}
which is an operator acting on a reduced Hilbert space, compute 
\begin{equation}
S\equiv-\mathrm{Tr}_{x_{1}}\left[\hat{\rho}\ln\hat{\rho}\right]
\end{equation}
which involves the trace over $x_{1}$ states.

\subsubsection*{Answer Requirements}
Provide the answer in the form of the \texttt{python} code. Implement the following function 
\begin{python}
def entropy(k:float,g:float,m:float)->float:
    pass
\end{python}

\subsubsection*{Comments about the Problem}

\textit{This problem, whose solution can be found in \cite{Srednicki:1993im} (with less detailed reasoning steps), has been rephrased in a pedagogical manner for graduate-level physics courses. It is a well-known question in quantum entanglement research, and the best performing LLMs are capable of solving it accurately, perhaps at least partially due to memorization.}

\subsubsection*{Solution}
Diagonalize the original Hamiltonian 
\begin{align}
H & =(x_{1}\quad x_{2}\quad p_{1}\quad p_{2})\left(\begin{array}{cccc}
\frac{k+g}{2} & -\frac{g}{2}\\
-\frac{g}{2} & \frac{k+g}{2}\\
 &  & \frac{1}{2m}\\
 &  &  & \frac{1}{2m}
\end{array}\right)\left(\begin{array}{c}
x_{1}\\
x_{2}\\
p_{1}\\
p_{2}
\end{array}\right).
\end{align}
One easily finds
\begin{equation}
x_{1}=\frac{y_{1}+y_{2}}{\sqrt{2}}\label{eq:x1}
\end{equation}
\begin{equation}
x_{2}=\frac{y_{1}-y_{2}}{\sqrt{2}}\label{eq:x2}
\end{equation}
diagonalizes the Hamiltonian such that in the $(y_{1},y_{2},q_{1}\equiv m\dot{y}_{1},q_{2}\equiv m\dot{y}_{2})$
basis, it is
\begin{align}
H & =(y_{1}\quad y_{2}\quad q_{1}\quad q_{2})\left(\begin{array}{cccc}
\frac{k}{2} & 0\\
0 & \frac{k}{2}+g\\
 &  & \frac{1}{2m}\\
 &  &  & \frac{1}{2m}
\end{array}\right)\left(\begin{array}{c}
y_{1}\\
y_{2}\\
q_{1}\\
q_{2}
\end{array}\right).
\end{align}
The ladder operators are
\begin{equation}
a_{j}=\frac{1}{\sqrt{2}}\left(\sqrt{m\omega_{j}}y_{j}+\frac{i}{\sqrt{m\omega_{j}}}q_{j}\right)
\end{equation}
\begin{equation}
\omega_{1}^{2}=\frac{k}{m}\,\,\,\,\,\,\,\,\omega_{2}^{2}=\frac{k+2g}{m}
\end{equation}
which allows one to rewrite the Hamiltonian as
\begin{equation}
H=\sum_{j=1}^{2}a_{j}^{\dagger}a_{j}\omega_{j}+\frac{\omega_{1}+\omega_{2}}{2}.
\end{equation}
In this basis, we denote the ground state as
\begin{equation}
a_{1}|00\rangle_{\vec{n}_{y}}=0=a_{2}|00\rangle_{\vec{n}_{y}}.
\end{equation}
Hence we have found $|\Omega\rangle=|00\rangle_{\vec{n}_{y}}$ .
We know that the wave function in the $\vec{y}$ coordinates is the
product of well known simple harmonic oscillator solutions:
\begin{equation}
\langle y'_{1},y'_{2}|00\rangle_{\vec{n}_{y}}=\frac{1}{\left(\pi b_{1}^{2}\right)^{1/4}}\exp\left[\frac{-\left(y_{1}'\right)^{2}}{2b_{1}^{2}}\right]\frac{1}{\left(\pi b_{2}^{2}\right)^{1/4}}\exp\left[\frac{-\left(y_{2}'\right)^{2}}{2b_{2}^{2}}\right]
\end{equation}
where
\begin{equation}
b_{n}\equiv\frac{1}{\sqrt{m\omega_{n}}}
\end{equation}
making this a convenient basis to work with. Note
\begin{align*}
\hat{y}_{1}\left(\left|a\right\rangle _{x_{1}}\otimes\left|b\right\rangle _{x_{2}}\right) & =\int dy_{1}'dy_{2}'\hat{y}_{1}|y_{1}'y_{2}'\rangle\langle y_{1}'y_{2}'|\left(\left|a\right\rangle _{x_{1}}\otimes\left|b\right\rangle _{x_{2}}\right)\\
 & =\int dy_{1}'dy_{2}'y_{1}'|y_{1}'y_{2}'\rangle\langle y_{1}'y_{2}'|\left(\left|a\right\rangle _{x_{1}}\otimes\left|b\right\rangle _{x_{2}}\right)\\
 & =\int dx_{1}'dx_{2}'y_{1}'\left(\left|x_{1}'\right\rangle _{x_{1}}\otimes\left|x_{2}'\right\rangle _{x_{2}}\right)\left(\,_{x_{1}}\langle x_{1}'|\otimes\,_{x_{2}}\langle x_{2}'|\right)\left(\left|a\right\rangle _{x_{1}}\otimes\left|b\right\rangle _{x_{2}}\right)\\
 & =\frac{a+b}{\sqrt{2}}\left(\left|a\right\rangle _{x_{1}}\otimes\left|b\right\rangle _{x_{2}}\right)\numberthis
\end{align*}
where we used the completeness of the basis, Eqs.~(\ref{eq:x1})
and (\ref{eq:x2}), and the usual delta function normalization of
the position basis. This and a similar relation for $\hat{y}_{2}$
imply
\begin{equation}
\left|a\right\rangle _{x_{1}}\otimes\left|b\right\rangle _{x_{2}}=\left|\frac{a+b}{\sqrt{2}},\frac{a-b}{\sqrt{2}}\right\rangle .
\end{equation}
This means
\begin{equation}
\,_{\vec{n}_{y}}\langle00|\left(|x_{1}'\rangle_{x_{1}}\otimes|w\rangle_{x_{2}}\right)=\,_{\vec{n}_{y}}\langle00|\frac{x_{1}'+w}{\sqrt{2}},\frac{x_{1}'-w}{\sqrt{2}}\rangle=\frac{1}{\left(\pi b_{1}^{2}\right)^{1/4}}\exp\left[\frac{-\left(\frac{x_{1}'+w}{\sqrt{2}}\right)^{2}}{2b_{1}^{2}}\right]\frac{1}{\left(\pi b_{2}^{2}\right)^{1/4}}\exp\left[\frac{-\left(\frac{x_{1}'-w}{\sqrt{2}}\right)^{2}}{2b_{2}^{2}}\right].
\end{equation}
The partial trace is defined through the following contraction of
$(2,2)$ tensor to a $(1,1)$ tensor: 
\begin{align*}
\hat{\rho} & =\int dx_{1}''\int dx_{1}'\int dw\left(|x_{1}''\rangle_{x_{1}}\,_{x_{1}}\langle x_{1}''|\otimes\,_{x_{2}}\langle w|\right)\left(|00\rangle_{\vec{n}_{y}}\,_{\vec{n}_{y}}\langle00|\right)\left(|x_{1}'\rangle_{x_{1}}\otimes|w\rangle_{x_{2}}\,_{x_{1}}\langle x_{1}'|\right)\\
 & =\int dx_{1}''\int dx_{1}'\int dw|x_{1}''\rangle_{x_{1}}\,_{x_{1}}\langle x_{1}'|\frac{1}{\left(\pi b_{1}^{2}\right)^{1/4}}\exp\left[\frac{-\left(\frac{x_{1}''+w}{\sqrt{2}}\right)^{2}}{2b_{1}^{2}}\right]\frac{1}{\left(\pi b_{2}^{2}\right)^{1/4}}\exp\left[\frac{-\left(\frac{x_{1}''-w}{\sqrt{2}}\right)^{2}}{2b_{2}^{2}}\right]\times\nonumber \\
 & \frac{1}{\left(\pi b_{1}^{2}\right)^{1/4}}\exp\left[\frac{-\left(\frac{x_{1}'+w}{\sqrt{2}}\right)^{2}}{2b_{1}^{2}}\right]\frac{1}{\left(\pi b_{2}^{2}\right)^{1/4}}\exp\left[\frac{-\left(\frac{x_{1}'-w}{\sqrt{2}}\right)^{2}}{2b_{2}^{2}}\right].\numberthis
\end{align*}
Integrate over $w$, we find
\begin{align*}
\hat{\rho} & =\int dx_{1}''\int dx_{1}'|x_{1}''\rangle_{x_{1}}\,_{x_{1}}\langle x_{1}'|\frac{1}{\left(\pi b_{1}^{2}\right)^{1/2}}\frac{1}{\left(\pi b_{2}^{2}\right)^{1/2}}\exp\left[-\frac{m}{4}\left(\omega_{1}+\omega_{2}\right)\left(\left[x_{1}'\right]^{2}+\left[x_{1}''\right]^{2}\right)\right]\times\nonumber \\
 & \frac{\sqrt{2\pi}}{\sqrt{m\left[\omega_{1}+\omega_{2}\right]}}\exp\left[\frac{(\frac{\sqrt{\omega_{2}}}{\sqrt{\omega_{1}}}-\frac{\sqrt{\omega_{1}}}{\sqrt{\omega_{2}}})^{2}(x_{1}'+x_{1}'')^{2}}{8\frac{1}{m}(\frac{1}{\omega_{1}}+\frac{1}{\omega_{2}})}\right]\\
 & =\int dx_{1}''\int dx_{1}'|x_{1}''\rangle_{x_{1}}\,_{x_{1}}\langle x_{1}'|\frac{1}{\left(\pi b_{1}^{2}\right)^{1/2}}\frac{1}{\left(\pi b_{2}^{2}\right)^{1/2}}\times\nonumber \\
 & \frac{\sqrt{2\pi}}{\sqrt{m\left[\omega_{1}+\omega_{2}\right]}}\exp\left[\frac{m(\omega_{2}-\omega_{1})^{2}2x_{1}'x_{1}''-m\left[8\omega_{1}\omega_{2}+(\omega_{1}-\omega_{2})^{2}\right]\left(\left[x_{1}'\right]^{2}+\left[x_{1}''\right]^{2}\right)}{8(\omega_{1}+\omega_{2})}\right].\numberthis
\end{align*}
Next, to identify the matrix, use
\begin{equation}
\frac{m(\omega_{2}-\omega_{1})^{2}2x_{1}'x_{1}''-m\left[8\omega_{1}\omega_{2}+(\omega_{1}-\omega_{2})^{2}\right]\left(\left[x_{1}'\right]^{2}+\left[x_{1}''\right]^{2}\right)}{8(\omega_{1}+\omega_{2})}=-\frac{1}{2b^{2}}\left[\left(\left[x_{1}'\right]^{2}+\left[x_{1}''\right]^{2}\right)-2\frac{(\omega_{2}-\omega_{1})^{2}}{\gamma}x_{1}'x_{1}''\right]
\end{equation}
\begin{equation}
\gamma\equiv8\omega_{1}\omega_{2}+(\omega_{1}-\omega_{2})^{2}
\end{equation}
\begin{equation}
\frac{1}{2b^{2}}\equiv\frac{m\gamma}{8(\omega_{1}+\omega_{2})}
\end{equation}
\begin{equation}
b=2\sqrt{\frac{\omega_{1}+\omega_{2}}{m[8\omega_{1}\omega_{2}+(\omega_{1}-\omega_{2})^{2}]}}
\end{equation}
to write
\begin{align}
\hat{\rho} & =\int dx_{1}''\int dx_{1}'|x_{1}''\rangle_{x_{1}}\,_{x_{1}}\langle x_{1}'|\frac{1}{\left(\pi b_{1}^{2}\right)^{1/2}}\frac{1}{\left(\pi b_{2}^{2}\right)^{1/2}} \times \nonumber\\
 & \frac{\sqrt{2\pi}}{\sqrt{m\left[\omega_{1}+\omega_{2}\right]}}\exp\left[-\frac{1}{2b^{2}}\left(\left[x_{1}'\right]^{2}+\left[x_{1}''\right]^{2}\right)\right]\exp\left(\frac{(\omega_{2}-\omega_{1})^{2}}{\gamma b^{2}}x_{1}'x_{1}''\right).
\end{align}
Change basis to energy with a new effective frequency
\begin{equation}
b_{3}=\frac{1}{\sqrt{m\omega_{3}}}
\end{equation}
\begin{equation}
\hat{\rho}=\sum_{nv}|v\rangle\langle v|\hat{\rho}|n\rangle\langle n|
\end{equation}
\begin{align}
\langle v|\hat{\rho}|n\rangle & =\int dx_{1}''\int dx_{1}'\langle v|x_{1}''\rangle_{x_{1}}\,_{x_{1}}\langle x_{1}'|n\rangle\frac{1}{\left(\pi b_{1}^{2}\right)^{1/2}}\frac{1}{\left(\pi b_{2}^{2}\right)^{1/2}}\times\nonumber \\
 & \frac{\sqrt{2\pi}}{\sqrt{m\left[\omega_{1}+\omega_{2}\right]}}\exp\left[-\frac{1}{2b^{2}}\left(\left[x_{1}'\right]^{2}+\left[x_{1}''\right]^{2}\right)\right]\exp\left(\frac{(\omega_{2}-\omega_{1})^{2}}{\gamma b^{2}}x_{1}'x_{1}''\right)
\end{align}
where
\begin{equation}
\langle x_{1}'|n\rangle=\frac{1}{\sqrt{n!b_{3}\sqrt{\pi}2^{n}}}e^{\frac{-\left(x_{1}'\right)^{2}}{2b_{3}^{2}}}H_{n}\left(\frac{x_{1}'}{b_{3}}\right)
\end{equation}
are the well known oscillator wave functions and $b_{3}$ still has
to be chosen. One can show by carrying out the integrals that the
matrix is diagonalized if 
\begin{align*}
b_{3} & =\frac{b}{\left(1-b^{4}\left[\frac{(\omega_{2}-\omega_{1})^{2}}{\gamma b^{2}}\right]^{2}\right)^{1/4}}\\
 & =\frac{1}{\sqrt{m}\omega_{1}^{1/4}\omega_{2}^{1/4}}.\numberthis
\end{align*}
This gives
\[
\langle v|\hat{\rho}|n\rangle=\lambda_{n}\delta_{vn}
\]
 where
\begin{align*}
\lambda_{n} & =\frac{\sqrt{2\pi}}{\sqrt{m\left[\omega_{1}+\omega_{2}\right]}}\frac{1}{\left(\pi b_{1}^{2}\right)^{1/2}}\frac{1}{\left(\pi b_{2}^{2}\right)^{1/2}}m_{11}\left(\frac{b^{2}\frac{(\omega_{2}-\omega_{1})^{2}}{\gamma b^{2}}}{1+\sqrt{1-b^{4}\left[\frac{(\omega_{2}-\omega_{1})^{2}}{\gamma b^{2}}\right]^{2}}}\right)^{n-1}\\
 & =\frac{\pi\sqrt{m}}{2\left[\omega_{1}+\omega_{2}\right]^{3/2}}\frac{1}{\left(\pi b_{1}^{2}\right)^{1/2}}\frac{1}{\left(\pi b_{2}^{2}\right)^{1/2}}\frac{(\omega_{2}-\omega_{1})^{2}}{\left(\sqrt{m}\omega_{1}^{1/4}\omega_{2}^{1/4}\right)^{3}\left(\frac{b_{3}^{2}}{b^{2}}+1\right)^{3/2}}\left(\frac{\frac{(\omega_{2}-\omega_{1})^{2}}{8\omega_{1}\omega_{2}+(\omega_{1}-\omega_{2})^{2}}}{1+\frac{b^{2}}{b_{3}^{2}}}\right)^{n-1}\numberthis
\end{align*}
where we used
\begin{align*}
m_{11} & =\frac{b^{3}\frac{(\omega_{2}-\omega_{1})^{2}}{\gamma b^{2}}\sqrt{2\pi}}{\left(1+\sqrt{1-b^{4}\left(\frac{(\omega_{2}-\omega_{1})^{2}}{\gamma b^{2}}\right)^{2}}\right)^{3/2}}\\
 & =\frac{m(\omega_{2}-\omega_{1})^{2}\sqrt{2\pi}}{4(\omega_{1}+\omega_{2})\left(\frac{1}{b^{2}}+\frac{1}{b_{3}^{2}}\right)^{3/2}}\numberthis
\end{align*}
\begin{align*}
\left(\frac{b_{3}}{b}\right)^{2} & =\frac{1}{m\omega_{1}^{1/2}\omega_{2}^{1/2}}\frac{1}{4\frac{\omega_{1}+\omega_{2}}{m[8\omega_{1}\omega_{2}+(\omega_{1}-\omega_{2})^{2}]}}\\
 & =\frac{1}{\omega_{1}^{1/2}\omega_{2}^{1/2}}\frac{8\omega_{1}\omega_{2}+(\omega_{1}-\omega_{2})^{2}}{4\left(\omega_{1}+\omega_{2}\right)}.\numberthis
\end{align*}
Simplify:
\begin{align*}
\lambda_{n} & =\frac{4\sqrt{\omega_{1}\omega_{2}}}{\sqrt{8\omega_{1}\omega_{2}+(\omega_{1}-\omega_{2})^{2}+4\omega_{1}^{1/2}\omega_{2}^{1/2}\left(\omega_{1}+\omega_{2}\right)}}\left(\frac{(\omega_{2}-\omega_{1})^{2}}{8\omega_{1}\omega_{2}+(\omega_{1}-\omega_{2})^{2}+\omega_{1}^{1/2}\omega_{2}^{1/2}4\left(\omega_{1}+\omega_{2}\right)}\right)^{n}\\
 & =\frac{4\sqrt{\omega_{1}\omega_{2}}}{\left(\sqrt{\omega_{1}}+\sqrt{\omega_{2}}\right)^{2}}\left[\frac{(\omega_{1}-\omega_{2})^{2}}{\left(\sqrt{\omega_{1}}+\sqrt{\omega_{2}}\right)^{4}}\right]^{n}.\numberthis
\end{align*}
Since we want to evaluate
\begin{equation}
-\mathrm{Tr}\left[\hat{\rho}\ln\hat{\rho}\right]=-\partial_{n}\ln\mathrm{tr}\hat{\rho}^{n}|_{n=1}
\end{equation}
we compute
\begin{align*}
\ln\mathrm{tr}\rho^{n} & =\ln\left(\sum_{j=0}^{\infty}\lambda_{j}^{n}\right)\\
 & =\ln\left(\sum_{j}\left[\frac{4\sqrt{\omega_{1}\omega_{2}}}{\left(\sqrt{\omega_{1}}+\sqrt{\omega_{2}}\right)^{2}}\left[\frac{(\omega_{1}-\omega_{2})^{2}}{\left(\sqrt{\omega_{1}}+\sqrt{\omega_{2}}\right)^{4}}\right]^{j}\right]^{n}\right)\\
 & =n\ln\left[\frac{4\sqrt{\omega_{1}\omega_{2}}}{\left(\sqrt{\omega_{1}}+\sqrt{\omega_{2}}\right)^{2}}\right]+\ln\left(\sum_{j}\left[\frac{(\omega_{1}-\omega_{2})^{2}}{\left(\sqrt{\omega_{1}}+\sqrt{\omega_{2}}\right)^{4}}\right]^{nj}\right)\\
 & =n\ln\left[\frac{4\sqrt{\omega_{1}\omega_{2}}}{\left(\sqrt{\omega_{1}}+\sqrt{\omega_{2}}\right)^{2}}\right]-\ln\left(1-\left[\frac{(\omega_{1}-\omega_{2})^{2}}{\left(\sqrt{\omega_{1}}+\sqrt{\omega_{2}}\right)^{4}}\right]^{n}\right).\numberthis
\end{align*}
Hence, we arrive at
\begin{align*}
S & =-\left\{ \ln\left[\frac{4\sqrt{\omega_{1}\omega_{2}}}{\left(\sqrt{\omega_{1}}+\sqrt{\omega_{2}}\right)^{2}}\right]-\frac{\left[\frac{(\omega_{1}-\omega_{2})^{2}}{\left(\sqrt{\omega_{1}}+\sqrt{\omega_{2}}\right)^{4}}\right]\ln\left[\frac{(\omega_{1}-\omega_{2})^{2}}{\left(\sqrt{\omega_{1}}+\sqrt{\omega_{2}}\right)^{4}}\right]}{\left(1-\left[\frac{(\omega_{1}-\omega_{2})^{2}}{\left(\sqrt{\omega_{1}}+\sqrt{\omega_{2}}\right)^{4}}\right]\right)}\right\} \\
 & =\boxed{-\ln\left(\frac{4\sqrt{\omega_{1}\omega_{2}}}{\left(\sqrt{\omega_{1}}+\sqrt{\omega_{2}}\right)^{2}}\right)-\left(\frac{(\omega_{2}-\omega_{1})^{2}}{4\sqrt{\omega_{1}\omega_{2}}\left(\sqrt{\omega_{1}}+\sqrt{\omega_{2}}\right)^{2}}\right)\ln\left(\frac{(\omega_{2}-\omega_{1})^{2}}{\left(\sqrt{\omega_{1}}+\sqrt{\omega_{2}}\right)^{4}}\right)}\numberthis
\end{align*}
where
\begin{equation}
\boxed{\omega_{1}=\sqrt{\frac{k}{m}}\,\,\,\,\,\,\,\,\omega_{2}=\sqrt{\frac{k+2g}{m}}}.
\end{equation}

\clearpage

\subsection{Level 4 - SUSY-Symmetry}\label{L4-susy}

\subsubsection*{Problem Statement}

Consider the theory
\begin{equation}
\mathcal{L}=i\bar{\xi}\bar{\sigma}^{\mu}\partial_{\mu}\xi+|\partial\phi|^{2}-|F|^{2}
\end{equation}
where $\xi$ is a 2-component Weyl spinor while $\phi$ and $F$ are
complex scalar fields. Suppose you want to make the following infinitesimal
transformation a symmetry of this theory:
\begin{equation}
\delta_{\eta}\xi_{\alpha}=i\sqrt{2}\sigma_{\alpha\dot{\alpha}}^{\mu}\bar{\eta}^{\dot{\alpha}}\partial_{\mu}\phi+\sqrt{2}\eta_{\alpha}F
\end{equation}
\begin{align*}
\delta_{\eta}\bar{\xi}_{\dot{\beta}} & = [i\sqrt{2}\sigma_{\beta\dot{\alpha}}^{\mu}\bar{\eta}^{\dot{\alpha}}\partial_{\mu}\phi+\sqrt{2}\eta_{\beta}F]^{\dagger}\\
 & = -i\sqrt{2}(\bar{\eta}^{\dot{\alpha}}\sigma_{\dot{\alpha}\beta}^{\mu*})^{*}\partial_{\mu}\bar{\phi}+\sqrt{2}\bar{\eta}_{\dot{\beta}}\bar{F}\\
 & = -i\sqrt{2}\eta^{\alpha}\sigma_{\alpha\dot{\beta}}^{\mu}\partial_{\mu}\bar{\phi}+\sqrt{2}\bar{\eta}_{\dot{\beta}}\bar{F}\numberthis
\end{align*}
\begin{equation}
\delta_{\eta}F=i\sqrt{2}\bar{\eta}_{\dot{\alpha}}\bar{\sigma}^{\mu\dot{\alpha}\alpha}\partial_{\mu}\xi_{\alpha}=i\sqrt{2}\bar{\eta}\bar{\sigma}^{\mu}\partial_{\mu}\xi
\end{equation}
\begin{align*}
\delta_{\eta}\bar{F} & = -i\sqrt{2}(\bar{\eta}\bar{\sigma}^{\mu}\partial_{\mu}\xi)^{\dagger}\\
 & = -i\sqrt{2}(\partial_{\mu}\xi)^{\dagger}(\bar{\sigma}^{\mu})^{\dagger}(\bar{\eta})^{\dagger}\\
 & = -i\sqrt{2}\partial_{\mu}\bar{\xi}\bar{\sigma}^{\mu}\eta\numberthis
\end{align*}
along with $\delta_{\eta}\phi$ and $\left(\delta_{\eta}\phi\right)^{\dagger}$
where $\eta$ is a spacetime-independent infinitesimal fermionic parameter
inducing the transformation. Find the transformation rule $\delta_{\eta}\phi$
and $\left(\delta_{\eta}\phi\right)^{\dagger}$ for the action associated
with $\mathcal{L}$ to remain invariant.

\subsubsection*{Answer Requirements}
Provide the answer in the form of the \texttt{python} code. Implement the following function 
\begin{python}
from math import sqrt
def find_delta_phi(eta:float, xi:float, bar_eta:float, bar_xi:float) -> Tuple[float, float]:
    """
    Returns the SUSY transformation rules for phi and its Hermitian conjugate: 
    a tuple (delta_phi, delta_phi_dagger)
    """
    pass
\end{python}

\subsubsection*{Comments about the Problem}
\textit{This problem is situated in advanced quantum field theory within the framework of supersymmetry (SUSY). It involves analyzing how bosonic and fermionic fields transform under an infinitesimal SUSY transformation and requires knowledge and careful application of Grassmann variables and the associated algebra. Such topics are typically encountered in advanced graduate-level physics courses.  Note that the Hermiticity of $\sigma^\mu$ matrix convention as well as the metric convention of $(1,-1,-1,-1)$ is implicit in the statement of the problem (the latter inherent in the kinetic minus the potential form of the Lagrangian).}

\subsubsection*{Solution}
Denoting the variation $\left(\delta_{\eta}\phi\right)^{\dagger}$
as $\delta_{\eta}\bar{\phi}$, we write 
\begin{align*}
\delta_{\eta}\mathcal{L} & = i\delta_{\eta}\bar{\xi}\bar{\sigma}^{\mu}\partial_{\mu}\xi+i\bar{\xi}\bar{\sigma}^{\mu}\partial_{\mu}\delta_{\eta}\xi+\partial_{\mu}\delta_{\eta}\bar{\phi}\partial^{\mu}\phi+\partial_{\mu}\bar{\phi}\partial^{\mu}\delta_{\eta}\phi-\delta_{\eta}\bar{F}F-\bar{F}\delta_{\eta}F\\
 & = i[-i\sqrt{2}\eta\sigma^{\beta}\partial_{\beta}\bar{\phi}+\cancel{\sqrt{2}\bar{\eta}\bar{F}}]\bar{\sigma}^{\mu}\partial_{\mu}\xi+i\bar{\xi}\bar{\sigma}^{\mu}\partial_{\mu}[i\sqrt{2}\sigma^{\beta}\bar{\eta}\partial_{\beta}\phi+\sqrt{2}\eta F]\\
 & +\partial_{\mu}\delta_{\eta}\bar{\phi}\partial^{\mu}\phi+\partial_{\mu}\bar{\phi}\partial^{\mu}\delta_{\eta}\phi-[-i\sqrt{2}\partial_{\mu}\bar{\xi}\bar{\sigma}^{\mu}\eta]F-\cancel{\bar{F}[i\sqrt{2}\bar{\eta}\bar{\sigma}^{\mu}\partial_{\mu}\xi]}\,.\numberthis
\end{align*}
Integrating by parts, we find (denoting with equality an equivalence
up to total derivative terms)
\begin{align}
\delta_{\eta}\mathcal{L} & =\sqrt{2}\eta\sigma^{\beta}\partial_{\beta}\bar{\phi}\bar{\sigma}^{\mu}\partial_{\mu}\xi+\partial_{\mu}\bar{\xi}\bar{\sigma}^{\mu}[\sqrt{2}\sigma^{\beta}\bar{\eta}\partial_{\beta}\phi-\cancel{i\sqrt{2}\eta F}]\nonumber \\
 & +\partial_{\mu}\delta_{\eta}\bar{\phi}\partial^{\mu}\phi+\partial_{\mu}\bar{\phi}\partial^{\mu}\delta_{\eta}\phi+\cancel{i\sqrt{2}\partial_{\mu}\bar{\xi}\bar{\sigma}^{\mu}\eta F} \,.
\end{align}
Integrate by parts the first two terms to eliminate the the $\sigma$
matrices using the identity $\bar{\sigma}^{\mu}\sigma^{\nu}+\bar{\sigma}^{\nu}\sigma^{\mu}=2g^{\mu\nu}$:
\begin{align}
\delta_{\eta}\mathcal{L} & =\sqrt{2}\left(\eta\partial_{\mu}\bar{\phi}\partial^{\mu}\xi+\partial^{\mu}\bar{\xi}\bar{\eta}\partial_{\mu}\phi\right)\nonumber \\
 & +\partial_{\mu}\delta_{\eta}\bar{\phi}\partial^{\mu}\phi+\partial_{\mu}\bar{\phi}\partial^{\mu}\delta_{\eta}\phi
\end{align}
again denoting with equality an equivalence up to total derivative
terms, and we are using the standard notation $\eta\xi\equiv\eta^{\alpha}\xi_{\alpha}$
and $\bar{\xi}_{\dot{\alpha}}\bar{\eta}^{\dot{\alpha}}\equiv\bar{\xi}\bar{\eta}$.
To make the remainder cancel, we solve
\begin{equation}
\sqrt{2}\eta\partial_{\mu}\bar{\phi}\partial^{\mu}\xi+\partial_{\mu}\bar{\phi}\partial^{\mu}\delta_{\eta}\phi=0
\end{equation}
yielding
\begin{equation}
\boxed{\delta_{\eta}\phi=-\sqrt{2}\eta\xi,\quad\left(\delta_{\eta}\phi\right)^{\dagger}=-\sqrt{2}\bar{\xi}\bar{\eta}}.\label{eq:L4-susy}
\end{equation}

\clearpage

\subsection{Level 3 - Slow-Roll Inflation}

\subsubsection*{Problem Statement}
For the action
\begin{equation}
S = \int dt a^3(t) \left\{ \frac{1}{2} \dot{\phi}^2 - V_0 \exp \left[ - \sqrt{\frac{2}{q}} \left( \frac{\phi}{M_P} \right) \right] \right\}
\end{equation}
where \( q \) and \(V_0\) are constants, derive and solve (integrate) the equation of motion for the field $\phi$ assuming slow-roll inflation and initial condition $\phi(t=0) = \phi_0$.

\subsubsection*{Answer Requirements}
Provide the answer in the form of the \texttt{python} code. Implement the following function 
\begin{python}
import numpy as np
def phi(q: float, M_p: float, phi_0: float, V_0: float, t: np.ndarray)->np.ndarray:
    pass
\end{python}

\subsubsection*{Comments about the Problem}
\textit{This problem lies in the field of cosmology, particularly in inflationary cosmology, and involves studying the dynamics of a scalar field (inflaton) driving the accelerated expansion of the early universe, before the ``hot Big Bang". It is typically encountered in specialized graduate-level courses in cosmology and requires familiarity with field theory in an expanding spacetime.}

\subsubsection*{Solution}
The equation of motion is
\begin{equation}
\ddot{\phi} + 3 H \dot{\phi} - \sqrt{\frac{2}{q}} \left( \frac{1}{M_P} \right) V_0 \exp \left[ - \sqrt{\frac{2}{q}} \left( \frac{\phi}{M_P} \right) \right] = 0.
\end{equation}
For the slow-roll inflation, the following must hold: 
\begin{equation}
    \ddot{\phi}\ll 3H\dot{\phi} \,.
\end{equation}
Hence, we have
\begin{equation}
3 H \dot{\phi} = \sqrt{\frac{2}{q}} \left( \frac{1}{M_P} \right) V_0 \exp \left[ - \sqrt{\frac{2}{q}} \left( \frac{\phi}{M_P} \right) \right].
\end{equation}
Slow-roll approximation also implies 
\begin{equation}
H^2\approx\frac{V(\phi)}{3M^2_P}
\end{equation}
so we need to solve the following ODE:
\begin{equation}
3 \sqrt{\frac{V_0 \exp \left[ - \sqrt{\frac{2}{q}} \left( \frac{\phi}{M_P} \right) \right]}{3 M_P^2}} \frac{d\phi}{dt} = \sqrt{\frac{2}{q}} \left( \frac{1}{M_P} \right) V_0 \exp \left[ - \sqrt{\frac{2}{q}} \left( \frac{\phi}{M_P} \right) \right]
\end{equation}

\begin{equation}
\int \frac{d\phi}{\sqrt{V_0}} \exp \left[ \sqrt{\frac{1}{2q}} \left( \frac{\phi}{M_P} \right) \right] = \sqrt{\frac{2}{3q}} t \,.
\end{equation}
Performing the integration and solving for $\phi(t)$ we get
\begin{equation}
\frac{1}{\sqrt{V_0}} M_P \sqrt{2q} \left( \exp \left[ \sqrt{\frac{1}{2q}} \left( \frac{\phi}{M_P} \right) \right] - \exp \left[ \sqrt{\frac{1}{2q}} \left( \frac{\phi_0}{M_P} \right) \right] \right) = \sqrt{\frac{2}{3q}} t
\end{equation}

\begin{equation}
\boxed{
\phi = \sqrt{2q} M_P \ln \left\{ \exp \left[ \sqrt{\frac{1}{2q}} \left( \frac{\phi_0}{M_P} \right) \right] + \frac{1}{M_P q} \sqrt{\frac{V_0}{3}} t \right\}}.
\end{equation}

\clearpage

\subsection{Level 3 - Scalar Particle Scattering}

\subsubsection*{Problem Statement}
Consider
\begin{equation}
\mathcal{L} = \left\{ \sum_{i=1}^2 \left[ \frac{1}{2} (\partial_\mu \phi_i)(\partial^\mu \phi_i) - \frac{m_i^2}{2} \phi_i \phi_i \right] - \frac{\lambda}{4} \phi_1^2 \phi_2^2 \right\}
\end{equation}
What is the differential cross section \( \frac{d\sigma}{d\Omega} \) for \( \phi_1 (\vec{k}_1) \phi_1 (-\vec{k}_1) \to \phi_2 (\vec{k}_1') \phi_2 (-\vec{k}_1') \) in the CM frame accurate to \( O(\lambda^2) \)? Express your final answer in terms of Mandelstam variables. 

\subsubsection*{Answer Requirements}
Provide the answer in the form of the \texttt{python} code. Implement the following function.
\begin{python}
def dsigma_domega(lam: float, s_m: float, p_m: float, u_m: float, 
                  m1: float, m2: float) -> float:
    pass
\end{python}

\subsubsection*{Comments about the Problem}
\textit{This is a question from quantum field theory. It involves calculating the differential cross section for a process where two \(\phi_1\) particles annihilate into two \(\phi_2\) particles. Such problems are typically encountered in graduate-level particle physics courses and require familiarity with perturbative field theory and the use of Mandelstam variables to express scattering amplitudes.}

\subsubsection*{Solution}
The amplitude for this process is
\begin{equation}
i \mathcal{M} = -4 i \frac{\lambda}{4} = -i \lambda
\end{equation}
In the CM frame, energy conservation gives
\begin{equation}
2 \sqrt{|\vec{k}_1|^2 + m_1^2} = 2 \sqrt{|\vec{k}_1'|^2 + m_2^2}
\end{equation}
A standard formula for differential cross section gives
\begin{align*}
\left( \frac{d\sigma}{d\Omega} \right)_{\text{CM}} &= \frac{1}{64 \pi^2 s} \frac{k_1'}{k_1} |\mathcal{M}|^2 \\
&= \frac{\lambda^2}{64 \pi^2 s} \frac{\sqrt{|\vec{k}_1|^2 + (m_1^2 - m_2^2)}}{k_1} \numberthis
\end{align*}
Since in the CM frame, we know
\begin{equation}
k_1 = \frac{1}{2 \sqrt{s}} \sqrt{s^2 - 4 m_1^2 s}
\end{equation}
\begin{align*}
\left( \frac{d\sigma}{d\Omega} \right)_{\text{CM}} &= \frac{2 \sqrt{s}}{64 \pi^2 s} \sqrt{\frac{1}{4s} \left[ s^2 - 4 m_1^2 s \right] + (m_1^2 - m_2^2)} \frac{\lambda^2}{\sqrt{s^2 - 4 m_1^2 s}} \\
&= \frac{\lambda^2}{64 \pi^2 s} \frac{\sqrt{s^2 - 4 m_1^2 s + 4s(m_1^2 - m_2^2)}}{\sqrt{s^2 - 4 m_1^2 s}}.\numberthis
\end{align*}

The final result is
\begin{equation}
\boxed{
\left( \frac{d\sigma}{d\Omega} \right)_{\text{CM}}
= \frac{\lambda^2}{64 \pi^2 s} \frac{\sqrt{s - 4 m_2^2}}{\sqrt{s - 4 m_1^2}}
}.
\end{equation}

\clearpage

\subsection{Level 2 - Dark Matter Capture as a Function of Time}

\subsubsection*{Problem Statement}
Suppose $C$ is the capture rate of dark matter in an astrophysical
body. Let $C_{A}$ be the dark matter annihilation rate per effective
volume. Then an approximate Boltzmann equation governing the number
$N$ of dark matter particles in the astrophysical body is
\[
\frac{d N}{dt}=C-C_{A}N^{2}.
\]
If initially, $N(0)=0$, what is $N(t)$ as a function of time?

\subsubsection*{Answer Requirements}

Provide the answer in the form of the \texttt{python} code. Implement the following function.
\begin{python}
def answer(C: float, C_A: float, t: float) -> float:
    pass
\end{python}

\subsubsection*{Comments about the Problem}
\textit{This problem mainly belongs to astrophysics, specifically involving dark matter dynamics in celestial bodies. It is typically encountered in advanced undergraduate or graduate-level courses and requires knowledge of differential equations and kinetic theory. This type of analysis is also important for understanding dark matter detection and its astrophysical implications.}

\subsubsection*{Solution}
We can integrate by quadrature.
\begin{equation}
\int\frac{dN}{C-C_{A}N^{2}}=t.
\end{equation}
We can express the integrand as a sum of two fractions:
\begin{align*}
\frac{1}{C-C_{A}N^{2}} & = \frac{1}{\sqrt{C}-\sqrt{C_{A}}N}\frac{1}{\sqrt{C}+\sqrt{C_{A}}N}\\
 & = \frac{1}{2\sqrt{C}}\left[\frac{1}{\sqrt{C}-\sqrt{C_{A}}N}+\frac{1}{\sqrt{C}+\sqrt{C_{A}}N}\right]\numberthis.
\end{align*}
Integrating, we find
\begin{align*}
t+K & = \frac{1}{2\sqrt{C}}\left[\frac{-1}{\sqrt{C_{A}}}\ln\left(\sqrt{C}-\sqrt{C_{A}}N\right)+\frac{1}{\sqrt{C_{A}}}\ln\left(\sqrt{C}+\sqrt{C_{A}}N\right)\right]\\
 & = \frac{1}{2\sqrt{C_{A}C}}\ln\left(\frac{\sqrt{C}+\sqrt{C_{A}}N}{\sqrt{C}-\sqrt{C_{A}}N}\right)\numberthis
\end{align*}
where $K$ is an integration constant. Setting the boundary condition
$N=0$ at $t=0$, we find
\[
K=0.
\]
We find the solution 
\begin{equation}
\boxed{N=\frac{\sqrt{C}}{\sqrt{C_{A}}}\frac{\left(e^{2\sqrt{C C_A}t}-1\right)}{\left(e^{2\sqrt{C C_A}t}+1\right)}}.
\end{equation}
Note that it is easy to check that it reaches the obvious steady state
in the limit $t\rightarrow\infty$.

\clearpage

\subsection{Level 2 - A 3-State QM Problem}

\subsubsection*{Problem Statement}
The Hamiltonian of a three-level system is given as $H = \begin{pmatrix}
    E_a & 0 & A \\
    0 & E_b & 0 \\
    A & 0 & E_a \\
\end{pmatrix}$ where $A$ is real. The state of the system at time $t=0$ is (in this basis) $\psi(t=0) = \frac{1}{\sqrt{2}}\begin{pmatrix}1 \\
1\\
0\end{pmatrix}$ What is the expectation value of the energy at time $t$?

\subsubsection*{Answer Requirements}
Provide the answer in the form of \texttt{python} code. Implement the following function
\begin{python}
def expectation_value(A: float, E_a:float, E_b:float, t:float) -> float:
    pass
\end{python}

\subsubsection*{Comments about the Problem}
\textit{This problem belongs to quantum mechanics, focusing on multi-level quantum systems found in areas like quantum optics or molecular physics. It is typically encountered in advanced undergraduate or early graduate-level courses and requires knowledge of linear algebra, time evolution, and the calculation of expectation values in quantum mechanics.}

\subsubsection*{Solution}
The eigenstates are easily found to be $\frac{1}{\sqrt{2}}(1,0,\pm 1)^T$ and $(0,1,0)^T$ with corresponding energies $E_a\pm A$, $E_b$. Let us denote them as $|1\rangle$, $|2\rangle$ and $|3\rangle$. Given state $\psi$ is decomposed as $\frac{1}{2}(|1\rangle +|2\rangle) + \frac{1}{\sqrt{2}}|3\rangle$, the expectation of energy stays constant: 
\begin{equation}
    \langle E\rangle = \frac{1}{4}((E_a+A)+(E_a-A)) + \frac{1}{2}E_b =\boxed{ \frac{1}{2}(E_a+E_b)}.
\end{equation}

\vspace{2cm}

\subsection{Level 1 - Blackbody in $d$ Dimensions}

\subsubsection*{Problem Statement}
Assume we live in a 4+1 dimensional spacetime. How does the total energy density of a black body scale with temperature T. Find the exponent $n$ in the expression $u \propto T^{n}$

\subsubsection*{Answer Requirements}
Provide the answer in the form of \texttt{python} code. Implement the following function
\begin{python}
def answer() -> float:
    pass
\end{python}

\subsubsection*{Comments about the Problem}
\textit{This problem lies in the realm of statistical mechanics and thermodynamics applied to higher-dimensional spacetimes, a topic typically encountered at the undergraduate level in theoretical physics.}

\subsubsection*{Solution}
The density of states scales as $k^{D-1}dk$ in D spatial dimensions giving 
$T^{D+1}$ scaling for the total energy density. Hence, $\boxed{n=5}.$

\vspace{2cm}

\subsection{Level 1 - Boosted Parabolic Trajectory}

\subsubsection*{Problem Statement}
Consider a situation where a space-probe very briefly fires its rockets while passing a planet of mass \(M\) at periapsis, its nearest point to the planet. Suppose that the probe is on a parabolic trajectory and at periapsis, when travelling at velocity $v_e$, it results in a boost of $\delta v$. What will be its speed once it escapes the planet's gravitational field only in terms of $v_e$ and $\delta v$?

\subsubsection*{Answer Requirements}
Provide the answer in the form of \texttt{python} code. Implement the following function
\begin{python}
def speed(v_e: float, delta_v:float) -> float:
    pass
\end{python}

\subsubsection*{Comments about the Problem}
\textit{This problem is part of orbital mechanics, typically covered at the undergraduate or advanced high school level in physics. It involves principle of energy conservation in Newtonian gravity.}

\subsubsection*{Solution}
Conservation of energy gives $\frac{1}{2}m(v_e+\delta v)^2-\frac{mMG}{r_p} = \frac{1}{2}mv^2_\infty$. We also know that $\frac{1}{2}m(v_e)^2-\frac{mMG}{r_p} = E = 0$ for the parabolic trajectory. We can solve for $v_e$: $v_e = \sqrt{\frac{2MG}{r_p}}$. Then we can substitute it in the first equation and get:
\begin{equation}
\boxed{v_\infty = \delta v\sqrt{1+\frac{2v_e}{\delta v}}}.
\end{equation}

\end{document}